\documentclass[runningheads]{llncs}

\usepackage{eccv}

\usepackage{eccvabbrv}

\usepackage{graphicx}
\usepackage{booktabs}

\usepackage[accsupp]{axessibility}  % Improves PDF readability for those with disabilities.

\usepackage{hyperref}

\usepackage{caption}
\usepackage{comment}
\usepackage{subcaption}
\usepackage{array}
\newcolumntype{H}{>{\setbox0=\hbox\bgroup}c<{\egroup}@{}}
\usepackage{multirow}
\usepackage{multicol}
\usepackage{amsmath}
\usepackage{arydshln} % for dotted vertical line in table
\usepackage[ruled]{algorithm2e} % For algorithms

\SetAlFnt{\small}
\SetAlCapFnt{\small}
\SetAlCapNameFnt{\small}
\SetAlCapHSkip{0pt}
\usepackage{wrapfig}
\usepackage{makecell}

\begin{document}

% ---------------------------------------------------------------
\title{StyleCity: Large-Scale 3D Urban Scenes Stylization}
\titlerunning{StyleCity: Large-Scale 3D Urban Scenes Stylization}

\author{Yingshu Chen \and
Huajian Huang{$^\dag$} \and
Tuan-Anh Vu \and
Ka Chun Shum \and
Sai-Kit Yeung}

\authorrunning{Y.~Chen et al.}

\institute{The Hong Kong University of Science and Technology\\
% \email{\{ychengw,hhuangbg,tavu,kcshum\}@connect.ust.hk, saikit@ust.hk} \\
{$^\dag$}Corresponding author}

\maketitle

\begin{figure*}
	\begin{center} 		\includegraphics[width=\linewidth]{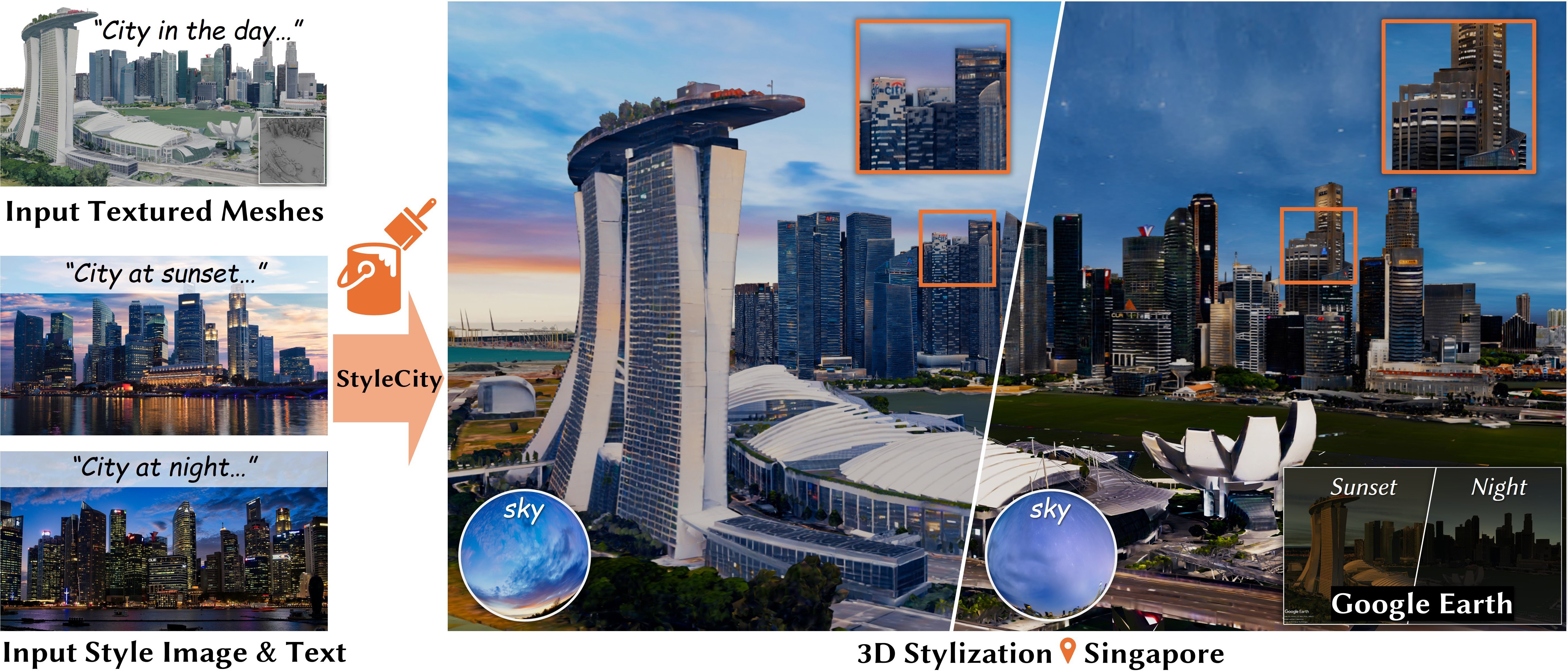}
	\end{center}
	\caption{3D city stylization in magic times of a day. Our proposed novel StyleCity framework can automatically stylize the textured meshes of large-scale urban scenes and generate harmonic omnidirectional sky backgrounds in a controllable manner with input style prompts, i.e., image and text references. Our effective solution has many potential applications, such as making city exploration in Google Earth a more personalized and visually captivating experience.}
    \label{fig:teaser}
\end{figure*}
\begin{abstract}
  Creating large-scale virtual urban scenes with variant styles is inherently challenging. 
  To facilitate prototypes of virtual production and bypass the need for complex materials and lighting setups, we introduce the first vision-and-text-driven texture stylization system for large-scale urban scenes, StyleCity. Taking an image and text as references, StyleCity stylizes a 3D textured mesh of a large-scale urban scene in a semantics-aware fashion and generates a harmonic omnidirectional sky background. To achieve that, we propose to stylize a neural texture field by transferring 2D vision-and-text priors to 3D globally and locally. During 3D stylization, we progressively scale the planned training views of the input 3D scene at different levels in order to preserve high-quality scene content. We then optimize the scene style globally by adapting the scale of the style image with the scale of the training views. Moreover, we enhance local semantics consistency by the semantics-aware style loss which is crucial for photo-realistic stylization. Besides texture stylization, we further adopt a generative diffusion model to synthesize a style-consistent omnidirectional sky image, which offers a more immersive atmosphere and assists the semantic stylization process. The stylized neural texture field can be baked into an arbitrary-resolution texture, enabling seamless integration into conventional rendering pipelines and significantly easing the virtual production prototyping process. Extensive experiments demonstrate our stylized scenes' superiority in qualitative and quantitative performance and user preferences. Project page: \url{https://chenyingshu.github.io/stylecity3d}.
\end{abstract}

\section{Introduction}
Large-scale urban digital twins, reconstructed from satellite and UAV imagery, are widely utilized in various domains such as video games, 3D city visualization, and virtual reality applications. In virtual production, the demand for visually appealing and unique representations of reconstructed meshes is high. Yet, simulating the visual appearance of virtual scenes—like the time simulation in Google Earth—often fails to produce truly impressive visuals. This shortfall is typically due to the extensive need for laborious material design and meticulous illumination setups for each instance.
To achieve automatically re-styling scene appearances, we seek to utilize 3D neural style transfer techniques.

%To address this challenge, we propose an automated system for the re-styling of virtual scene appearances, utilizing 3D neural style transfer techniques. 

%for textured mesh stylization in this work. It serves as a lightweight alternative f

Style transfer is a fundamental problem that seeks to automatically transfer the target style according to a given reference while maintaining the original structure.
With the maturity of 2D neural stylization~\cite{gatys2016image, huang2017arbitrary, kwon2022clipstyler}, 3D scene stylization has become more flexible and achievable by lifting content and style features priors from 2D to 3D. 
For example, image-guided mesh stylization \cite{hollein2022stylemesh} and text-guided object and room texturing \cite{cao2023texfusion, hwang2023text2scene}
realize 3D neural style transfer on meshes with 2D neural visual or textual priors.
Due to the 2D-to-3D stylization fashion, we have to do view planning additionally to obtain comprehensive training views covering surfaces in balance \cite{michel2022text2mesh, richardson2023texture, chen2023text2tex}. 
However, these mesh stylization works primarily focus on object- and room-scale scenarios. Their view planning and style optimization methods are not suitable for large-scale scenes.
Particularly, explicit texture or latent features optimization is resource-intensive. It results in affordable texture resolutions that are too limited to scale up to large-scale scenes.
Moreover, these methods seldom address local semantic consistency in stylization. Given that large-scale scenes possess significant semantic complexity, we contend that maintaining semantic style feature consistency at a local level is crucial, particularly for achieving photo-realistic stylization. However, how to maintain global and local style harmony is challenging.

To narrow the gap, we propose to use a UV-based neural texture field to model scene appearances. The UV-based neural texture field is composed of a multi-resolution feature grid for UV encoding and a multilayer perceptron (MLP) for feature decoding. As a compact hybrid representation, it reduces resource usage while maintaining global style consistency to some extent.
%and harmony within the stylized 3D scenes. UV
After initializing the neural texture field with the original texture map, we perform optimization by minimizing global content and style losses among multiple planned views.

We employ a multi-scale progressive rendering strategy combined with scale-adaptive style reference enhancement during the texture field optimization to achieve high-fidelity texture renovation. The training views are progressively augmented into finer views at different levels with random translation and zoom-in. Simultaneously, we obtain multi-scale style reference features by randomly cropping the input reference image. Structural similarity scores guide us in adaptively selecting the scale of the style references to best match the scale of the current training views, ensuring the appropriate transfer of style patterns at different levels and providing precise and effective supervision. This approach allows us to progressively transfer style patterns to the 3D scene while maintaining fine-grained content structures.

In addition to global style optimization, our semantics-aware style loss preserves local style consistency, contributing to a holistic style enhancement. We also leverage a generative diffusion model to omnidirectionally denoise latent features and synthesize a visual and textual style-aligned omnidirectional sky image. Serving as the background of urban scenes, it provides a consistent and immersive atmospheric context for the stylized scene, despite having no illumination effect on the foreground during optimization and rendering. By converting the stylized neural texture field into a classical UV texture map, we efficiently render stylized novel views of the 3D scene in conventional pipelines for real-time applications. Fig. \ref{fig:whole_flow} illustrates the overview of our StyleCity framework.

To summarize, the contributions of this work include, 1) we propose a novel 3D textured mesh stylization framework using a neural texture field; 2) 
we introduce a multi-scale progressive rendering strategy and a scale-adaptive style optimization method to globally transfer 2D style to large-scale 3D scenes while preserving structural photorealism.
3) we propose a new local semantics-aware style loss to enhance local style correspondence;
4) We adopt the diffusion model to generate high-resolution 360-degree images as environment backgrounds enhancing the appeal of the stylized urban scene; 5) The comprehensive experiments prove that we realize high-quality 3D urban scene stylization without any manual rendering material and illumination setups. The automatic pipeline guided by vision and text references can be a quick substitute for virtual production and inspire scene design.

\section{Related Work}
% 3D style transfer
In early works for 3D stylization, researchers applied interactive and iterative optimization on 3D data such as mesh texture and geometry to achieve non-photorealistic stylization \cite{fivser2016stylit, sykora2019styleblit, hauptfleisch2020styleprop, liu2019cubic}. Later, some methods \cite{yin20213dstylenet, cao2020psnet} trained generative neural networks with large 3D data and facilitated 3D appearance and geometry style transfer. For example, 
% FrankenGAN \cite{kelly2018frankengan} produces building texture using generative networks with texture style reference.
3DStyleNet \cite{yin20213dstylenet} transferred both shape and texture between 3D models with a textured mesh exemplar by a 3D generative model. 3D data collection is tedious and time-consuming. Fortunately, empowered by large-data pre-trained 2D neural models such as VGG, CLIP \cite{radford2021learning}, and Stable Diffusion \cite {rombach2021high}, there appeared a lot of automatic vision or text example-based 2D neural style transfer approaches \cite{gatys2016image, huang2017arbitrary, luan2017deep, patashnik2021styleclip, gal2022stylegan, kwon2022clipstyler, kim2022diffusionclip, yang2023zero}. Inspired by this, researchers tend to lift these 2D priors to 3D representations for automatic 2D example-based 3D stylization.
Next, we review 3D neural style transfer methods on meshes and neural field representations, which transfer style with vision or text guidance and preserve to some degree original structure and geometry. We refer to \cite{chen2023advances} for more 3D stylization works on other 3D data.

% \subsection{Mesh-based Stylization}
\noindent\textbf{Mesh-based Stylization.} Kato \etal \cite{kato2018neural} proposed a neural renderer for mesh such that it enables artistic geometric and appearance stylization for the object mesh using image-example-based neural style transfer technique \cite{gatys2016image}.  
When it turns to image-guided 3D indoor scene artistic stylization, StyleMesh \cite{hollein2022stylemesh} introduced a texture optimization scheme specifically designed for 3D reconstructed indoor meshes, incorporating depth and angle information.
Later, with the emergence of large-language model (LLM), text-guided stylization on mesh geometry and appearance \cite{michel2022text2mesh, ma2023x}, text-guided texture transfer or synthesis \cite{jin2022language, lei2022tango, richardson2023texture, chen2023text2tex, yang20233dstyle} became a popular topic in academia.
\cite{jin2022language} accomplished 3D semantic style transfer for indoor textures with text guidance. \cite{michel2022text2mesh, ma2023x} achieved text-guided object geometric and texture stylization using a vertex-based neural style field network. They largely rely on fine-grained triangulation on the mesh and cannot synthesize detailed patterns or motifs. \cite{lei2022tango, yang20233dstyle} took object materials into account, achieving text-guided texture stylization with material properties. With the advent of large-data pre-trained diffusion models \cite{rombach2021high}, recent works \cite{richardson2023texture, chen2023text2tex, yang2023dreamspace, cao2023texfusion, knodt2023consistent, song2023roomdreamer} proposed to use the pre-trained diffusion models \cite{zhang2023adding, rombach2021high, bar2023multidiffusion} to better supervise text-guided photorealistic texture synthesis for objects or rooms. 
\cite{yang20233dstyle} utilized 2D diffusion priors \cite{poole2023dreamfusion} for object texture stylization. However, globally optimizing explicit texture requires huge computational resources, i.e., GPU memory. These methods generally focus on small-scale scenes and lack scalability. 

% \subsection{Neural Field Stylization} 
\noindent\textbf{Neural Field Stylization.} 
There appeared numerous neural-field-based stylization works for implicit reconstructed scenes \cite{huang2021learning, chiang2022stylizing, huang2022stylizednerf, fan2022unified, nguyen2022snerf, zhang2023transforming, wang2023nerf, haque2023instructnerf}. Some data-driven generative training models \cite{huang2021learning, liu2023stylerf} support zero-shot stylization for a scene, but require large training references and sophisticated model design. By contrast, feature optimized-based approaches usually have more impressive results without other style dilutions for one model \cite{zhang2022arf, zhang2023ref, pang2023locally}. For example, ARF \cite{zhang2022arf} proposed a nearest-neighbor feature matching loss for local style matches in latent space transferring local artistic patterns.
Empowered by 2D generative models, there are generative-model supervised optimization methods, such as InstructN2N \cite{haque2023instructnerf}, which utilized an alternative multi-view updating scheme to progressively optimize the 3D scene. However, multi-view supervision by independent one-shot generated images easily tends to collapse due to 3D inconsistency.
These neural field stylization works utilized 3D neural field representations to maintain global style consistency and render novel views via volume rendering. However, it is not wise to 
convert mesh-based rendering into volume rendering for mesh stylization which is prone to generating blurry views.

\vskip 0.2cm
Compared to the aforementioned methods, our stylization system is the first work devised for mesh-based large complex outdoor urban scenes. Importantly, we explore the UV-based neural texture field to model mesh appearance and essentially take large-scale characteristics into account to devise style optimization strategies, achieving highly harmonious stylization in urban scenes.
%---------------------------------------------------

\section{Methodology} \label{sec:method_tex_opt}
\subsection{Data Pre-processing}\label{sec:data_prepare}
To perform style transfer on a reconstructed mesh without known camera poses, it is indispensable to tackle view planning. In addition, we need to annotate 3D mesh with semantic labels such that we can efficiently obtain pixel-wise multi-view consistent semantic information during stylization.
\begin{figure}[t]
    \hfill
    \begin{subfigure}[b]{0.48\linewidth}
        \centering
        \includegraphics[width=\linewidth]{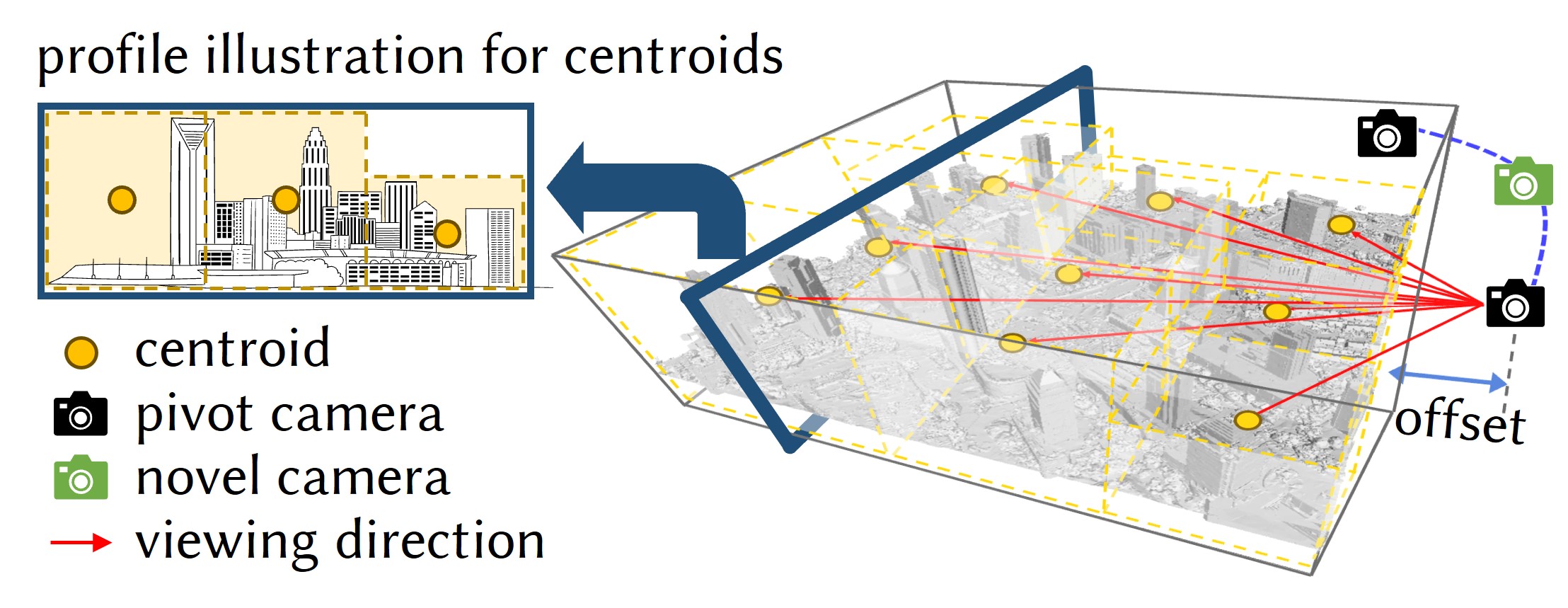}
        \caption{Planning illustration}\label{fig:method_view_plan}
    \end{subfigure}    
    \begin{subfigure}[b]{0.48\linewidth}
        \centering        \includegraphics[width=0.8\linewidth]{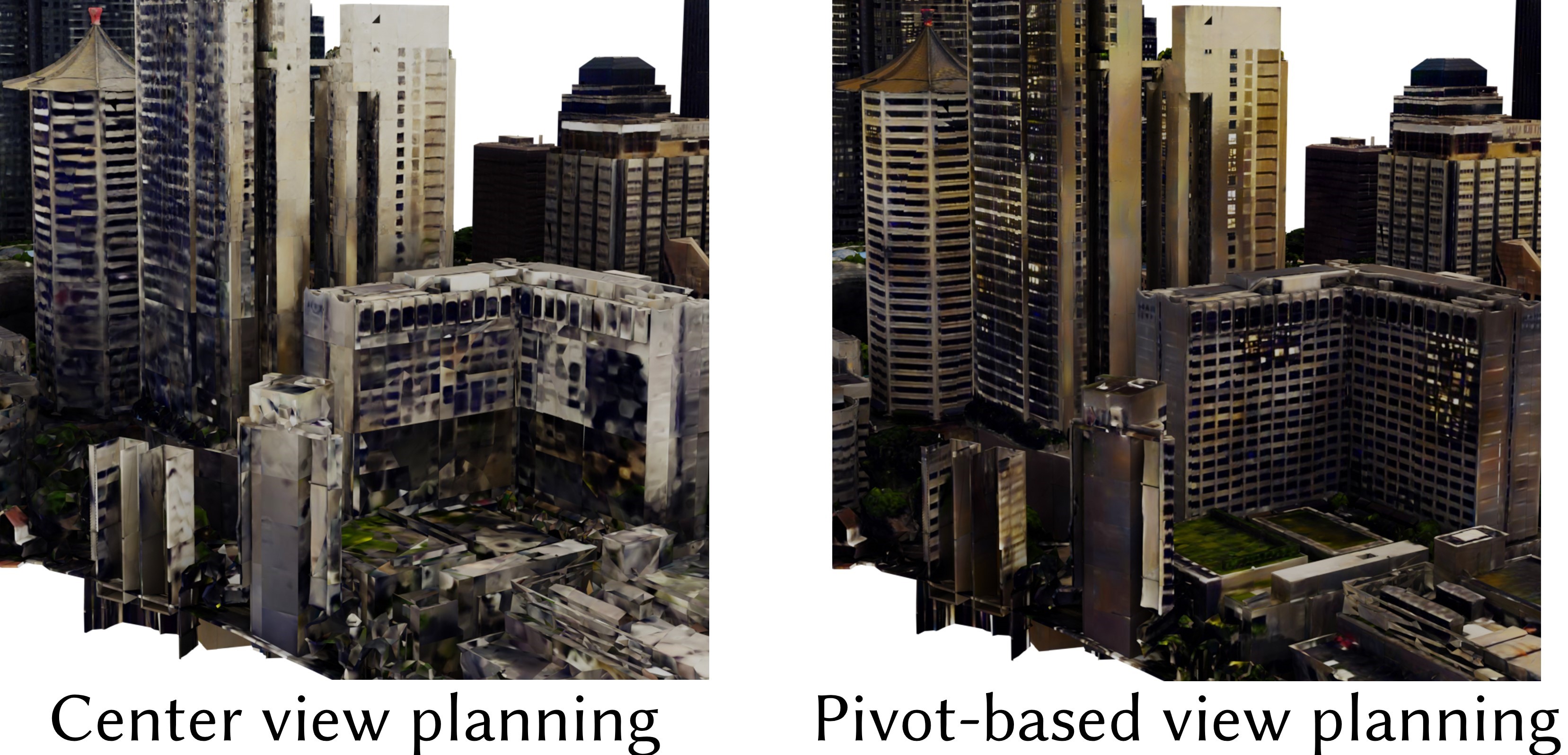}
        \caption{Stylization comparison}\label{fig:method_view_plan_comp}
    \end{subfigure}
    \caption{Pivot-based view planning.}\label{fig:view_plan}
\end{figure}

\noindent\textbf{Pivot-Based View Planning.}\label{sec:method_view_plan}
Naive scene center view planning methods applied to stylize single objects \cite{michel2022text2mesh, richardson2023texture, kim2019transport} are insufficient for large-scale surface coverage. Direct application results in blurry texture with broken structure after stylization as shown in Fig. \ref{fig:method_view_plan_comp}. 
We developed a pivot-based view planning method that overcome their limitations. 
We uniformly sample $P$ camera positions on the upper and side faces of the mesh bounding box with a proper offset. The mesh is then subdivided into $r$ sub-regions, and the centroids of these regions are used as camera viewing points. Finally, we obtain $P\times r$ pivot views, locating at $P$ pivot positions and looking at $r$ centroids.
Fig. \ref{fig:method_view_plan} shows a pivot view planning example.
Pivot views cover the majority of visible surfaces with associated semantics, serving as initialization for the novel views (Fig. \ref{fig:method_view_plan} green camera) augmenting training, referred to Sec.~\ref{sec:method_multi_scale}  Multi-Scale Progressive Optimization.

\begin{figure*}[t]
    \begin{center}     
        \includegraphics[width=\linewidth]{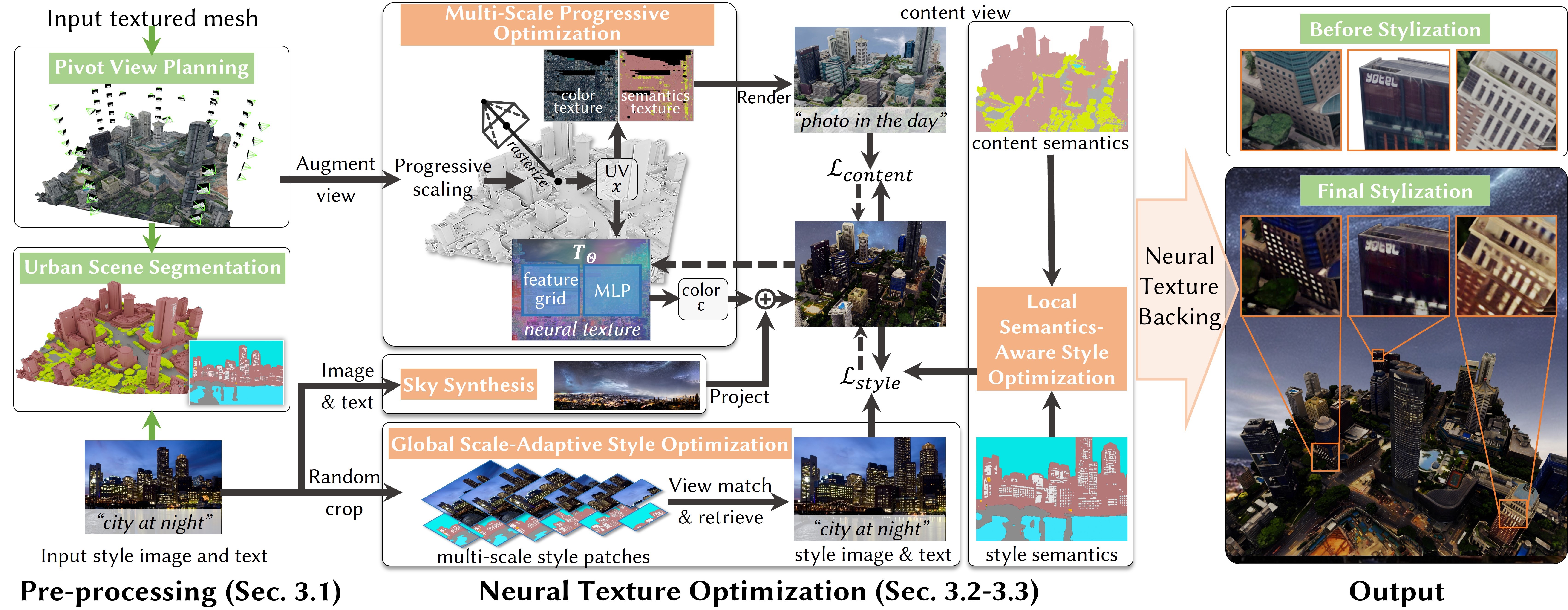}
    \end{center} %height=0.42\linewidth
    \caption{Framework overview of StyleCity. Taking style image and text references as input, StyleCity optimizes the neural texture field in a semantics-aware fashion with progressively scaled training views. Synthesized omnidirectional sky enhances style atmosphere and assists semantic style supervision.}
    \label{fig:whole_flow}
\end{figure*}

\noindent\textbf{Urban Scene Segmentation.}
For semantics-aware urban scenes stylization, we particularly consider classes of interest in architectural scenarios for segmentation, including "\textit{sky}", "\textit{building}",  "\textit{window}", "\textit{road}", "\textit{person}", "\textit{plant}", "\textit{car}", "\textit{water}" and "\textit{lights}". We adapted and fine-tuned Mask2Former \cite{cheng2022masked}, a pre-trained 2D semantic segmentation model, and implemented an automated 2D-to-3D segmentation tool via texture mapping for input 3D models. We also segment the input style reference image using the same segmentation model. 

Please refer to the supplementary for more data pre-processing details.

\subsection{Neural Texture Field} \label{sec:method_tex_repr}
\textit{Definition.}
Aiming at large urban scenes with megapixel or gigapixel texture images, we employ neural texture field representation and re-parameterize the huge texture map into a 2D continuous function. We define the neural texture field $T_{\Theta}(\cdot)$ (shown in \textit{neural texture} in Fig.~\ref{fig:whole_flow}) by an MLP that maps a normalized UV texture coordinate ($x \in \mathbb{R}^2$) with hash grid-based feature encoding \cite{muller2022instant} to a color RGB value ($\varepsilon \in \mathbb{R}^3$): 
\(T_{\Theta}(x) = \varepsilon \). 
The neural texture field theoretically supports arbitrary texture resolution with a condensed amount of parameters. It is compact but sufficient to represent scene appearance with on average 90\% texture size compression from experiments.
Moreover, neural continuous representation learns smooth interpolation among UVs, tolerating inevitably incomplete training views and maintaining global style consistency. By contrast, explicit texture optimization may lead to an out-of-memory issue with high resolution or a degraded discrete stylized texture with insufficient resolution.

\textit{Neural Rendering.}
Given a camera pose, we rasterize mesh and retrieve UVs
to query texture model $T_{\Theta}(\cdot)$. Accordingly, we then obtain corresponding texture RGB values, and then resemble values into a rendered image. 

\textit{Conventional Rendering.}
To get stylized texture with width $W_{tex}$ and height $H_{tex}$, we query $T_{\Theta}(UV)$ with $U \in [0,1]$ with $W_{tex}$ uniformly sampled values and $V \in [0,1]$ with $H_{tex}$ uniformly sampled values. The output texture with mesh can be rendered via the conventional rendering pipeline or applied to VR.

\textit{Initialization.}
Before stylization, we distill original full-resolution texture content to neural texture by a simple MSE photometric loss \(
    \mathcal{L}_{tex} =  \sum_{\mathit{x \in UV}} \| \mathit{T}_c(x) -  T_{\Theta}(x)\|^2_2,
\)
which $T_c$ is original texture image, $x$ is a random sampled batch of UVs. After distillation, the neural texture is initialized with original texture style.

\subsection{Content and Style Joint Progressive Optimization} \label{sec:method_joint_opt}
We jointly optimize the neural texture field through multiple views with both source content and target style constraints in each iteration.
As illustrated in Fig. \ref{fig:whole_flow}, we sample a viewpoint, render the content view and its segmentation, and obtain the stylized view from neural texture. To achieve high-quality stylization, we augment training views into multi-scales for progressive optimization.

\subsubsection{Multi-Scale Progressive Optimization.} \label{sec:method_multi_scale}
\begin{wrapfigure}{r}{0.5\textwidth}
    \centering
    \includegraphics[width=\linewidth]{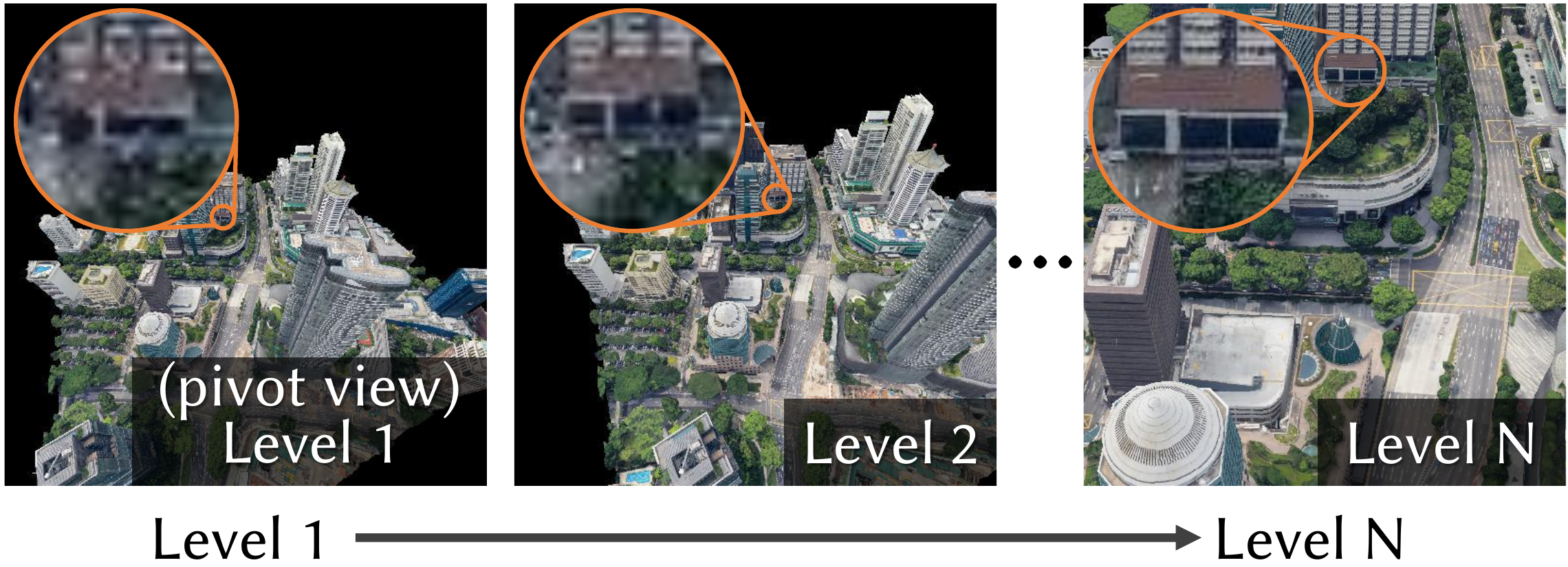}
    \caption{Multi-scale training views augmented based on same pivot cameras.
    } 
    \label{fig:multi_scale_render} 
\end{wrapfigure}

% \noindent\textbf{Multi-Scale Progressive Optimization.}\label{sec:method_multi_scale}
During optimization, we randomly sample novel views along Bezier curves with nearby planned pivot cameras as control points in order to enlarge covered angles, as shown in Fig.~\ref{fig:view_plan} green camera. Furthermore, we randomly translate novel view cameras horizontally and vertically in an offset of \( [-\alpha, \alpha] \times dist_{cam}\), where $dist_{cam}$ is the distance from camera to model center, $\alpha$ is hyperparameter (default 0.25). However, although the dynamic view augmentation ensures each surface is comprehensively stylized and globally harmonic, we found that the stylized mesh still lacks high-frequency details. To address this issue, we progressively scale sampled views at a preset field of view (FoV) with a "zoom-in" effect during training, as demonstrated in Fig.~\ref{fig:multi_scale_render}. 
Specifically, FoV will decrease uniformly (e.g. from 90$^\circ$ to 20$^\circ$) while the number of novel sampling views increases at each progressive optimization level.  
We set all rendered images in a fixed resolution and the default number of levels N=5.
The experiment validates that dynamic multi-scale progressive optimization is effective for high-fidelity large-scale scene stylization. For simplicity, we only demonstrate the optimization process and loss equations for a single view at any optimization level in the subsequent description, unless explicitly stated otherwise.

\subsubsection{Content and Photorealism Preservation Optimization.}
% \noindent\textbf{Content and Photorealism Preservation Optimization.}
% During stylization, we utilize content features from rendered views and Laplacian values for supervision, aiming to preserve texture content and scene identity. To maintain content integrity and photorealism, we mask out the background areas during the training phase.
During stylization, we utilize not only content features but also Laplacian values of rendered views for supervision, preserving texture content and scene identity. During training, we mask out sky background areas.

Latent content feature loss can help preserve the basic appearance of the original texture, such as original colors and abstract structures, while remaining possible for style variations:
\begin{equation}
 \label{eq:content_loss}
    \mathcal{L}_{c} =  \sum_{\mathit{l\in l_c}} \| \mathit{F}^l(c) - \mathit{F}^l(z)\|^2_2, 
\end{equation}
where $F^{l}(\cdot)$ denotes embedded features of $l^{th}$ layer in VGG-19 \cite{simonyan2015very}
taking an image as input. $c$ and $z$ are rendered content image and stylized image on the current viewpoint respectively, the feature layer $l_c = \{4\}$ is used.

Beyond the content feature loss, we incorporated a photorealism regularizer, akin to the approach in~\cite{luan2017deep}, to maintain high content fidelity and preserve identity, such as by maintaining texture edges irrespective of chrominance.
The regularization term is formulated as 
\begin{equation}
\mathcal{L}_{pht} = \sum_{h=1}^{3} (\mathcal{V}^{z}_{h})^T \mathcal{M}^{c} \mathcal{V}^{z}_{h},
\end{equation}
where $h$ denotes a RGB channel, and $\mathcal{V}^{z}$ is the flattened version of the stylized view. The kernel $\mathcal{M}^{c}$ is the matting Laplacian matrix \cite{levin2007closed} of the content view.

\subsubsection{Global Scale-Adaptive Style Optimization.}
% \noindent\textbf{Global Scale-Adaptive Style Optimization.}
Global style optimization is responsible for globally transferring style features to the neural texture field for overall atmosphere alignment.

\begin{wrapfigure}{tr}{0.49\textwidth} 
    \includegraphics[width=\linewidth]{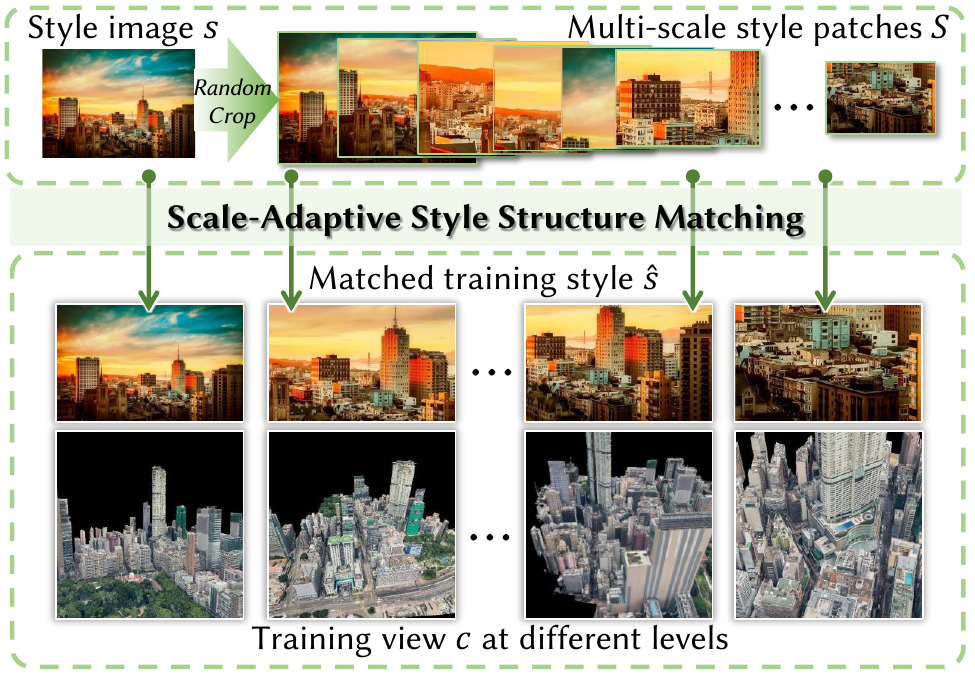}
    \caption{Scale-adaptive style structure matching. We match multi-scale styles based on structure similarity between training views and style patches.}
    \label{fig:method_multi_style_match}
\end{wrapfigure}

As described before, our multi-scale progressive optimization strategy renders different-scale images at different training levels. To keep the style consistent, the style reference image should adapt to the scale of the training views. It means that style reference features should be scale-adaptive.
To this end, we exploit patch-based style references at different resolutions for optimization. During optimization, we randomly crop the input style image $s$ into multi-scale style patches $S$ with different dimensions. The long side of each style patch is no shorter than 256 pixels. Then we select a structure-matched style patch $\hat{s}$ for each training view $c$, as illustrated in Fig.~\ref{fig:method_multi_style_match}. Inspired by \cite{shechtman2007matching, tumanyan2023disentangling}, we apply a self-similarity descriptor derived from latent features to represent spatial structure. The training style patch $\hat{s}$ is selected by minimizing the structure disparity between the training view $c$ and the style patch: 
\begin{equation}
    \begin{split}
        \arg\min_{\hat{s} \in S} \sum & \|\mathcal{D}(c)- \mathcal{D}(\hat{s})\|^2_2, 
    \end{split}
\end{equation}
where \(\mathcal{D}(\cdot) =  sim(\mathit{F}^{l_d}(\cdot),\mathit{F}^{l_d}(\cdot)^{T})\) is the self-similarity structure descriptor derived from channel-wise flattened VGG features, $l_d=\{3\}$ for simplicity, and  $sim(\cdot, \cdot)$ is the cosine similarity. With scale-adaptive style matching for optimization, we can better guarantee sufficient and appropriate style feature transfer.

\textit{Global Vision-and-Text Driven Style Loss.}
To quickly grab and generate meaningful new stylized texture, we deliver global multi-level visual style features to texture, by penalizing the discrepancy of global style feature distributions:
\begin{equation}
    \mathcal{L}_{gs} = \sum_{l \in l_s} (\| \mu(\mathit{{F}}^l(\hat{s})) - \mu(\mathit{F}^l(z))\|^2_2 + \| \sigma(\mathit{{F}}^l(\hat{s}))- \sigma(\mathit{F}^l(z))\|^2_2), 
\end{equation}
where $\hat{s}, z$ respectively denote the matched style image and output stylized image; feature layers $l_s = \{2,3,4\}$;  $\mu$ and $\sigma$ indicate mean and standard deviations.

To transfer textual style features for holistic style semantics harmony, we apply global and directional CLIP losses \cite{radford2021learning, gal2022stylegan}:
\begin{equation} \label{eq:clip_loss}
    \begin{split}
        \mathcal{L}_{clip} &=  \mathcal{L}_{glb} + \mathcal{L}_{dir},\\
        \mathcal{L}_{glb}(s_t, z) = 1 - sim(\mathit{E_{T}(s_t)}, \mathit{E_{I}(z)}),\, &
        \mathcal{L}_{dir}(c, z, c_{t}, s_{t}) = 1- sim(\Delta I, \Delta T), \\
        % \mathcal{L}_{dir}(c, z, c_{t}, s_{t}) = 1- sim&(E_{I}(z) - E_{I}(c), E_{T}(s_t) - E_{T}(c_t)), 
    \end{split}
\end{equation}
where $sim(\cdot, \cdot)$ is cosine similarity, $c_t$ and $s_t$ are source and style text prompts. $E_{I}$ and $E_{T}$ are image and text encoders, \(\Delta I=E_{I}(z) - E_{I}(c), \Delta T=E_{T}(s_t) - E_{T}(c_t)\).

\subsubsection{Local Semantics-Aware Style Optimization.}
For urban scenes with complex contexts, global style transfer easily leads to mismatches of style semantics. Thus, we introduce a local style optimization strategy for class-wise feature regularization to achieve more realistic stylization. 

\textit{Local Semantics-Aware Style Loss.}
The local semantics-aware optimization is performed by minimizing the mean and standard deviation of class-wise style features between the rendered stylized view $z$ and style reference $s$. 
We render a corresponding semantic segmentation map for the current training view using the semantics texture of the mesh while retrieving a semantic segmentation map for the matched style reference image. Based on semantic segmentation maps, we sample region of interest (ROI) features of each class for the training view and style reference image respectively: 
\begin{equation}
    f^{l}_{i}(z) = ROI_{i}(F^{l}(z)), \,
    {f}^{l}_{i}(s) = ROI_{i}(F^{l}(\hat{s})), 
\end{equation}
where $ROI_{i}(\cdot) \in \mathbb{R}^{1 \times M_{i}}$ indicates flattened vector of regions of $i^{th}$ class, $M_{i}$ is the number of non-zero value in the binary mask of $i^{th}$ class; $F^{l}_{i}(\cdot)$ is the $l^{th}$ layer VGG features in $i^{th}$ class regions, $\hat{s}$ is structure-matched style image.
Different from \cite{luan2017deep} which introduced style augmentation using masked Gram matrix taking non-ROI area into account, our semantic style loss only calculates the style feature of the target area representing a more accurate feature distribution. Moreover, the ROI features have different dimensions among different views. With $N_{cls}$ classes of interest, we formulate local semantics-aware style as: 
\begin{equation}
    % \begin{split}
        \mathcal{L}_{ls} = \sum_{l \in l_s} \sum_{i=1}^{N_{cls}} ( \|\mu(f^{l}_{i}(s))- \mu({f}^{l}_{i}(z)) \|^2_2  
        + \|\sigma(f^{l}_{i}(s))- \sigma(f^{l}_{i}(z))\|^2_2 ).
    % \end{split}
\end{equation}

\begin{figure}[t!]
    \centering
    \includegraphics[width=0.7\linewidth]{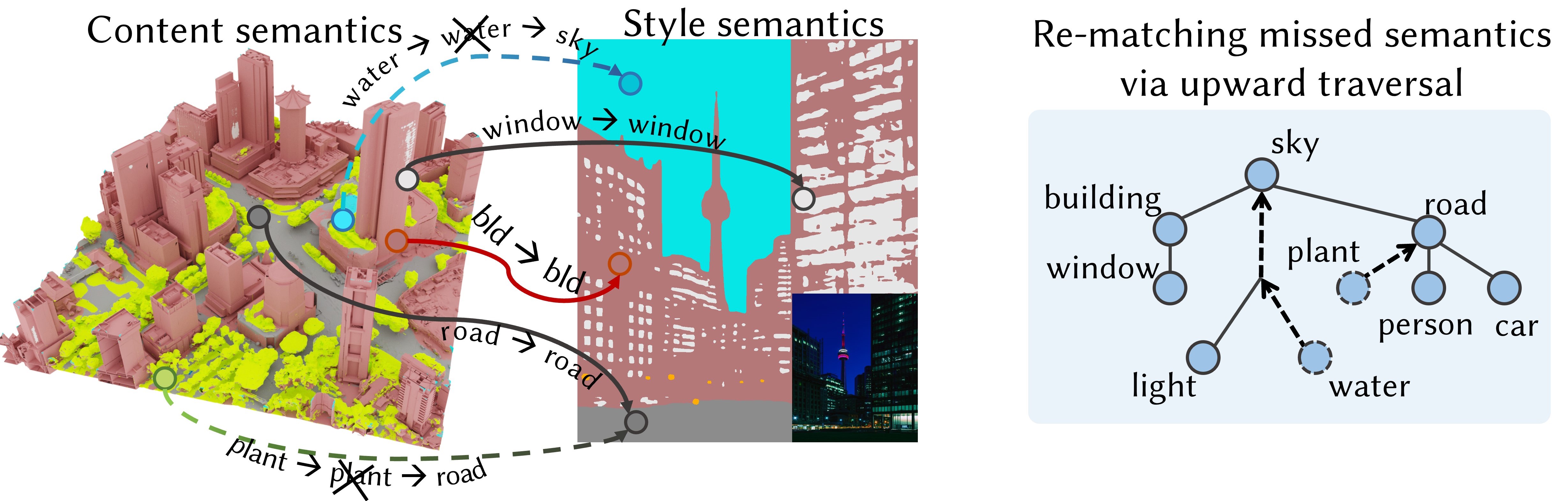}
    \caption{Semantics matching (left) and hierarchy graph for semantics re-matching (right graph). Style reference lacks semantics "\textit{water}" and "\textit{plant}" contained in the content scene. During semantic stylization, these semantics are re-matched to upper-level existing semantics in style reference.}
    \label{fig:method_2dstyle_sem_match}
\end{figure}
\textit{Semantic Hierarchical Re-match.} \label{sec:method_2dstyle_rematch}
During style transfer, it may occur mismatched and missed semantics between content view and style image as shown in Fig. \ref{fig:method_2dstyle_sem_match}. Content scene contains semantics  "\textit{building}", "\textit{window}", "\textit{plant}", "\textit{water}" and "\textit{road}", while style image lacks corresponding semantics such as "\textit{plant}" and "\textit{water}". To obtain references for missed semantics, we designed a hierarchy graph for semantics re-matching based on illumination effects in the real world. For example, cars, trees, and people on the road should share similar illumination of the road; all elements have the same ambient illumination from the sky, etc. In Fig.~\ref{fig:method_2dstyle_sem_match}, "\textit{water}" content can match "\textit{sky}" style luminance, "\textit{plant}" content can match "road" reference luminance.
If the matched style patch fails to provide style of some classes, we turn back to full-scale style reference for help.

\subsubsection{Total Objective.} 
We now have the texture stylization model with all losses: %the above-mentioned
\begin{equation}
\begin{split}
    \mathcal{L}_{total} &= \underbrace{\lambda_{c} \mathcal{L}_{c}  + \lambda_{pht} \mathcal{L}_{pht}}_{\mathcal{L}_{content}} + \underbrace{\lambda_{gs} \mathcal{L}_{gs} + \lambda_{ts} \mathcal{L}_{clip} +\lambda_{ls} \mathcal{L}_{ls}}_{\mathcal{L}_{style}} ,
    % \mathcal{L}_{total} &= \lambda_{c} \mathcal{L}_{c}  + \lambda_{pht} \mathcal{L}_{pht} + \lambda_{gs} \mathcal{L}_{gs} +
    % \lambda_{ts} \mathcal{L}_{clip} +\lambda_{ls} \mathcal{L}_{ls} ,
\end{split}
\end{equation}
By default, we set $\lambda_{c}=10, \lambda_{pht}=1e-3, \lambda_{ls}=1e-1, \lambda_{gs}=1, \lambda_{ts}=5$, and optimize each model for 20 epochs in 10 hours. 
%Usually, training for 4 hours gets plausible stylization. 
Particularly, we set $\lambda_{pht}=0$ in the first 2 epochs to get basic style patterns in case of style suppression from $\mathcal{L}_{pht}$ in the beginning.

\subsection{Style-Aligned Omnidirectional Sky Synthesis}
\begin{figure}[t]
    \centering
    \begin{subfigure}[b]{0.68\linewidth}
        \centering
        \includegraphics[width=\textwidth]{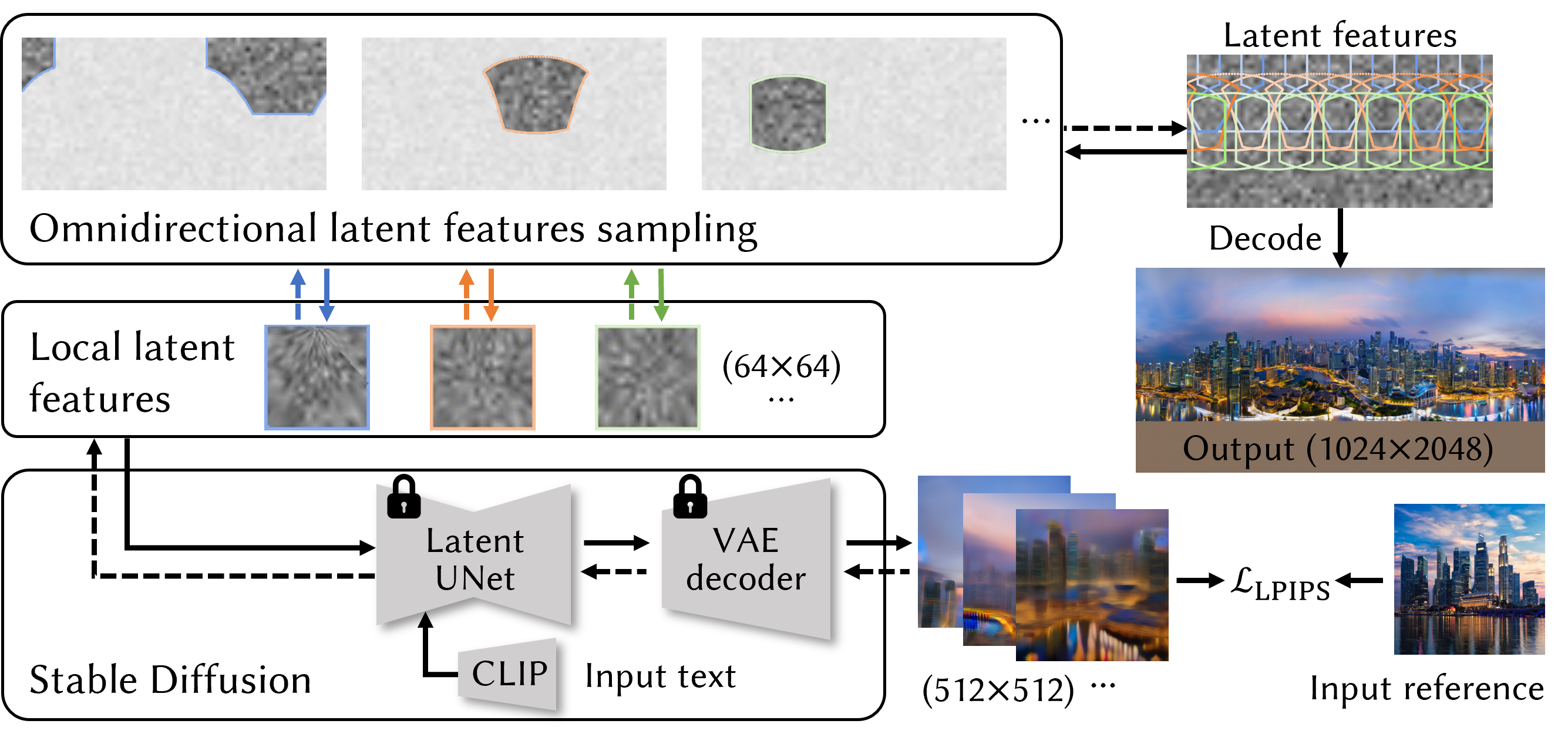}  
        \caption{Sky synthesis pipeline}\label{fig:sky_gen}
    \end{subfigure}      
    % \begin{subfigure}[b]{0.3\linewidth}
    \begin{subfigure}[b]{0.27\linewidth}
        \centering
         \includegraphics[width=\linewidth]{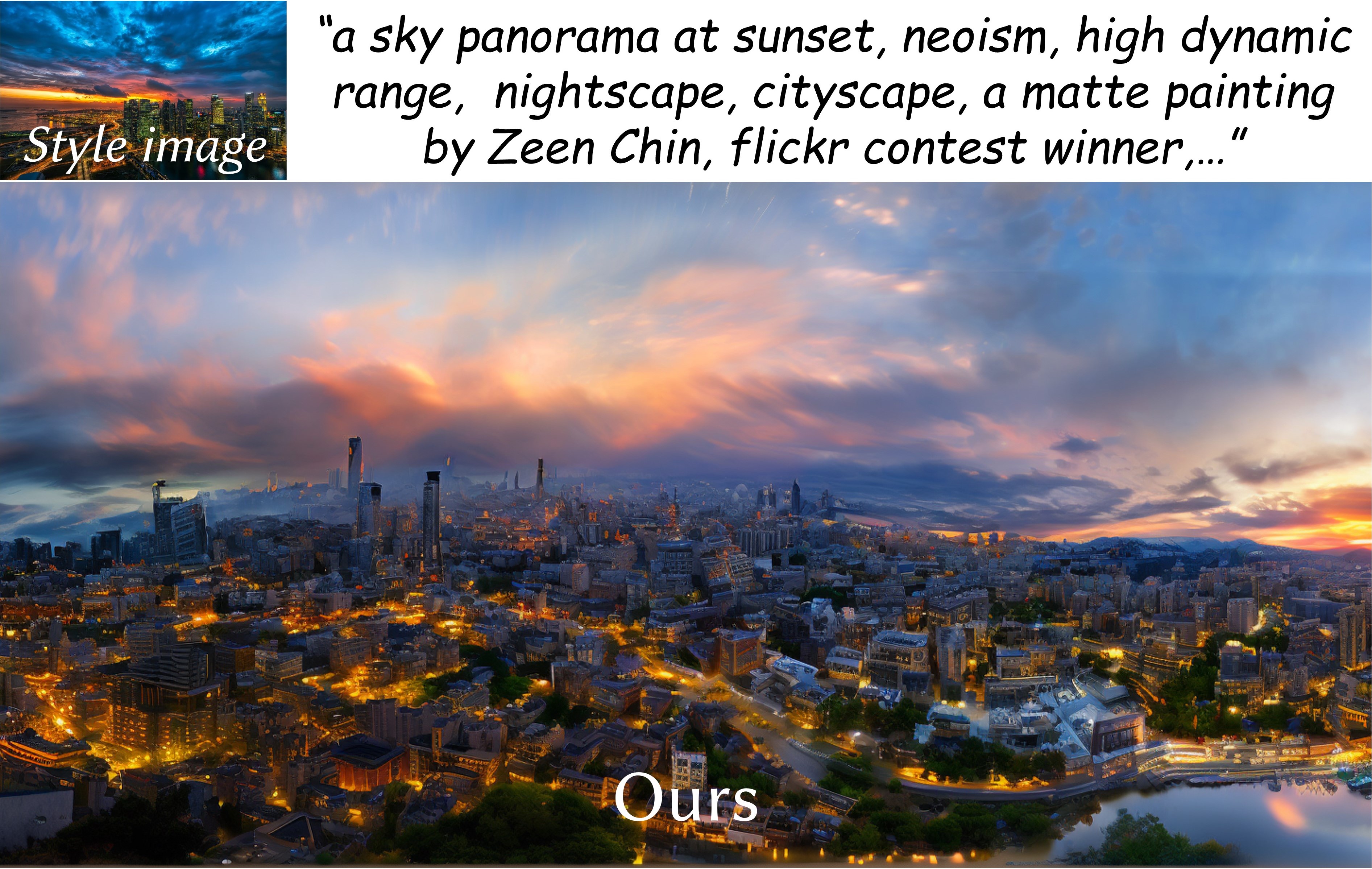}
         \includegraphics[width=\linewidth]{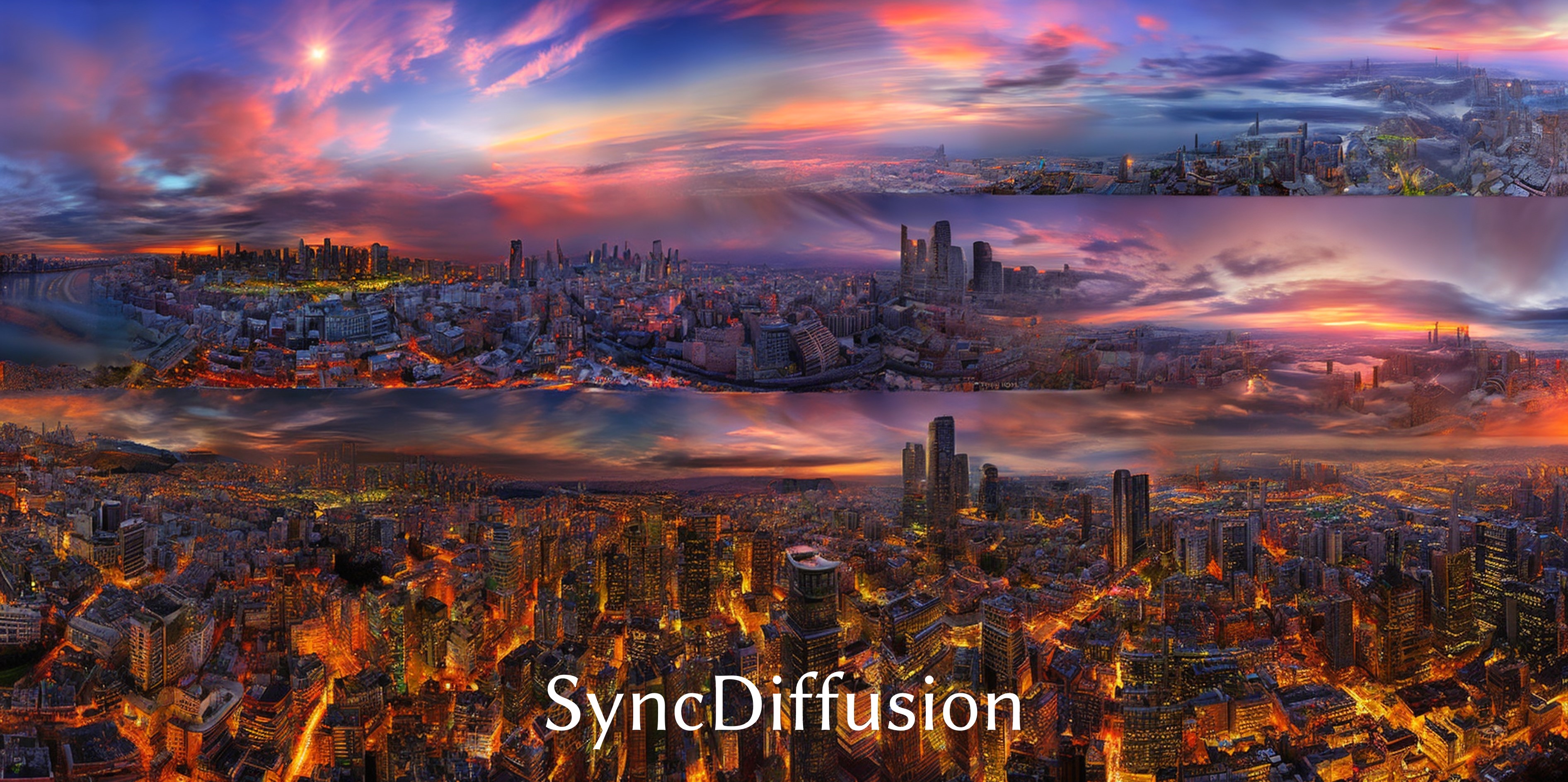}
        \caption{Sky comparison}\label{fig:method_sky_pano}
    \end{subfigure}  
    \caption{(a) High-resolution sky synthesis with omnidirectional sampling. (b) Panoramic sky synthesis comparisons with SyncDiffusion \cite{lee2023syncdiffusion} using same seeds and text prompts.}
\end{figure}

Given the same style image and text references, we adopt the pre-trained Stable Diffusion model \cite{rombach2021high} and conduct vision-and-text aligned sky panorama synthesis. 
A style-aligned sky, serving as the unique background of urban scenes, is essential to establish a global atmosphere for the final rendering. Moreover, the sky background supports comprehensive visual semantics, enhancing the effectiveness of textual style loss (Eq.~\ref{eq:clip_loss}). The pre-trained diffusion model excels in generating small image patches (e.g., $512 \times 512$), while diffusion panorama synthesis methods \cite{bar2023multidiffusion, lee2023syncdiffusion} utilize multi-window joint diffusion technique to generate high-resolution panorama. 
However, to support rendering with an environment texture, we require a high-resolution sky image covering 360-degree FoV. Due to uniform diffusion among sliding windows, the panorama is generally flattened and suffers layered content when diffusing latent features vertically, such as the SyncDiffusion \cite{lee2023syncdiffusion} results demonstrated in Fig. \ref{fig:method_sky_pano}.
To achieve high-resolution panorama synthesis as a whole entity, we extend the multi-window joint diffusion technique and propose omnidirectional latent feature sampling for view supervision. At each denoising step, we leverage a 90-degree bounding field of view (BFoV)~\cite{huang2023360vot} and omnidirectionally sample the latent features to extract noisy latent feature sub-regions. Each sub-region latent feature is then denoised by the denoising diffusion implicit model (DDIM)~\cite{song2020denoising} and decoded by the variational autoencoder (VAE) decoder to generate a color image. %local latent features 
Meanwhile, we apply the perceptual similarity constraint $\mathcal{L}_{LPIPS}$ \cite{zhang2018perceptual} between the input style reference and generated images to supervise similar visual style. Fig. \ref{fig:sky_gen} illustrates the sky synthesis process; Fig. \ref{fig:method_sky_pano} and \ref{fig:exp_more_results} display comparison and synthesis examples. Please refer to the supplementary for more details and comparisons. 

\section{Experiments}
We conducted experiments on UV textured models in different cities collected by Google 3D Tiles API \cite{google2023tileapi}. For evaluation, we rendered a sequence of testing views around each model.  
For 2D style images, we referred to city-view evaluation images in \cite{chen2022timeofday,luan2017deep}, and applied the text-guided synthesis model \cite{zhang2023adding} to generate some style images given customized style text prompts.
Source texts follow templates such as "a photo in the day", and non-customized target texts are generated by \cite{clipinterrogator2023} given style images. 
All experiments were conducted on RTX 3090 GPU. We implemented our method based on Pytorch and Pytorch3D \cite{ravi2020pytorch3d}, and used tiny-cuda-nn \cite{tiny-cuda-nn} for multi-resolution grid-based hash encoding and network training acceleration. Resolution $512 \times 512$ was used for optimization.

\subsubsection{Result Comparisons.}
Given the downgraded performance of image or video stylization in 3D scenarios regarding 3D geometric and appearance consistency \cite{chiang2022stylizing, huang2021learning, huang2022stylizednerf}, our focus lies on 3D stylization baselines.
We compared our method with the image-guided explicit texture stylization StyleMesh \cite{hollein2022stylemesh}, image- and text-guided NeRF stylization methods ARF \cite{zhang2022arf} and InstructN2N \cite{haque2023instructnerf}.
We ran baselines using official source codes and increased the content loss hyperparameter used in StyleMesh and ARF by 10 times for photorealism. We took the same pivot views (\cref{sec:method_view_plan}) as their training input. 
We evaluated the performance of foreground content similarity by edge-SSIM (eSSIM) and masked LPIPS \cite{zhang2018perceptual} between source and stylized testing views, and semantic style similarity by image-text CLIP score \cite{hessel2021clipscore} between the text reference and stylized testing views with the same sky background. Moreover, we conducted a user study on content preservation and style matching of stylized models. 
Table \ref{tab:exp_quan} and Fig. \ref{fig:exp_qual_results} demonstrate quantitative and qualitative comparisons. 

\begin{table*}[t]
    \centering
    \footnotesize
    \setlength{\tabcolsep}{3pt}
    \renewcommand{\arraystretch}{1.}
    \caption{Quantitative comparisons. 
    We evaluated 5 city models and 7 styles, resulting in 35 stylized outputs for each method. These city models, spanning approximately 0.5km $\times$ 0.5km in physical area, represent diverse locations worldwide, including London, Sydney, New York, Hong Kong, and Singapore.}
    \resizebox{\linewidth}{!}{ 
    \begin{tabular}{l|c c c|c c H c c c}
    \toprule
     \multirow{2}{*}{Method} &\multirow{2}{*}{3D Representation}& \multirow{2}{*}{Condition}& \multirow{2}{*}{Scalability} & \multirow{2}{*}{eSSIM$\uparrow$  } & \multirow{2}{*}{LPIPS$\downarrow$} & & \multirow{2}{*}{CLIP$\uparrow$} & \multicolumn{2}{c}{User Study (score in 0-5)}\\
     \cline{9-10}
      &  & &&& &&& Content Preserve$\uparrow$ & Style Match$\uparrow$\\
    \midrule
     ARF~\cite{zhang2022arf}& NeRF &image & object-centric &0.394 & 0.812 & 14.693 & 0.188 & 1.576& 1.430 \\
    StyleMesh~\cite{hollein2022stylemesh}&explicit texture+mesh & image &room-level &0.400 & 0.576 & 2.481 &0.188 &  2.074& 2.402 \\
     % 2D-to-3D & 0.623& \textbf{0.339}& 2.883& 0.193& & \\
     InstructN2N~\cite{haque2023instructnerf} &NeRF &text & object-centric &0.503& 0.370& 3.756& 0.186& 3.156& 2.024 \\
     Ours &neural texture+mesh & image+text & urban-level&\textbf{0.623}& \textbf{0.342}& 5.031 & \textbf{0.244} & \textbf{4.586} & \textbf{4.572} \\
    \bottomrule
    \end{tabular}
    }
    \label{tab:exp_quan}
\end{table*}

\begin{table*}[t!]
    \centering
    \def\wsix{0.14}
    \def\wsixinput{0.16}
    \def\wfour{0.23}
    \renewcommand{\arraystretch}{0.5}
    \resizebox{0.95\linewidth}{!}{ 
    \begin{tabular}{lcccc}
         \rotatebox{90}{ \parbox{0.09\linewidth}{\centering Input}} &
         \multicolumn{2}{c}{         \includegraphics[width=\wsixinput\linewidth]{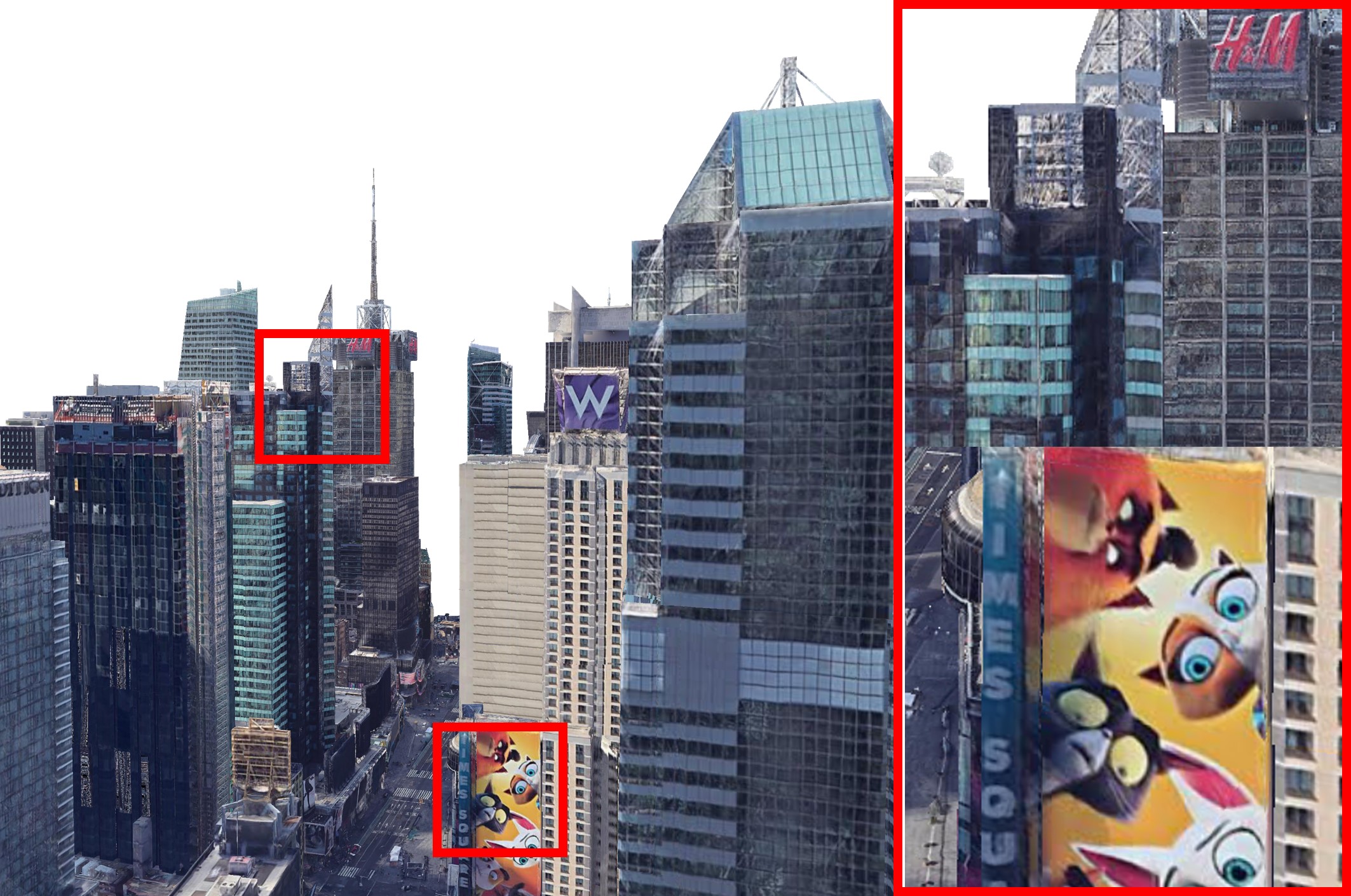}         \includegraphics[width=\wsixinput\linewidth]{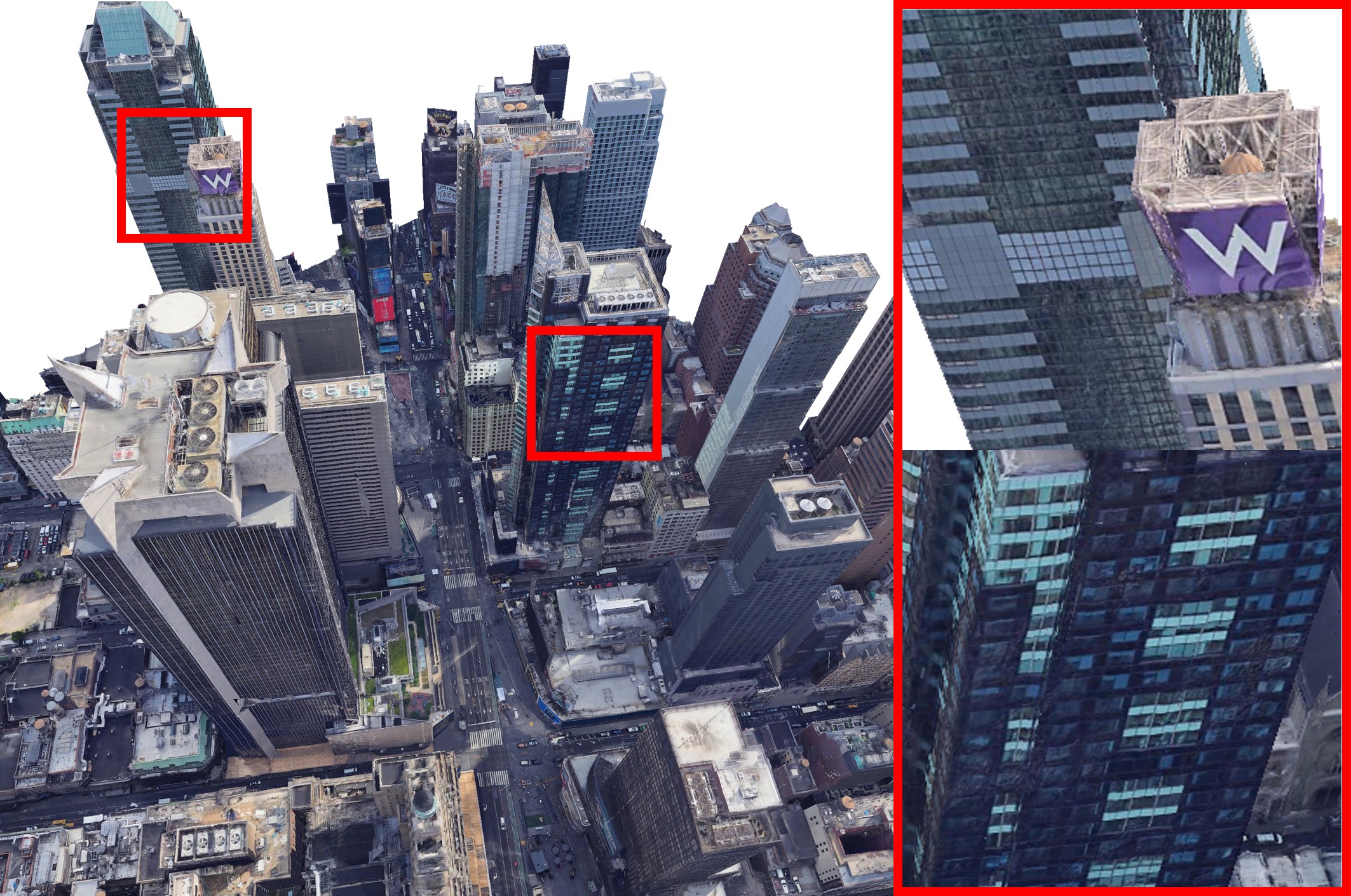}
         \includegraphics[width=\wsix\linewidth]{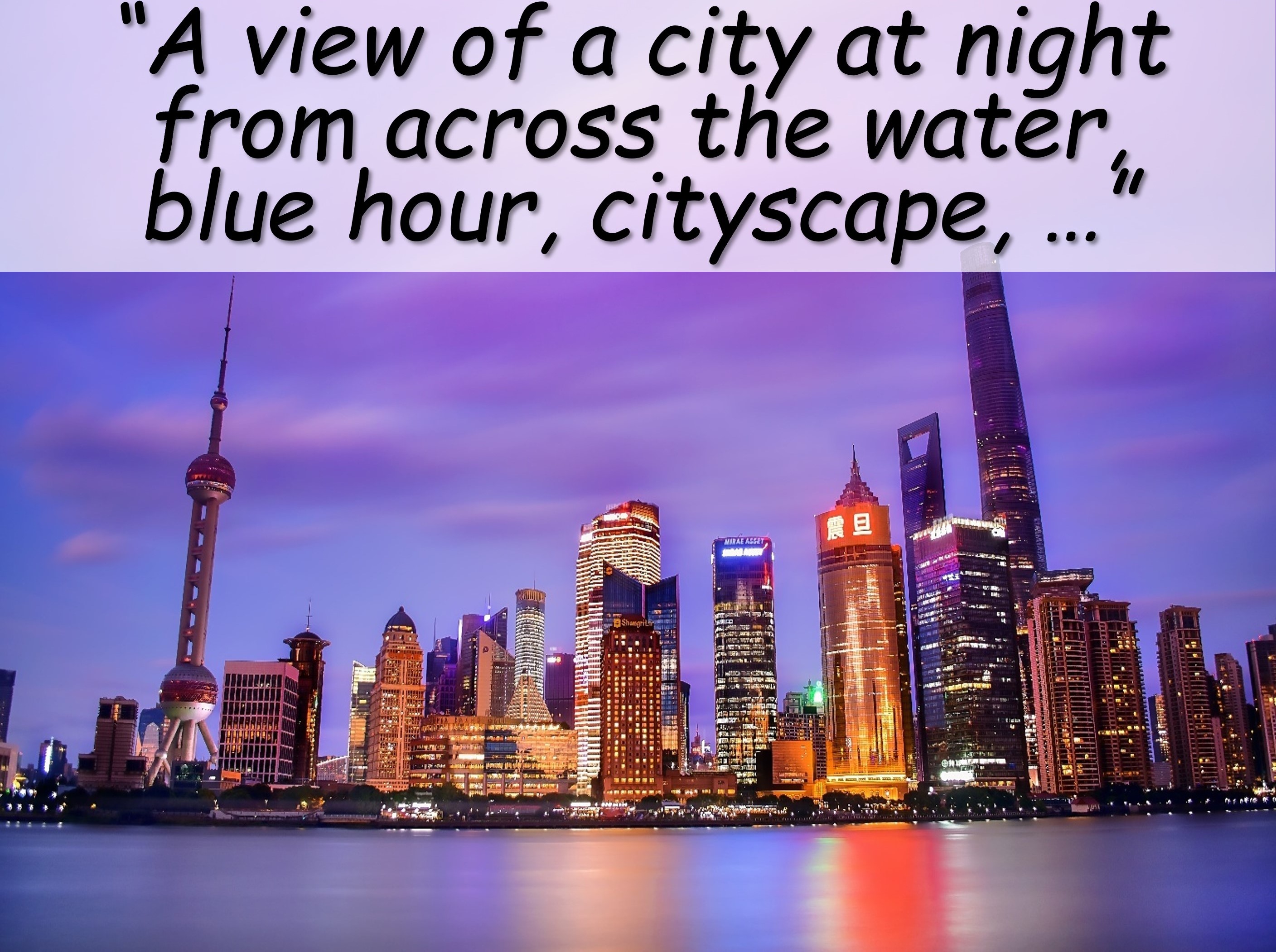}}  & 
         \multicolumn{2}{c}{\includegraphics[width=\wsixinput\linewidth]{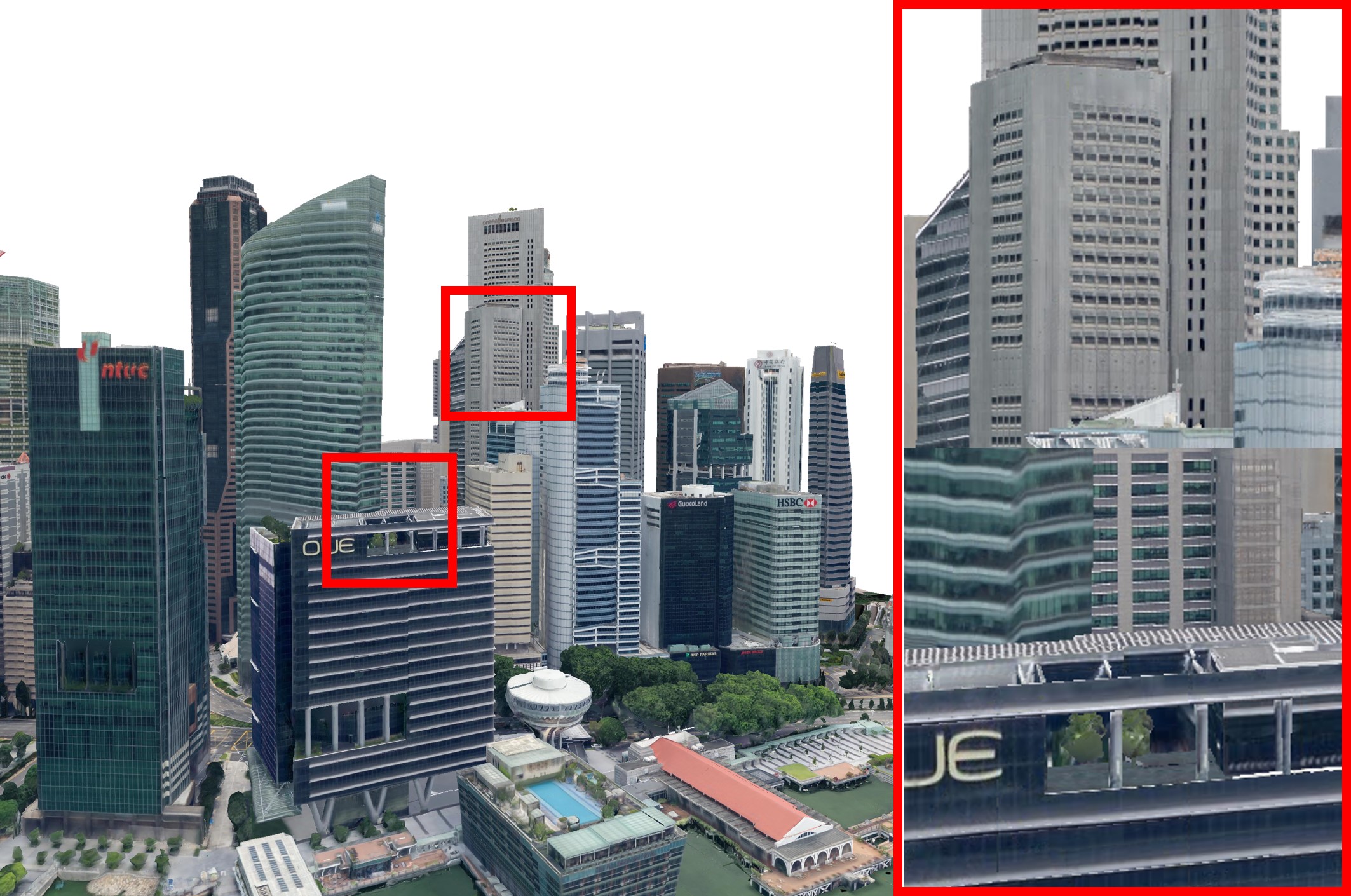}         \includegraphics[width=\wsixinput\linewidth]{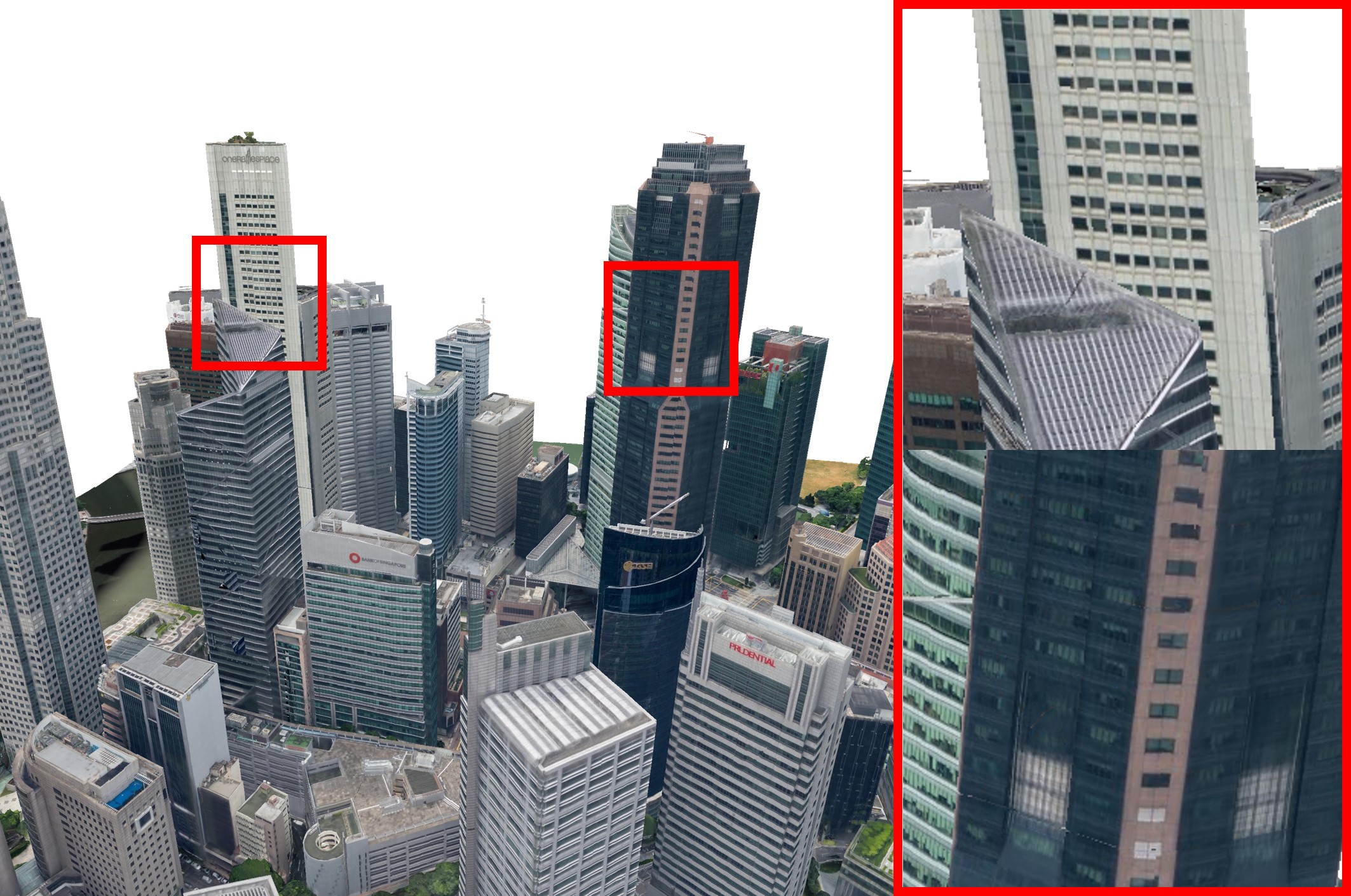}         \includegraphics[width=\wsix\linewidth]{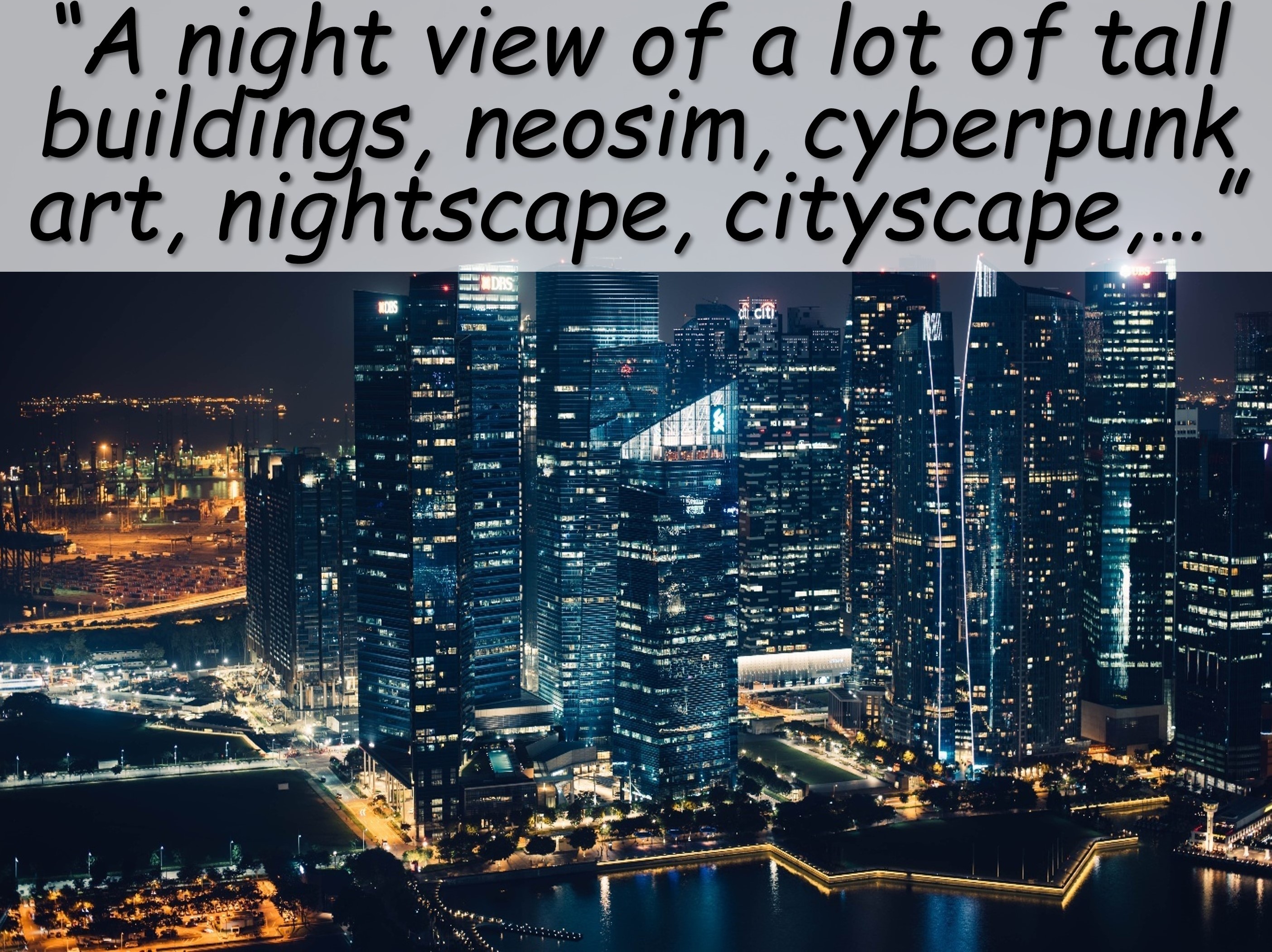}}\\         
         \toprule
         \rotatebox{90}{ \parbox{0.12\linewidth}{\centering \textbf{Ours}}} & 
         \multicolumn{2}{c}{\includegraphics[width=\wfour\linewidth]{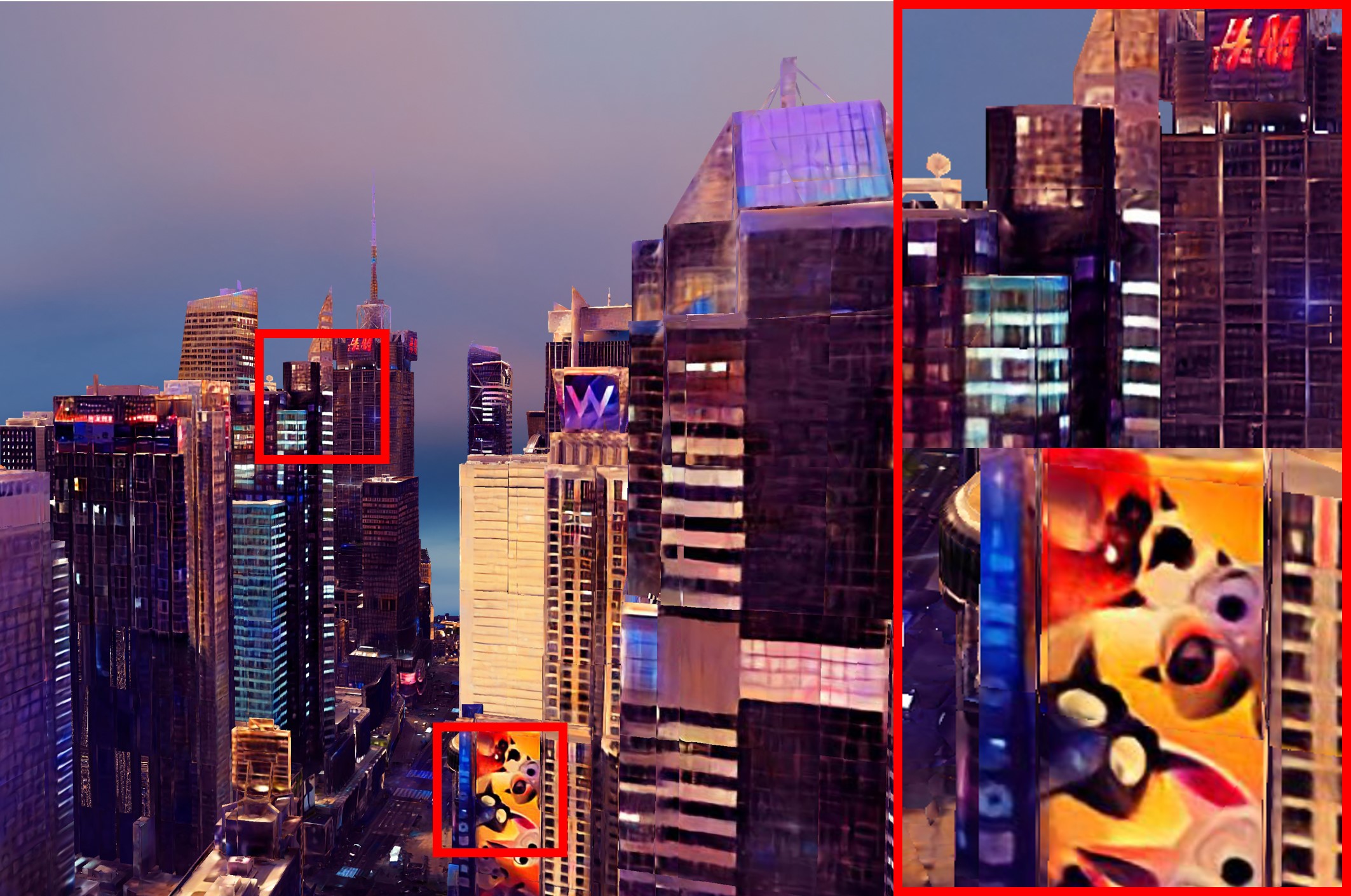}         \includegraphics[width=\wfour\linewidth]{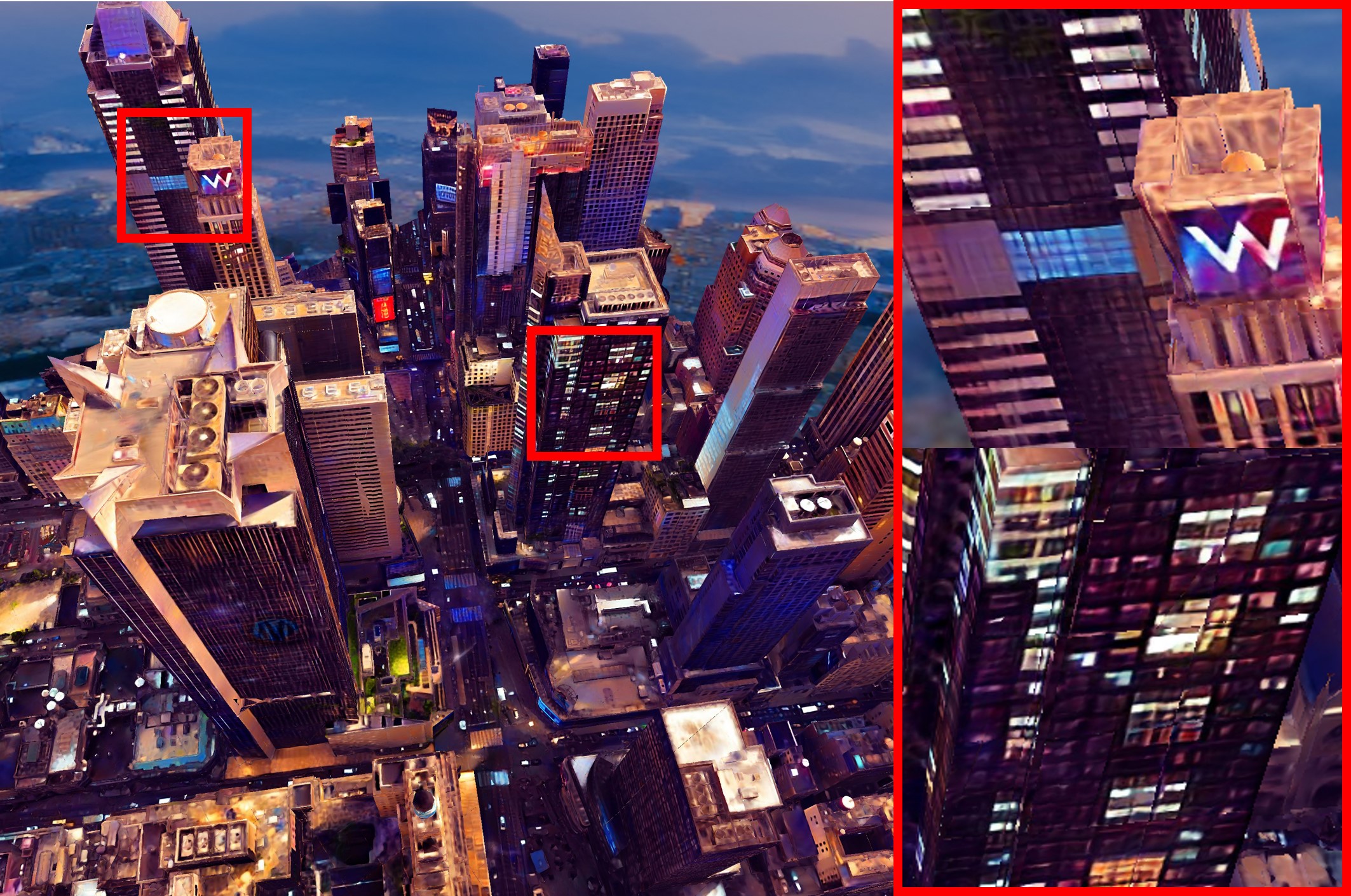}} & 
         \multicolumn{2}{c}{\includegraphics[width=\wfour\linewidth]{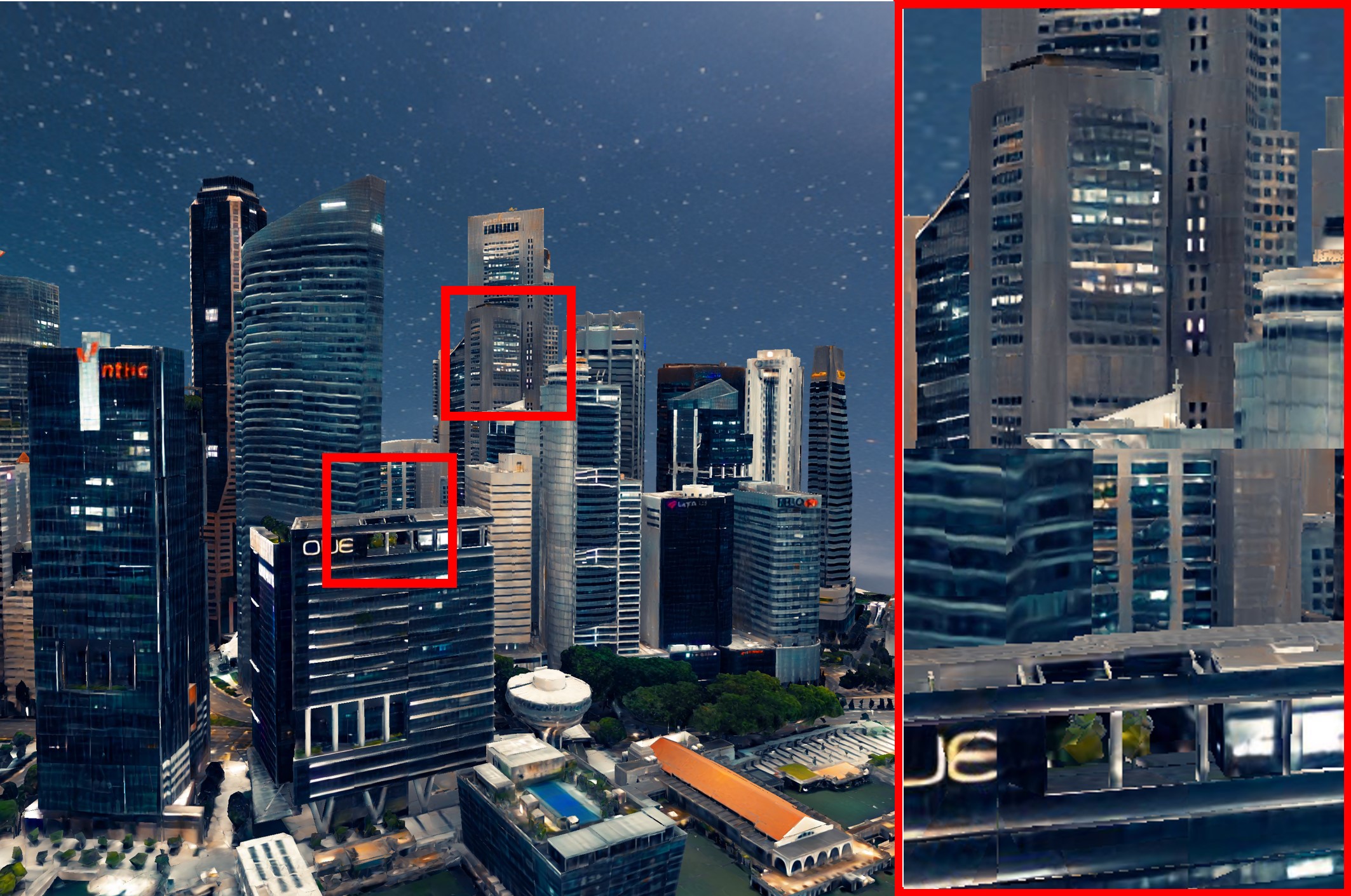}         \includegraphics[width=\wfour\linewidth]{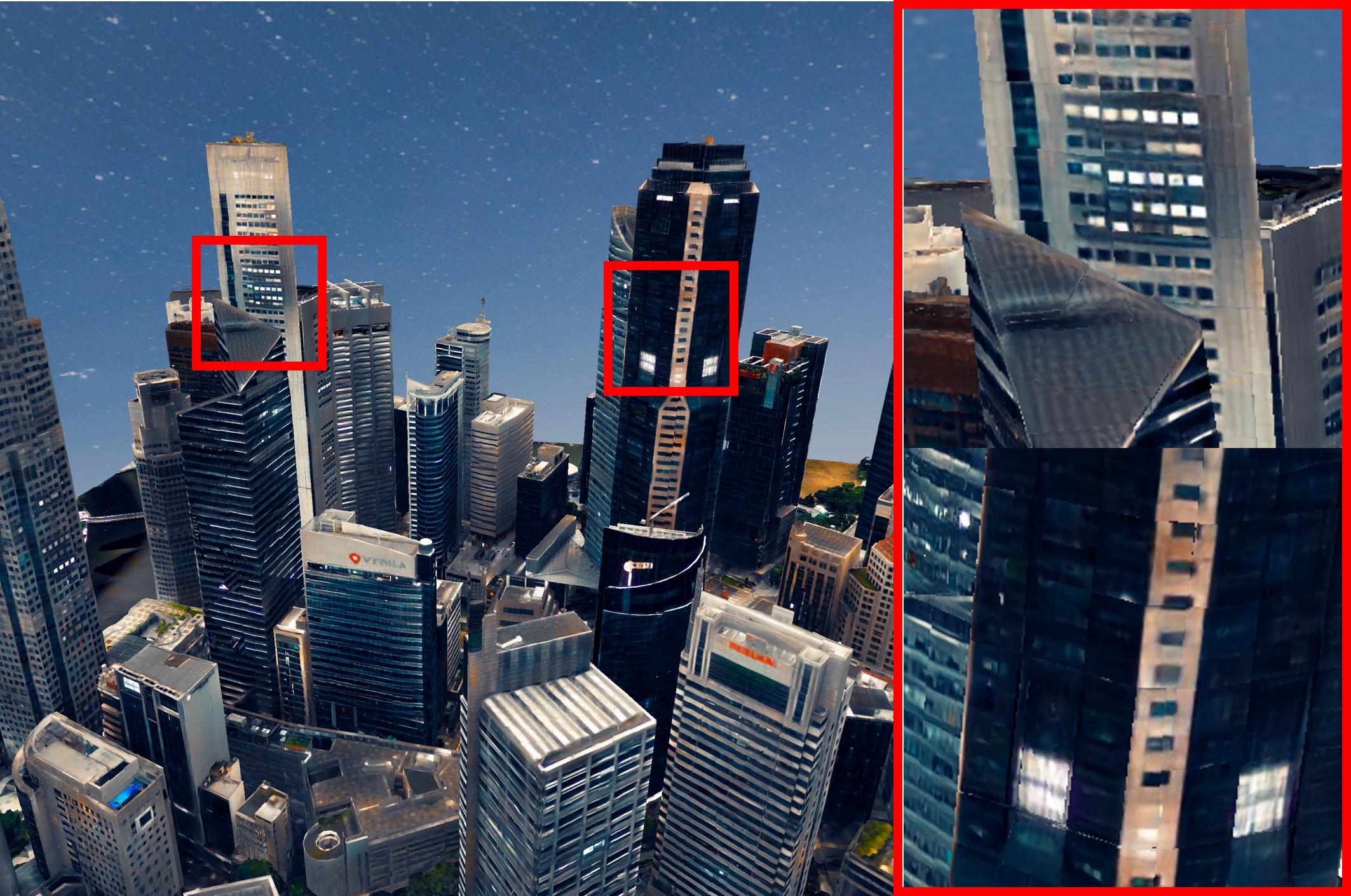}} \\       
        \rotatebox{90}{ \parbox{0.13\linewidth}{\centering StyleMesh}} & 
         \multicolumn{2}{c}{\includegraphics[width=\wfour\linewidth]{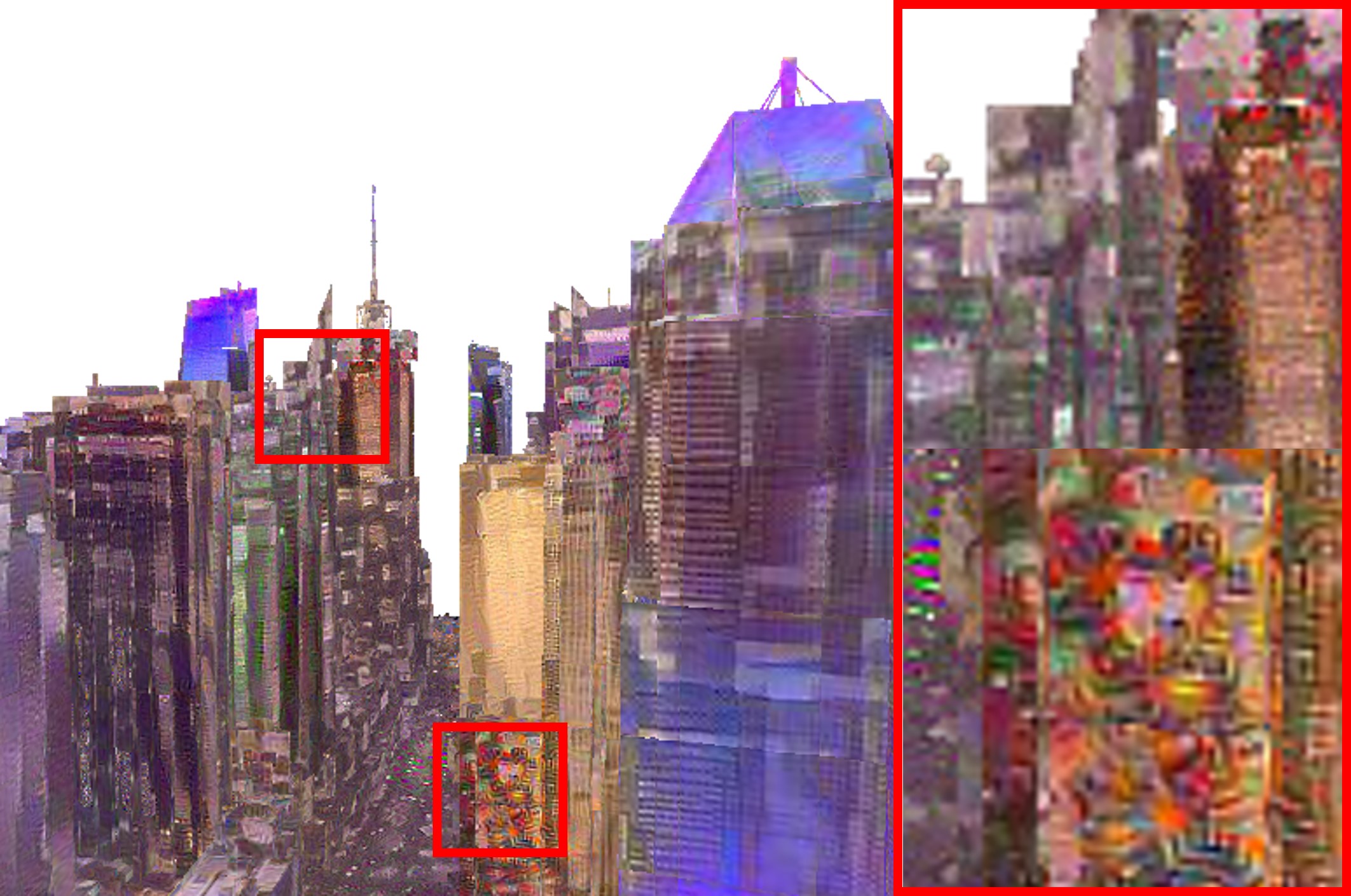}         \includegraphics[width=\wfour\linewidth]{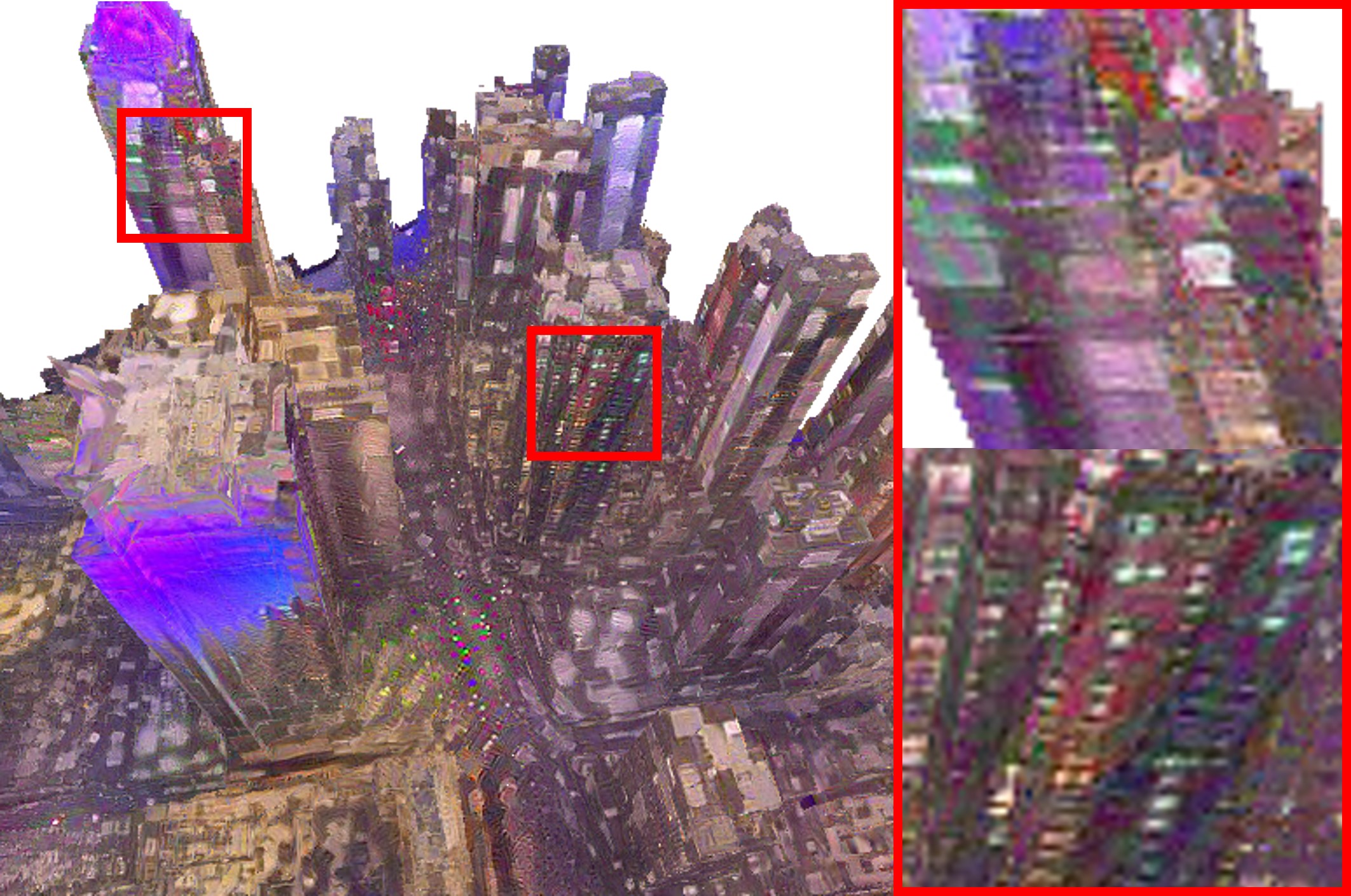}} & 
         \multicolumn{2}{c}{\includegraphics[width=\wfour\linewidth]{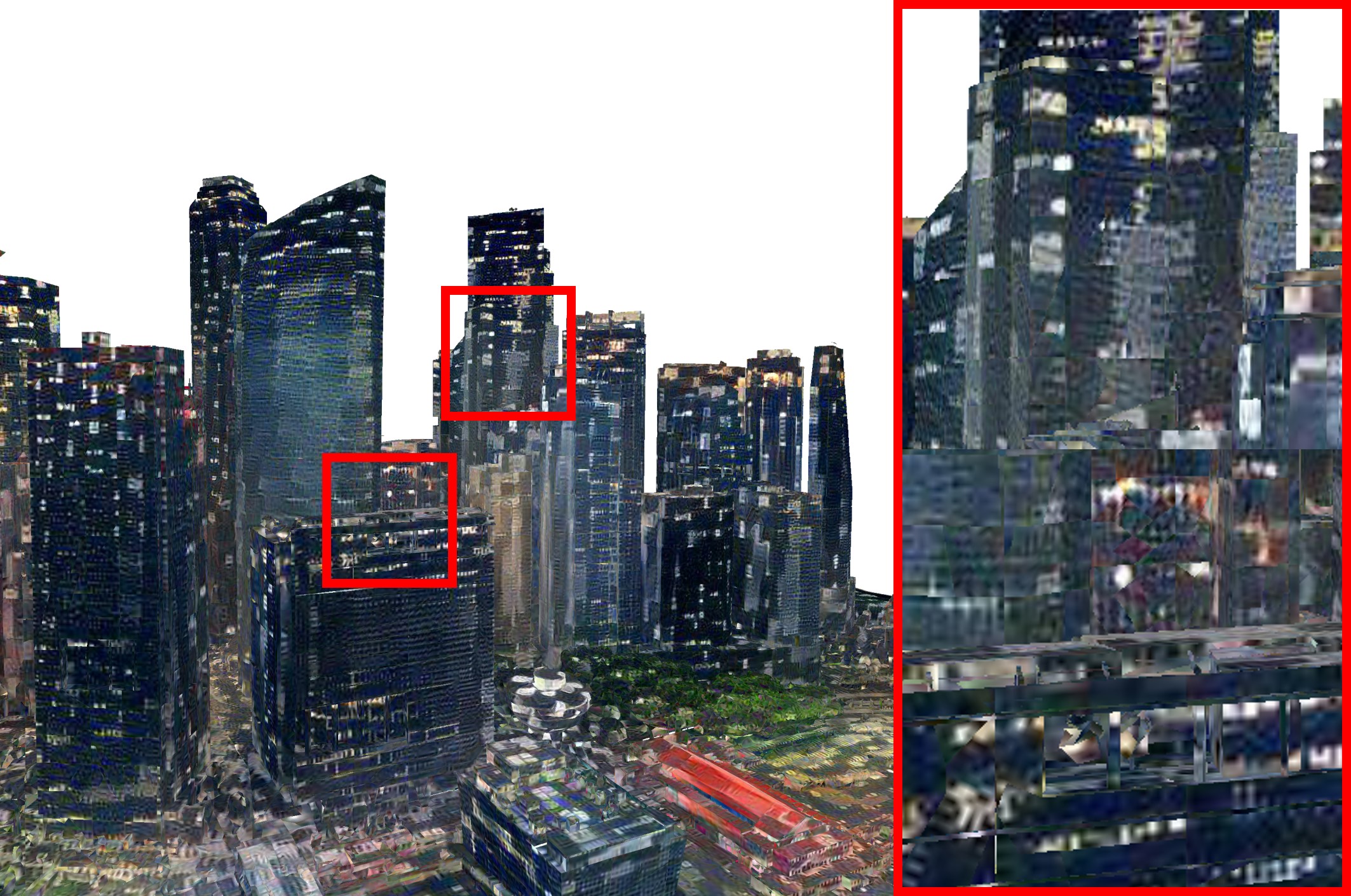}         \includegraphics[width=\wfour\linewidth]{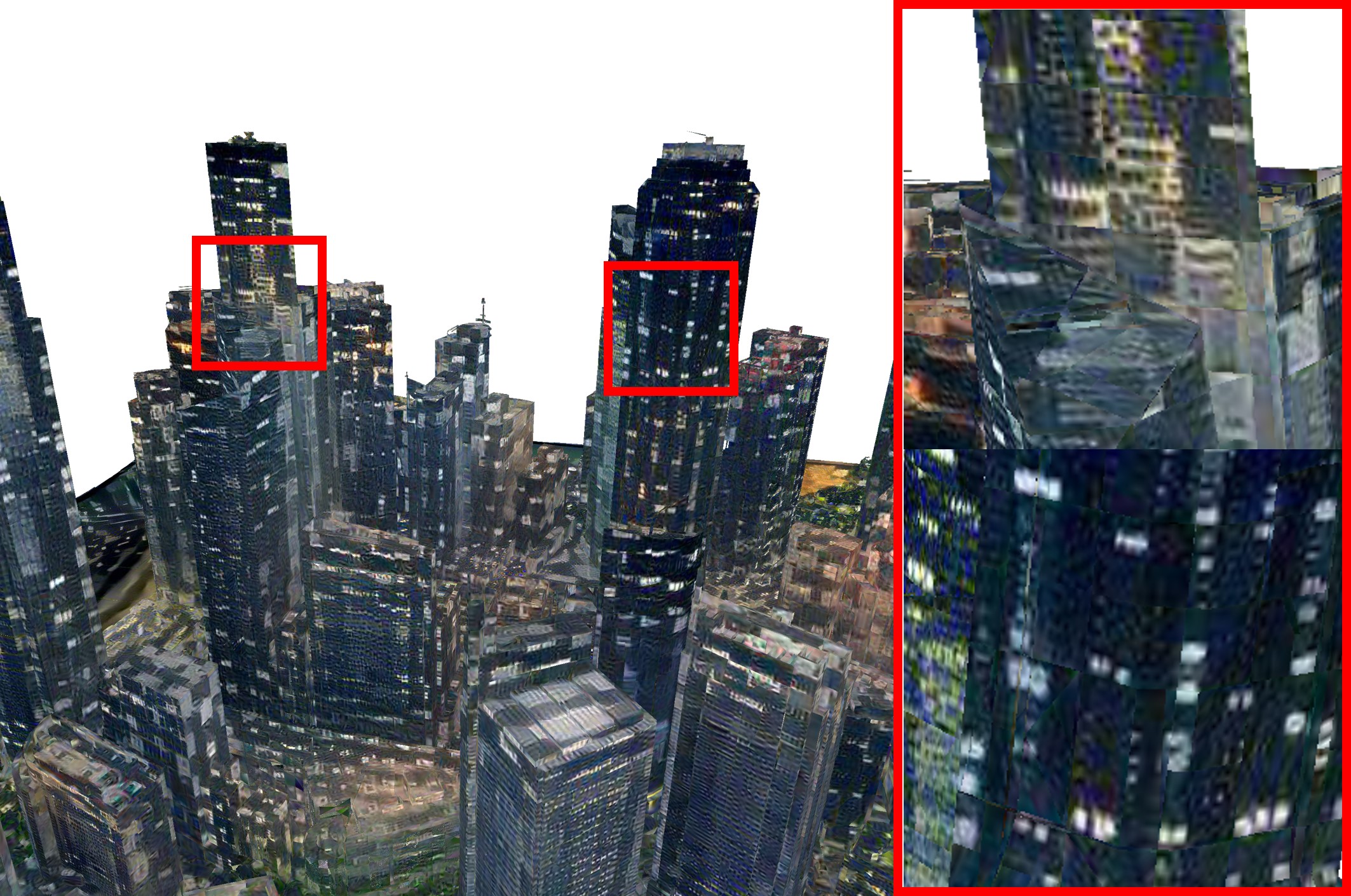}} \\
         
        \rotatebox{90}{ \parbox{0.12\linewidth}{\centering ARF}} & 
         \multicolumn{2}{c}{\includegraphics[width=\wfour\linewidth]{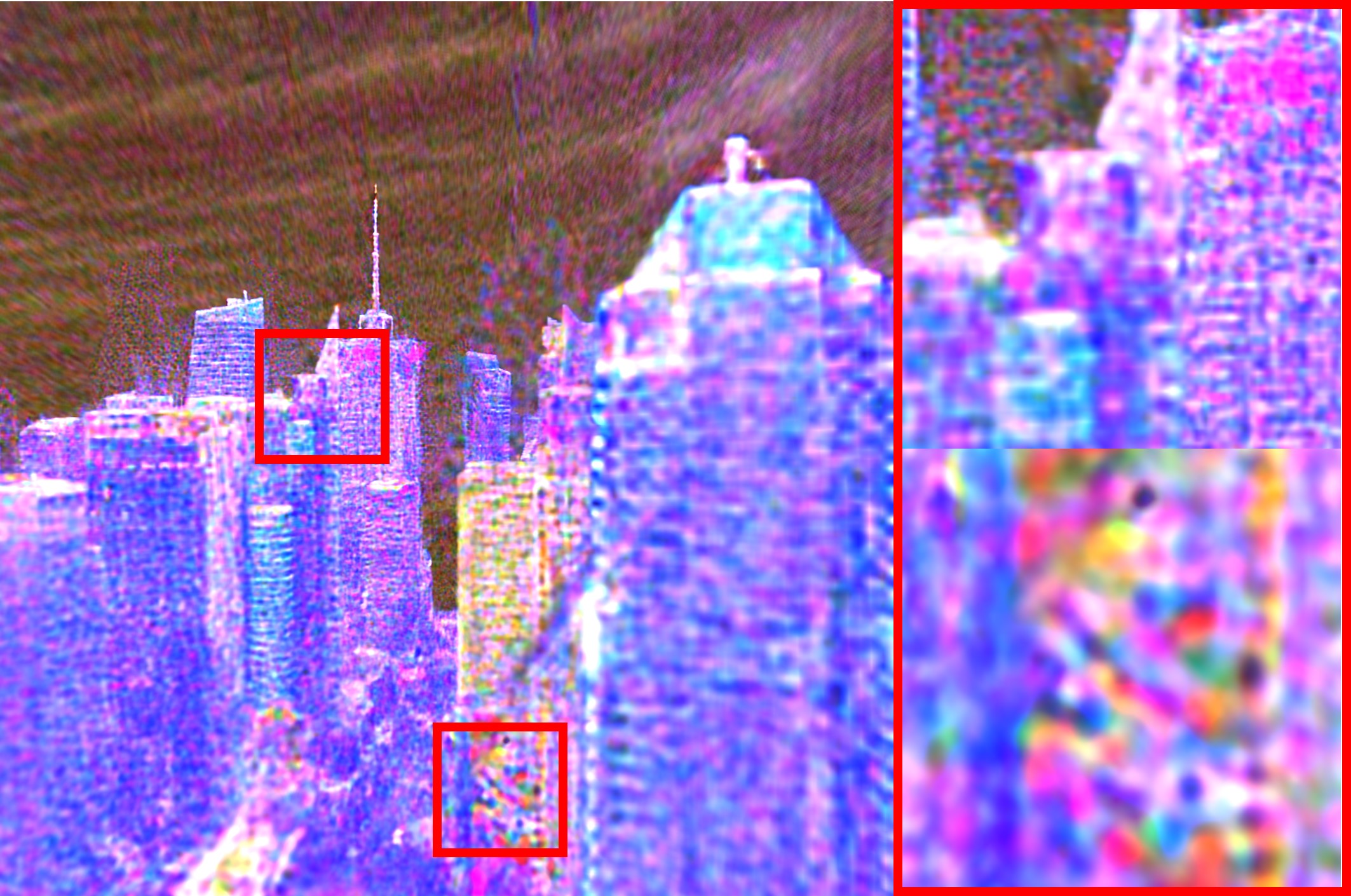}         \includegraphics[width=\wfour\linewidth]{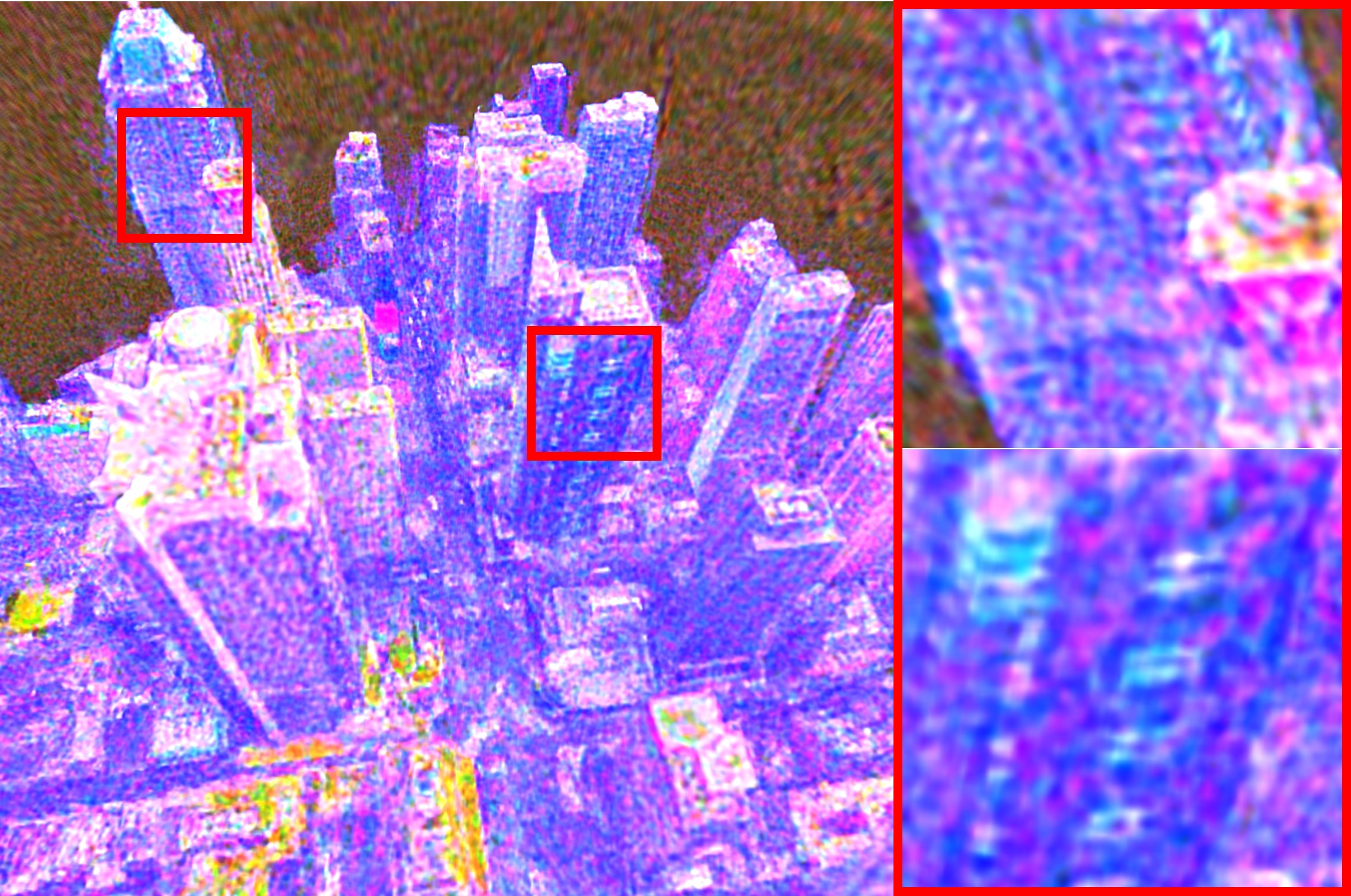}} & 
         \multicolumn{2}{c}{\includegraphics[width=\wfour\linewidth]{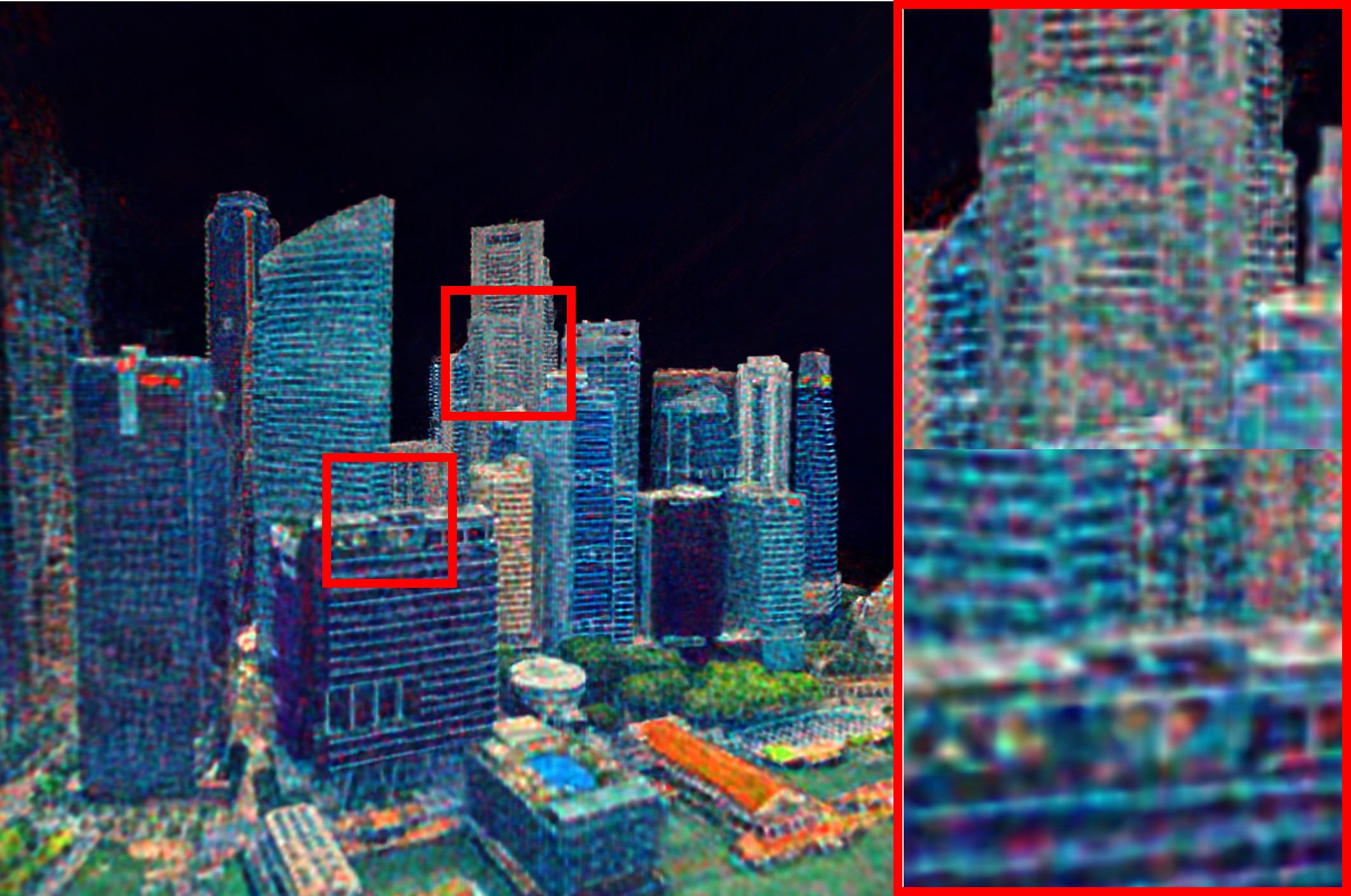}         \includegraphics[width=\wfour\linewidth]{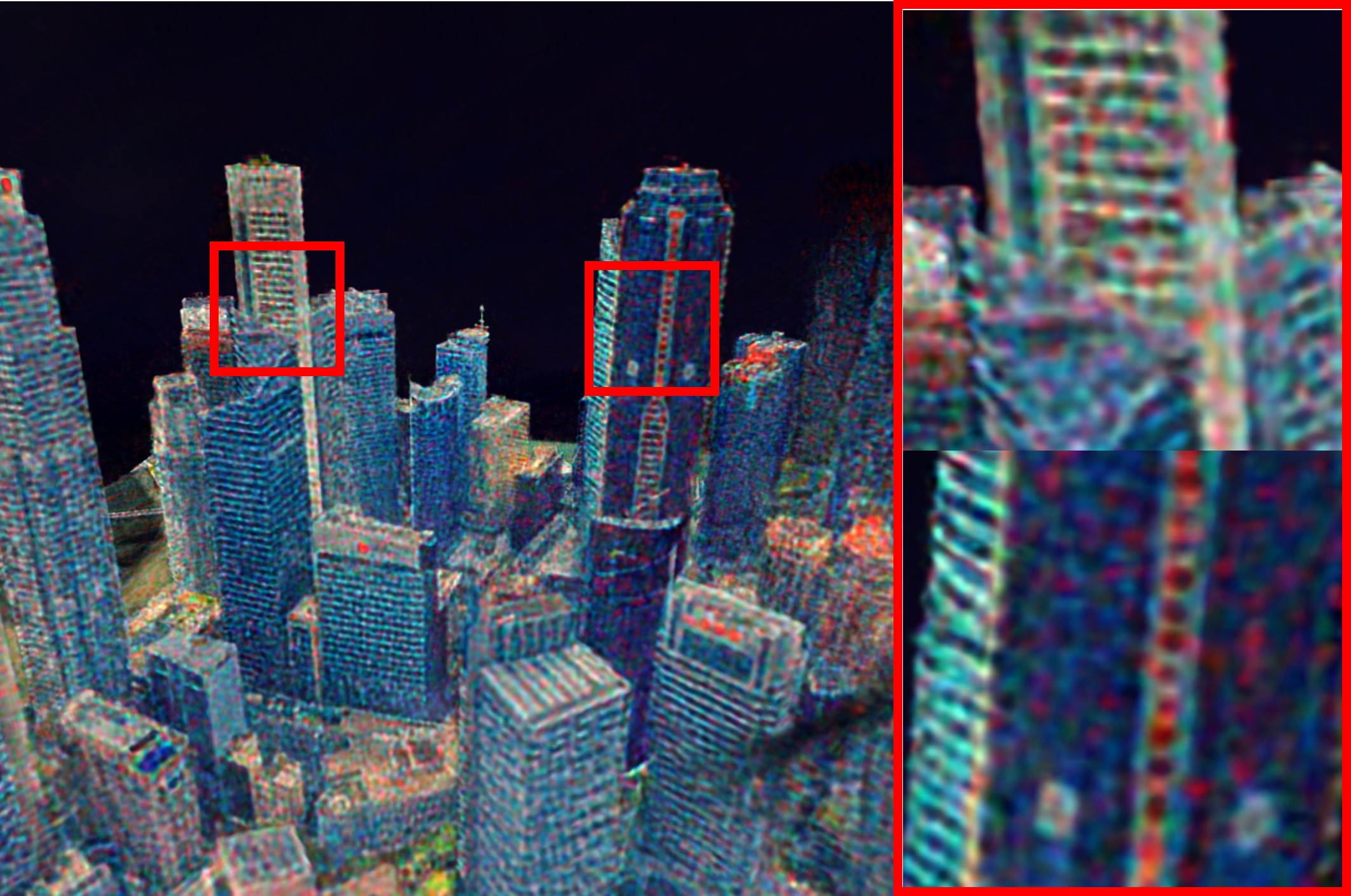}} \\
        \rotatebox{90}{ \parbox{0.13\linewidth}{\centering InstructN2N}} & 
         \multicolumn{2}{c}{\includegraphics[width=\wfour\linewidth]{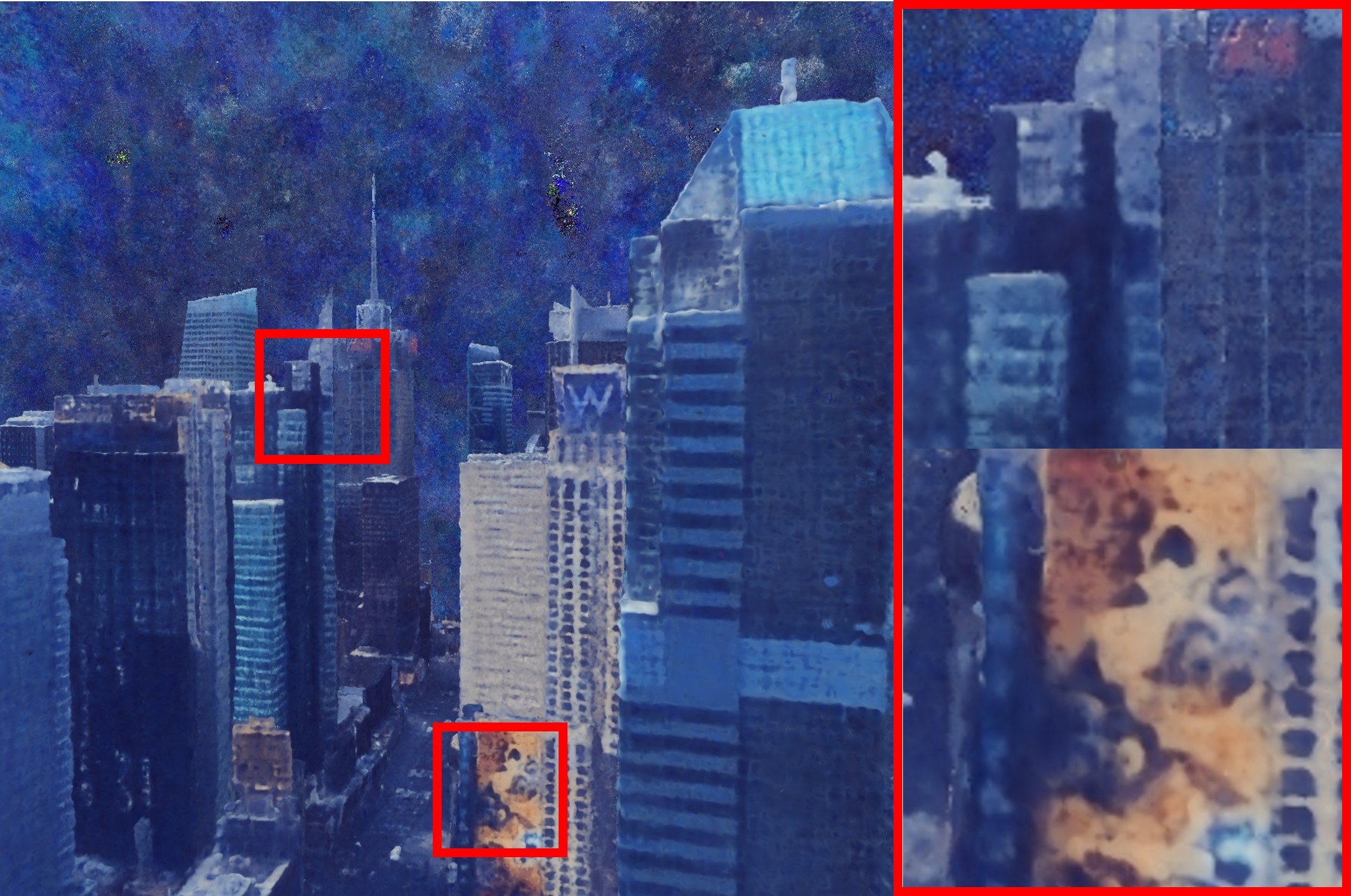}
         \includegraphics[width=\wfour\linewidth]{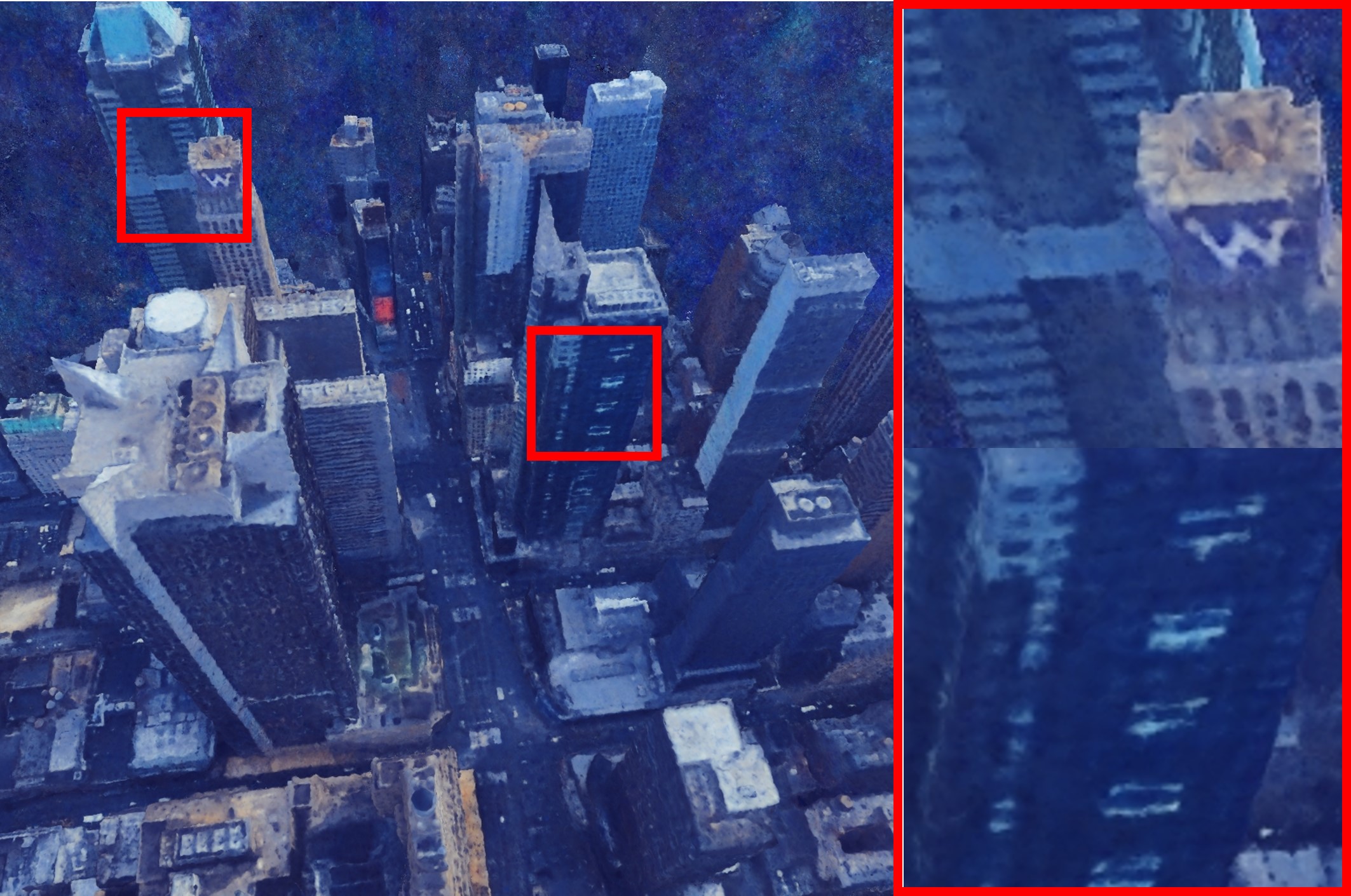}} & 
         \multicolumn{2}{c}{\includegraphics[width=\wfour\linewidth]{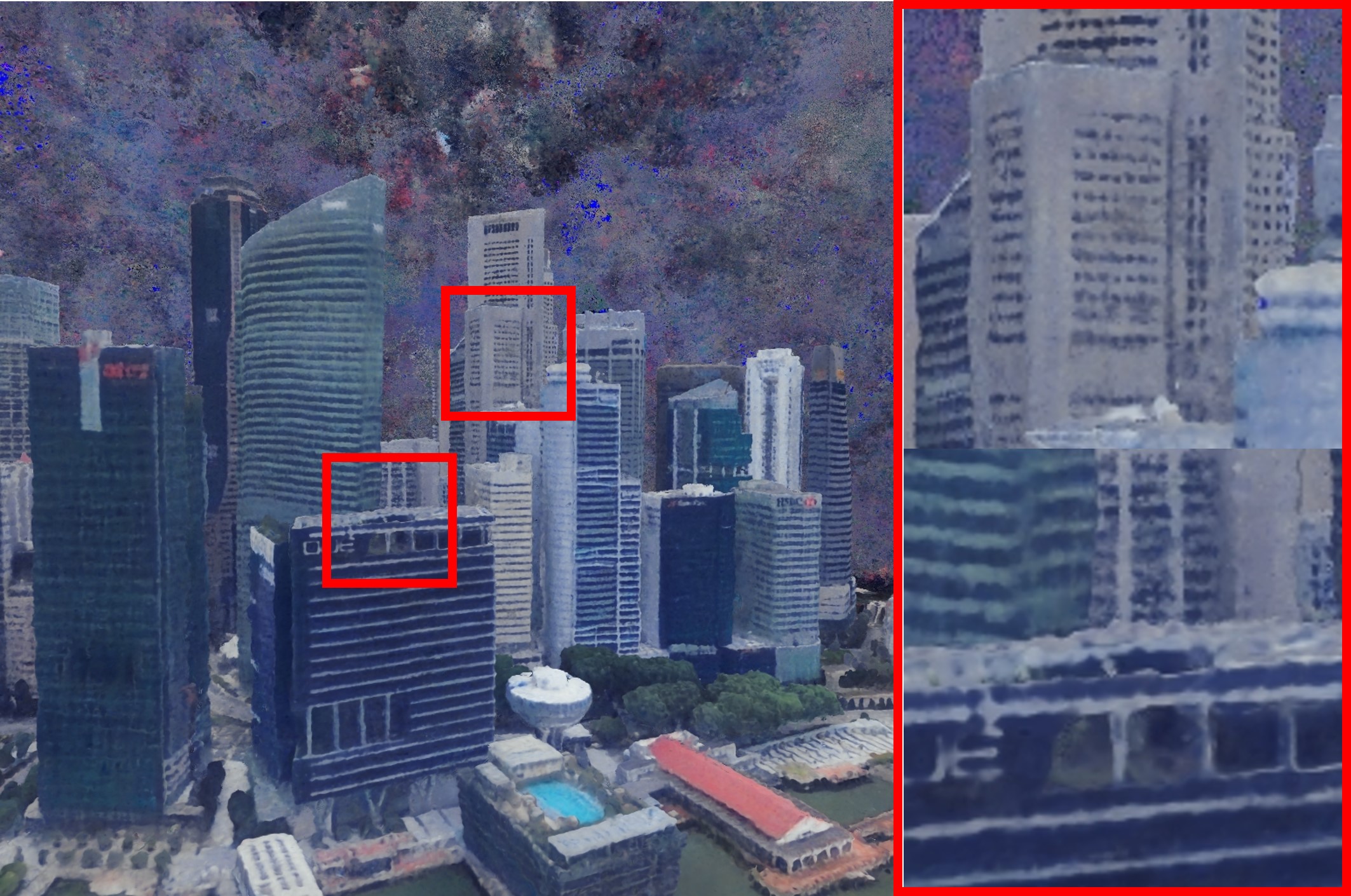}
         \includegraphics[width=\wfour\linewidth]{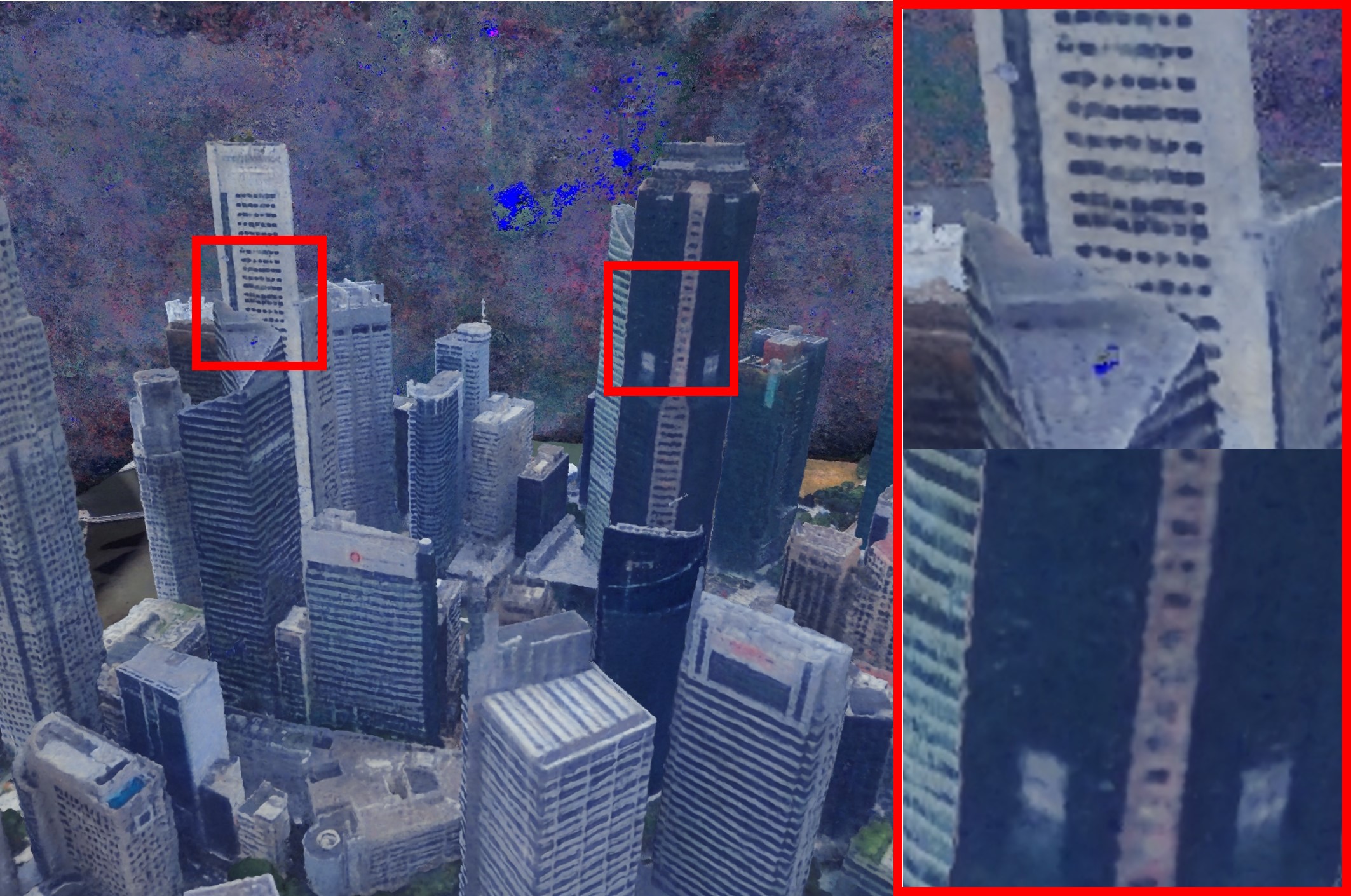}} \\
    \end{tabular}}
    \captionof{figure}{Visual comparisons with baselines. Zoom in for details. }
    \label{fig:exp_qual_results}
\end{table*}
By comparison, our results achieved higher fidelity of motifs and style details such as logos, and semantically matched styles like lit windows, while other baselines lead to blurriness or non-photorealism. The 2D generative diffusion-based 3D lifting method InstructN2N tends to have minor global color changes although preserves more original content features. We also found it difficult to learn abstract-style prompts. For example, it always coats the scene with blue for "blue hour" prompts even along with other style-related descriptions.
Optimization-based artistic style transfer methods (StyleMesh and ARF) lead to dramatic content distortion without photorealism concerns. From visual results, hazy artifacts appear in all scenes in volume rendering based neural field stylization (ARF and InstructN2N). From the user study's feedback, we found ARF got the lowest "style match" score because of its blurry appearance. 
Besides, in the absence of semantic matching, image-guided methods tend to transfer mismatched styles, such as ARF stylizing buildings with the purple sky and background with building colors in Fig.~\ref{fig:exp_qual_results}.

\begin{figure*}[t]
    \centering  
    \def\fabwidth{0.36}
    \def\abwidth{0.2}
    \begin{subfigure}[b]{\fabwidth\linewidth}
        \centering
        \includegraphics[width=\textwidth]{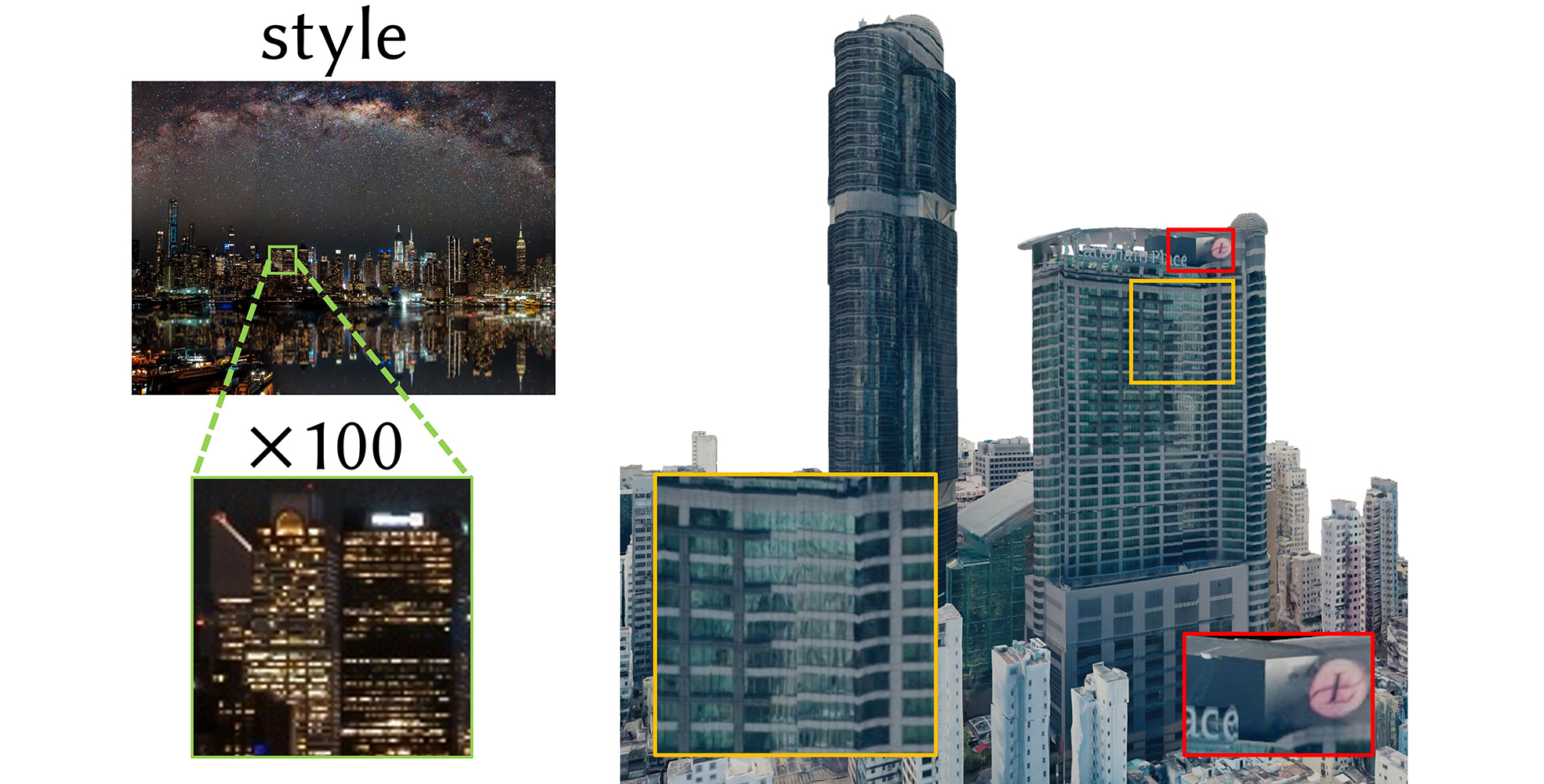} 
        \caption{Input}
    \end{subfigure}    
    \begin{subfigure}[b]{\abwidth\linewidth}
        \centering
        \includegraphics[width=0.9\textwidth]{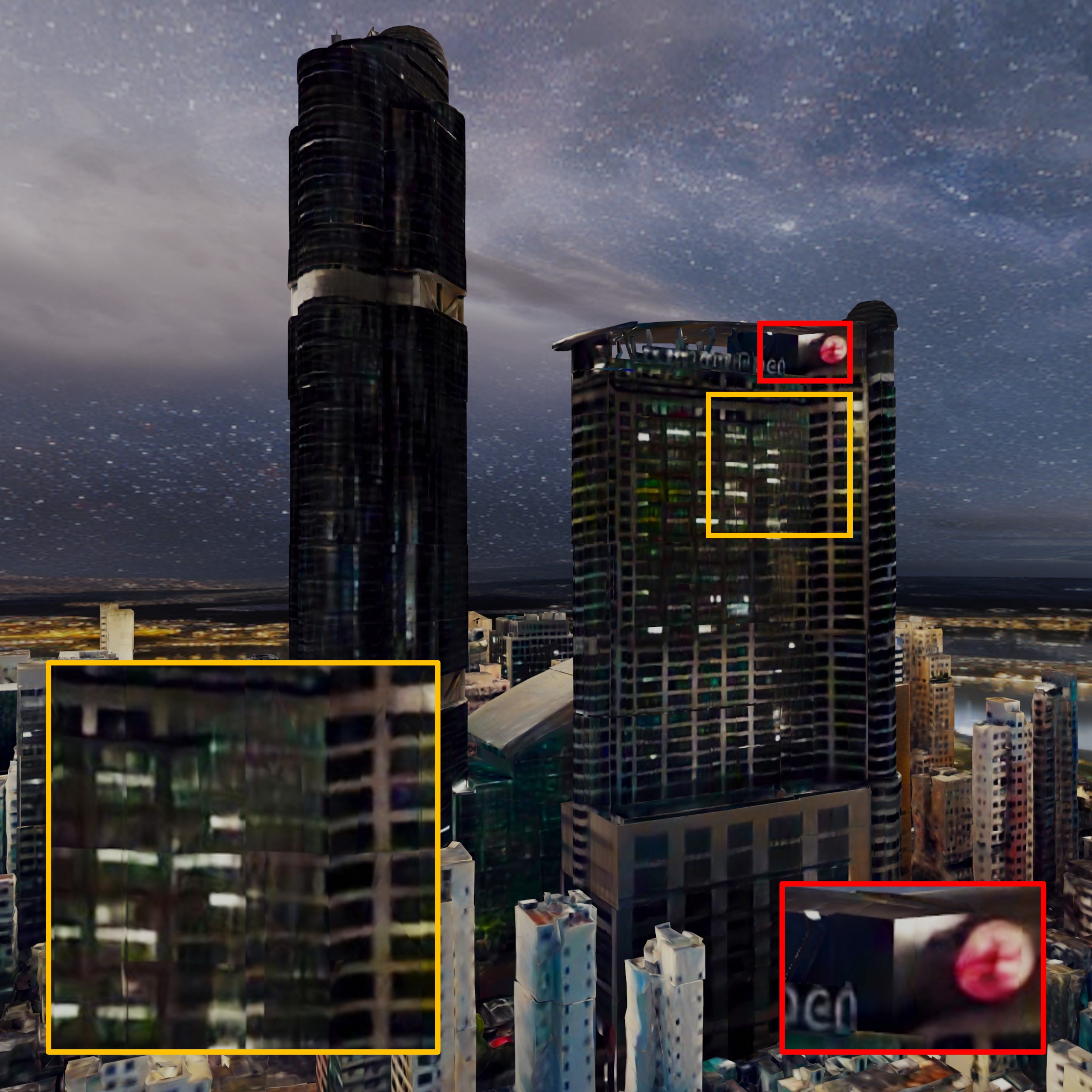}
        \caption{-aug}
        \label{fig:exp_ab_aug}
    \end{subfigure}
    \begin{subfigure}[b]{\abwidth\linewidth}
        \centering
        \includegraphics[width=0.9\textwidth]{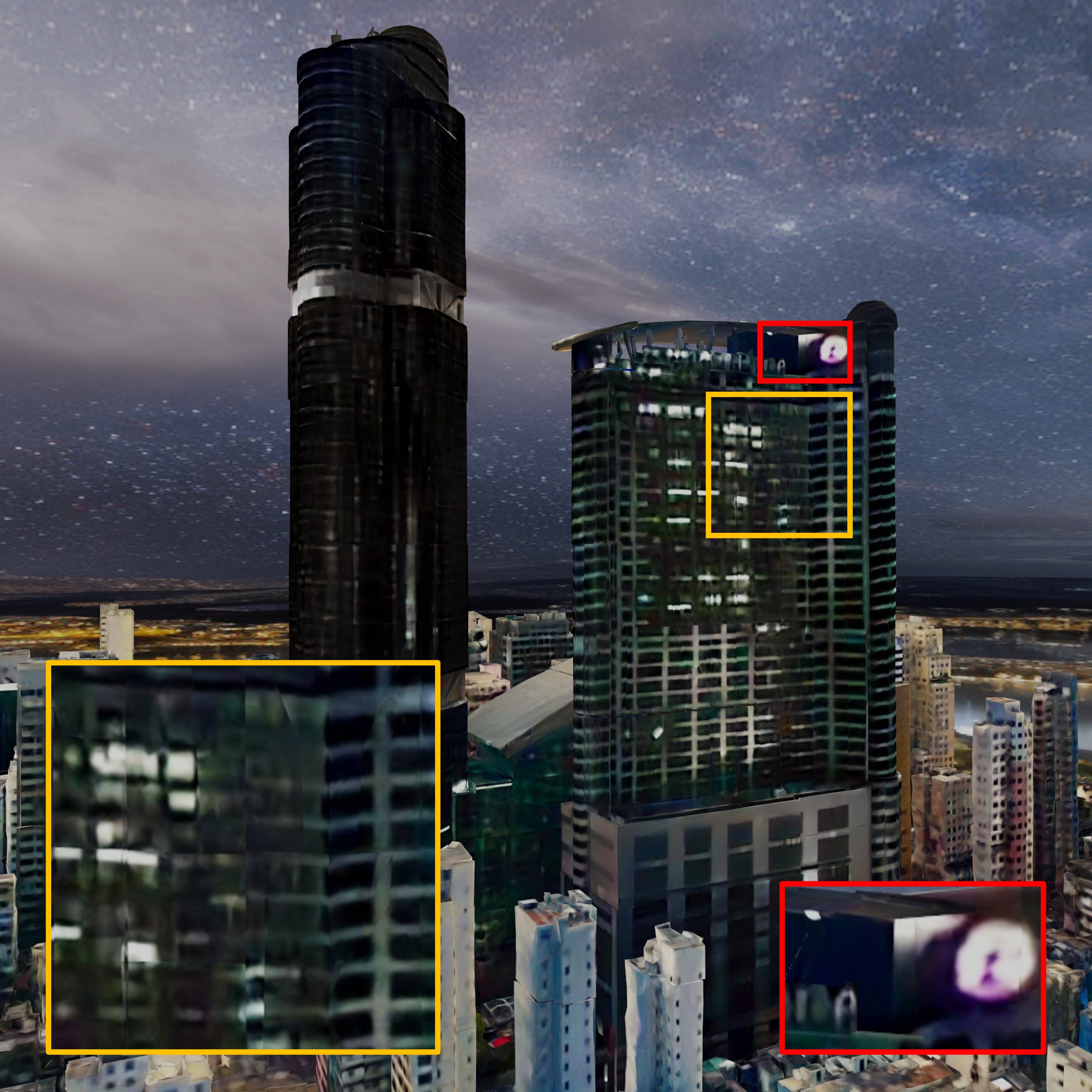}
        \caption{-prog} \label{fig:exp_ab_prog}
    \end{subfigure}    
    \begin{subfigure}[b]{\abwidth\linewidth}
        \centering
        \includegraphics[width=0.9\textwidth]{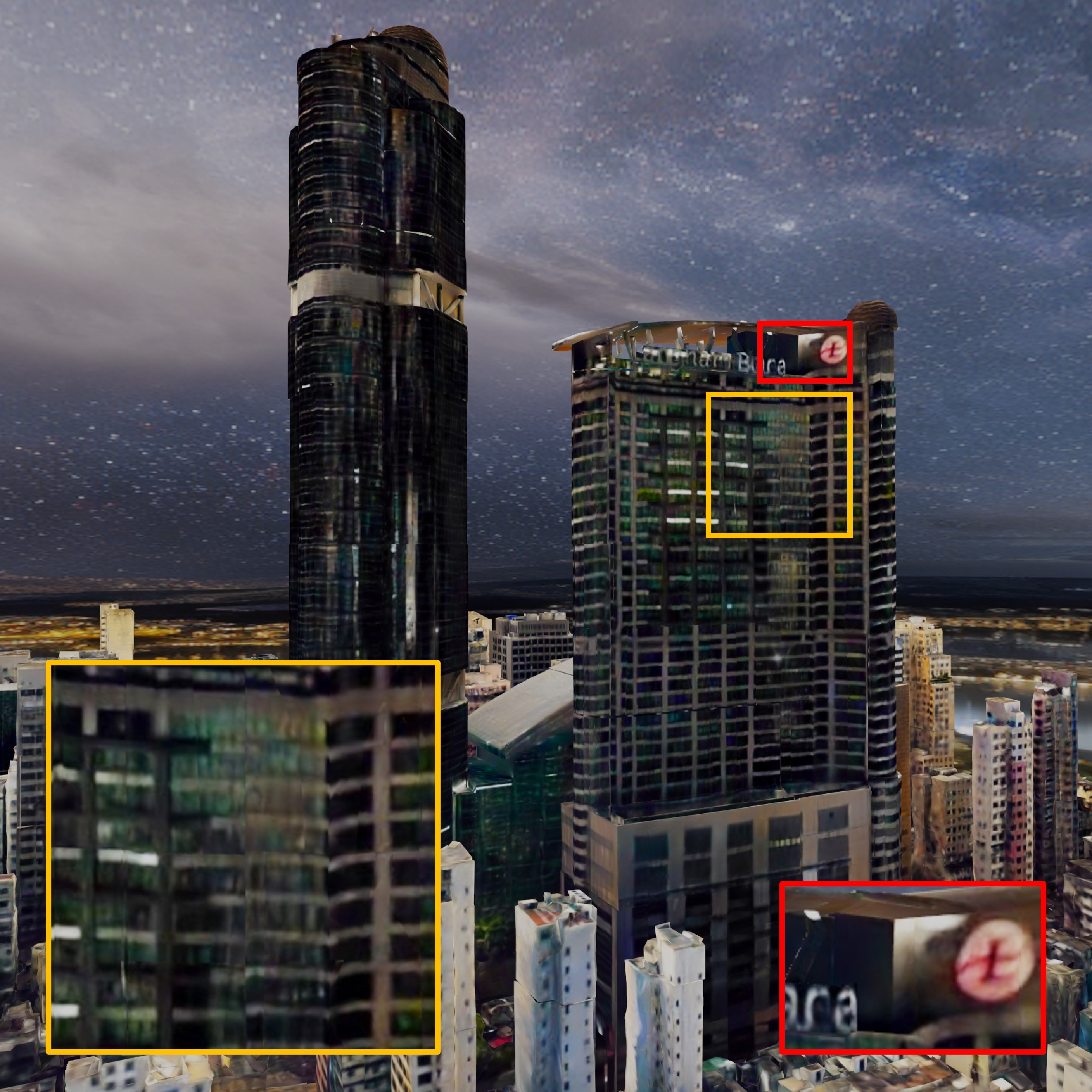}
        \caption{-$\mathcal{L}_{ls}$}
        \label{fig:exp_ab_local}
    \end{subfigure}    
    \begin{subfigure}[b]{\fabwidth\linewidth}
        \centering
        \resizebox{\textwidth}{!}{
        \begin{tabular}{c c c c}
            \toprule
             Ablation & eSSIM $\uparrow$ & LPIPS $\downarrow$ & CLIP Score $\uparrow$ \\
             \toprule
             - aug & 0.595 & 0.289 & 0.302 \\
             - prog &  0.605& 0.285& 0.285 \\
             - scale & 0.598& 0.282& 0.304 \\
             - $\mathcal{L}_{ls}$ &0.633& 0.234& 0.307 \\
             - $\mathcal{L}_{clip}$ &\textbf{0.643}& \textbf{0.232}& 0.293 \\
             Full model &0.633& 0.234& \textbf{0.309} \\
             \bottomrule
        \end{tabular}}
        \caption{Evaluation}
    \end{subfigure}  
    \begin{subfigure}[b]{\abwidth\linewidth}
        \centering
        \includegraphics[width=0.9\textwidth]{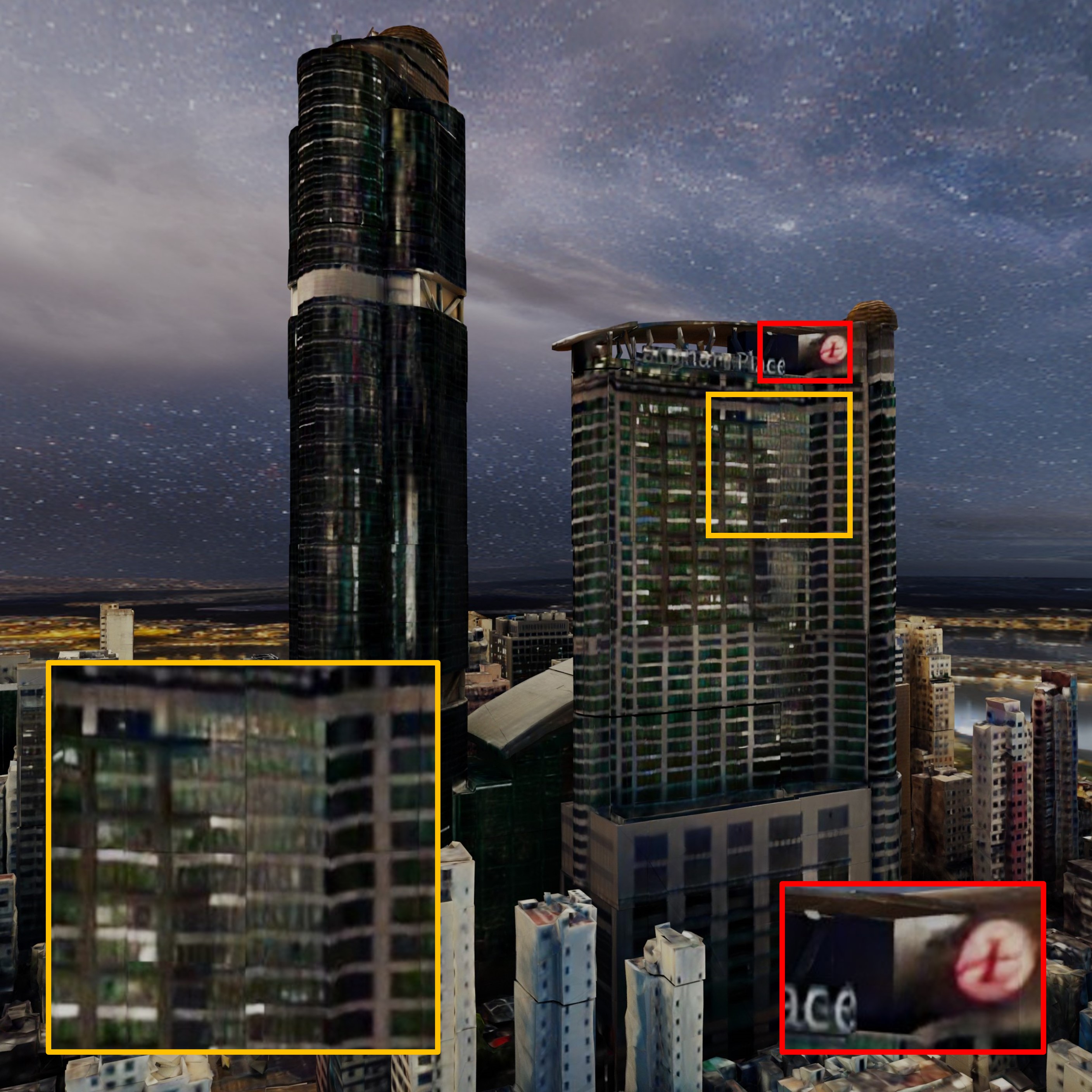}
        \caption{-$\mathcal{L}_{clip}$}\label{fig:exp_ab_clip}
    \end{subfigure}
    \begin{subfigure}[b]{\abwidth\linewidth}
        \centering
        \includegraphics[width=0.9\textwidth]{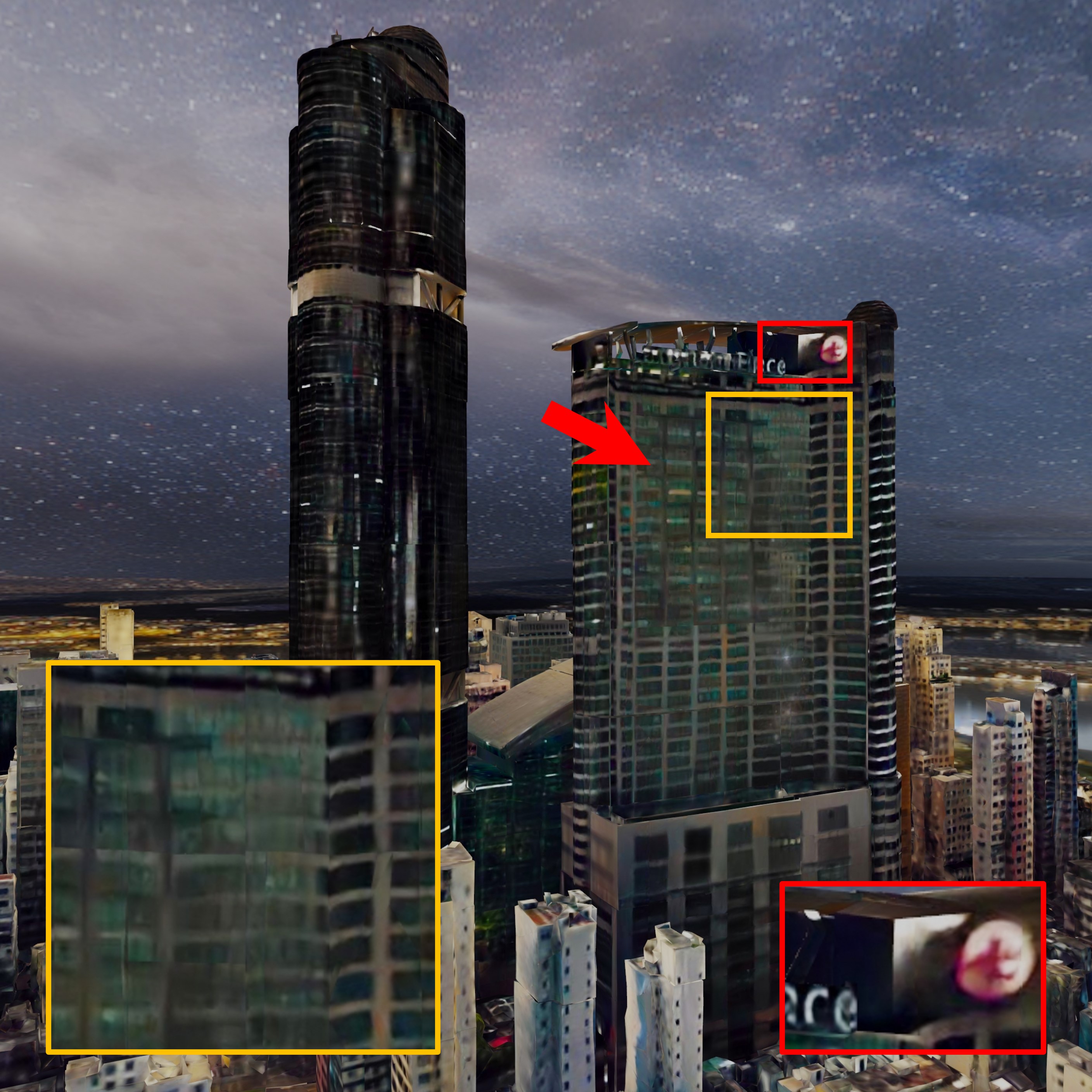}
        \caption{-scale}\label{fig:exp_ab_scale}
    \end{subfigure}
     \begin{subfigure}[b]{\abwidth\linewidth}
        \centering
        \includegraphics[width=0.9\textwidth]{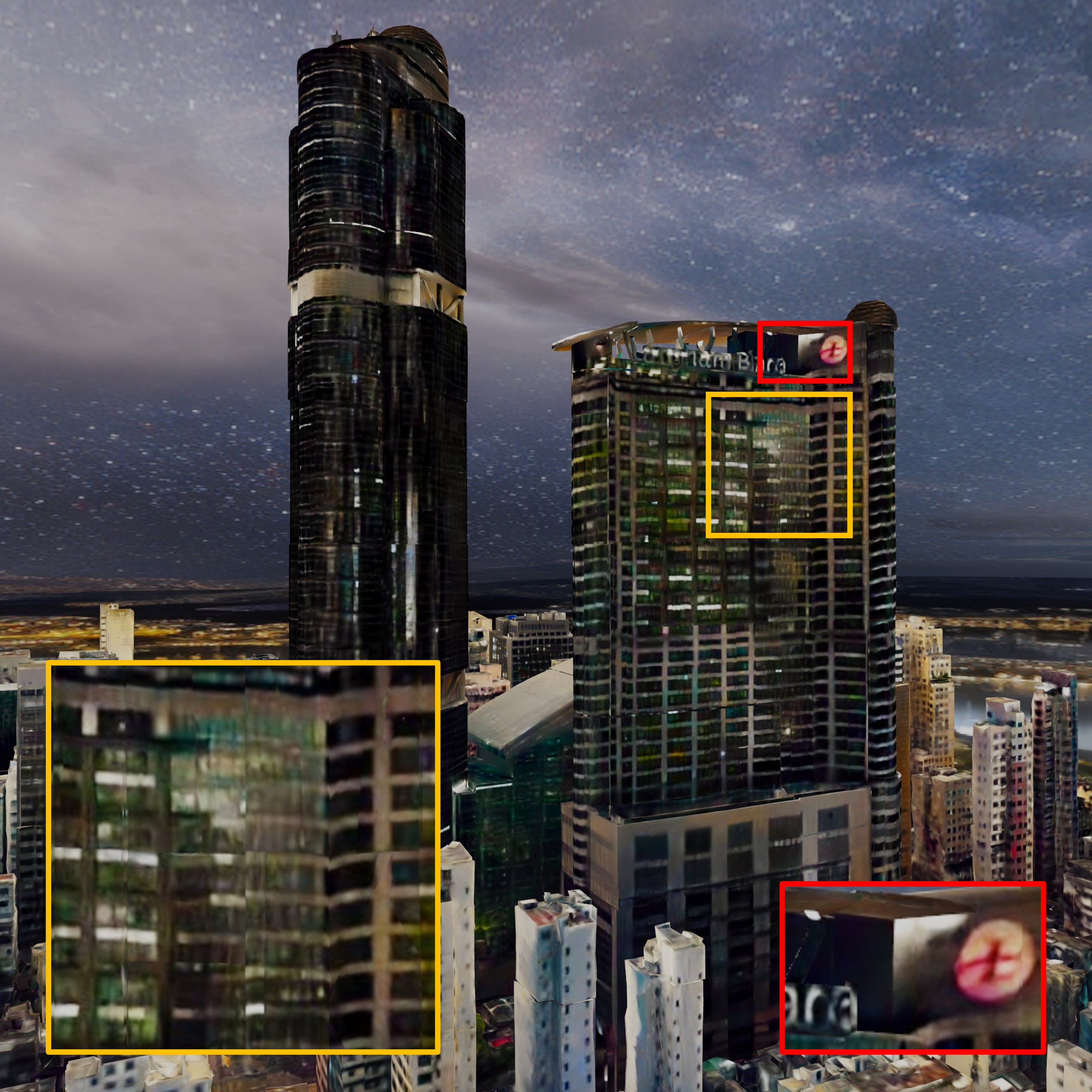}
        \caption{Full model}\label{fig:exp_ab_full}
    \end{subfigure}
    \caption{Ablation studies. Without (b) view augmentation, (c) progressive optimization or (g) scale-adaptive styles, it may perform worse on sharp content preservation. Without (d) local style loss, (f) CLIP loss or (g) scale-adaptive styles, it failed to perform some detailed effects, such as lit windows.}
    \label{fig:exp_ablation}
\end{figure*}

\subsubsection{Ablation Study.}
Fig. \ref{fig:exp_ablation} shows comparisons for ablation studies. We randomly pick a scene with 4 different styles for quantitative evaluation.

\noindent\textit{Effect on augmented view planning and progressive optimization.}
Augmentation with novel view sampling and view translation (\textit{aug}), and multi-scale progressive optimization (\textit{prog}) assist in high-fidelity content structure preservation. As shown in Fig.~\ref{fig:exp_ab_aug} and \ref{fig:exp_ab_prog}, without either of them the system only restores rough textures of blurry logos and windows with worse eSSIM and LPIPS performance.

\noindent\textit{Effect on semantics style loss.} 
Without local semantics-aware style loss ($\mathcal{L}_{ls}$) or textual semantic CLIP loss ($\mathcal{L}_{clip}$), the system generates fewer detailed novel patterns such as lit windows (Fig. \ref{fig:exp_ab_local} and \ref{fig:exp_ab_clip}). It leads to fewer style changes but may improve original content similarity, e.g., slightly better eSSIM and LPIPS. 

\noindent\textit{Effect on scale-adaptive style optimization.} 
Scale-adaptive style references help with robust stylization. In some challenging scenarios, unmatched-scale style can dull colors and even destroy geometry structure. For example, the subtle window patterns in the style image compromised geometric definition and the synthesis of lit windows, as shown in the area highlighted by the arrow in Fig. \ref{fig:exp_ab_scale}.

\begin{figure*} [!t]
    \def\wsc{0.244}
    \def\hsc{0.24}
    \def\wscthree{0.72}
    \centering    
    \includegraphics[width=\wsc\linewidth]{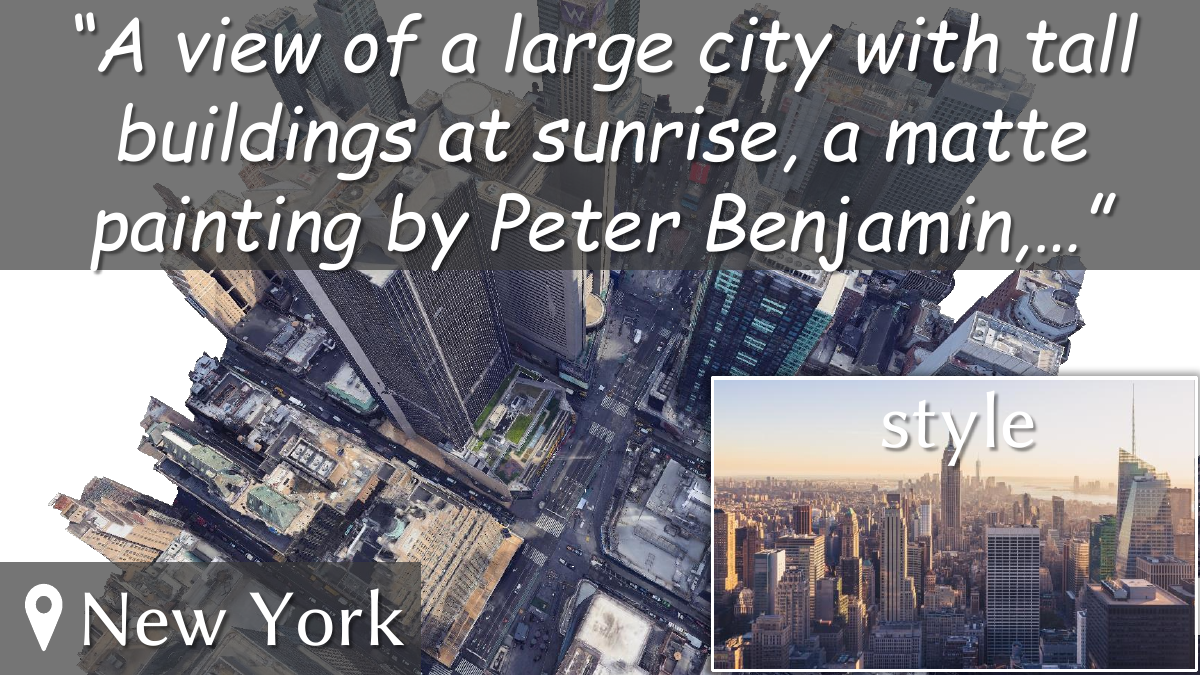}
    \includegraphics[width=\wsc\linewidth]{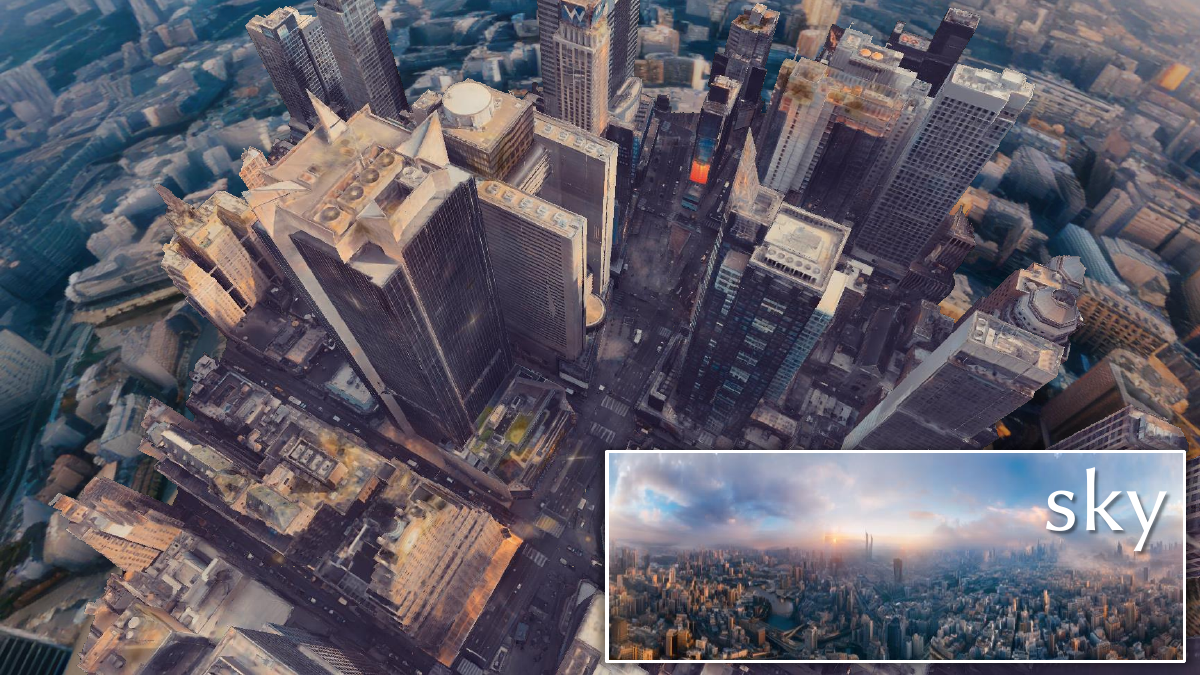}
    \includegraphics[width=\wsc\linewidth]{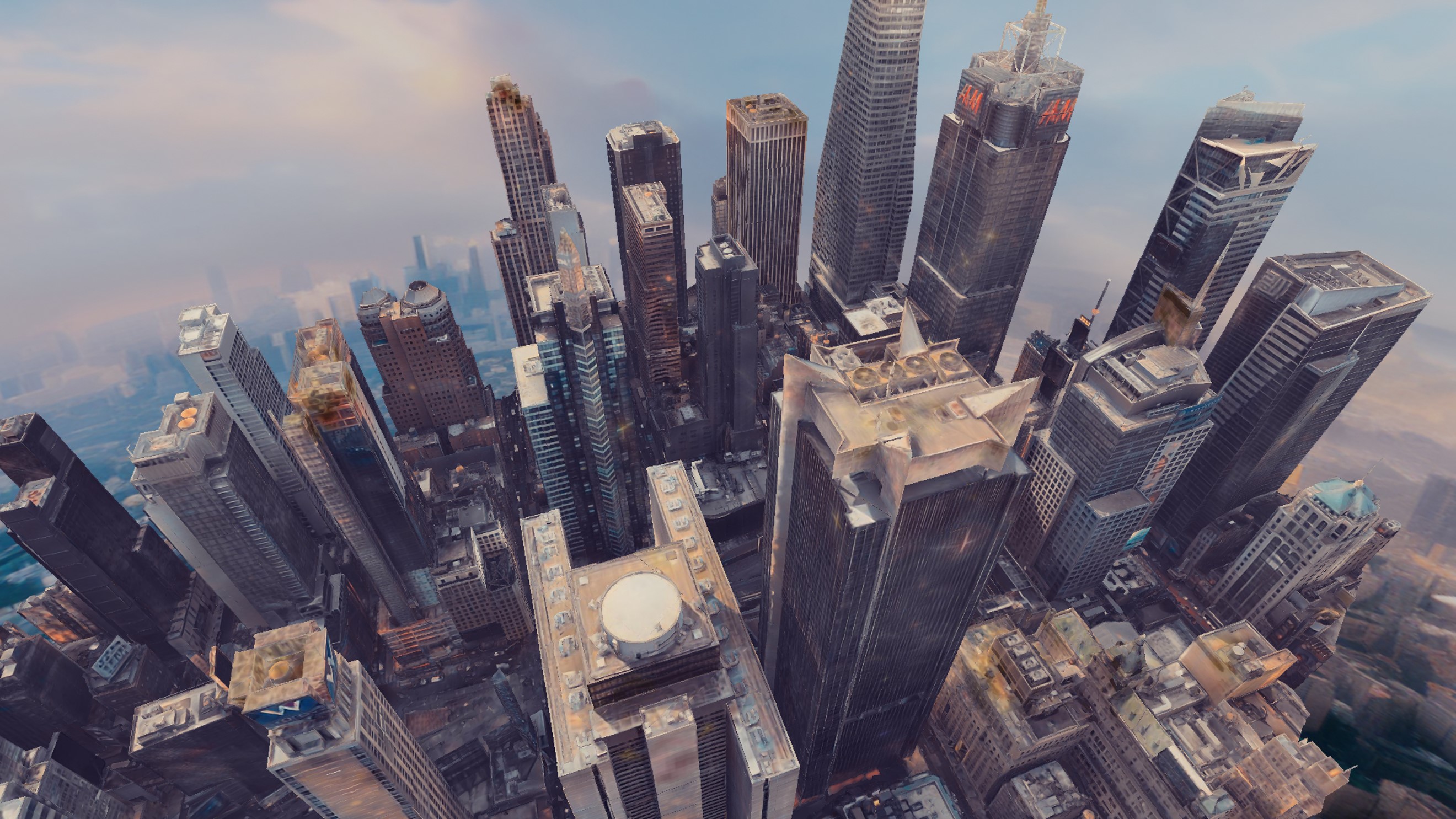}
    \includegraphics[width=\wsc\linewidth]{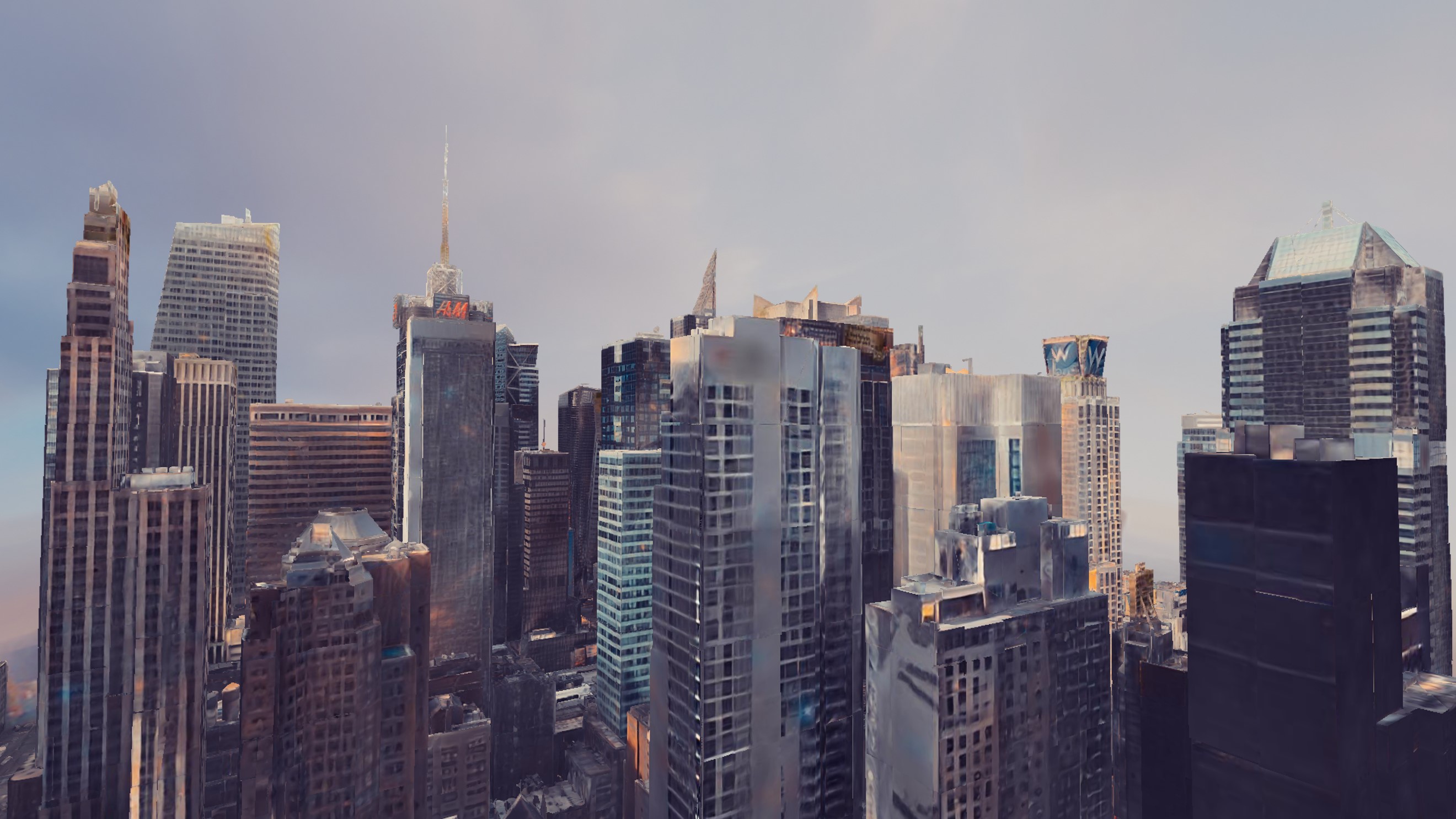}\\

    \includegraphics[width=\wsc\linewidth]{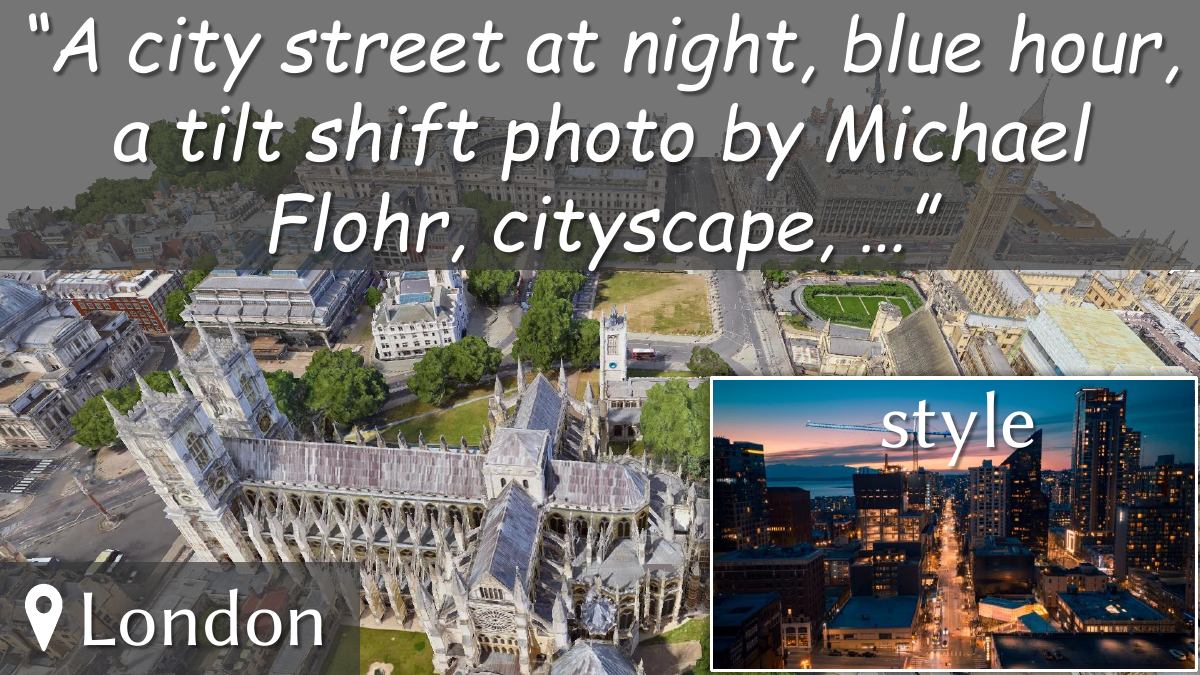}
    \includegraphics[width=\wsc\linewidth]{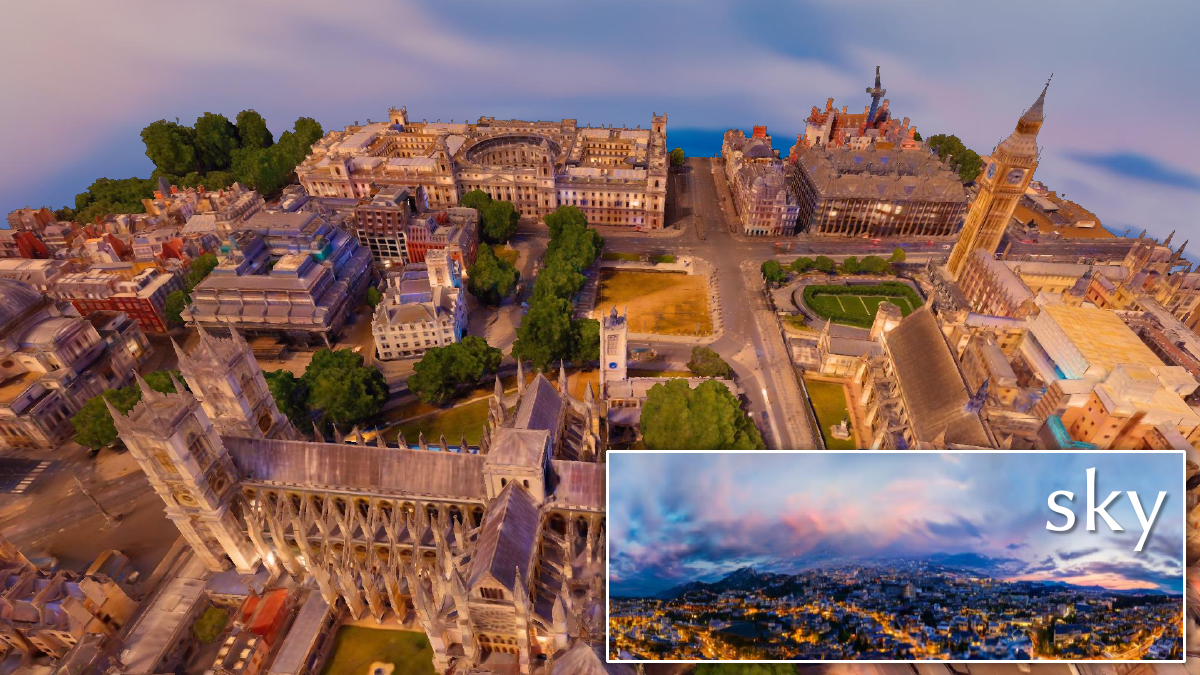}
    \includegraphics[width=\wsc\linewidth]{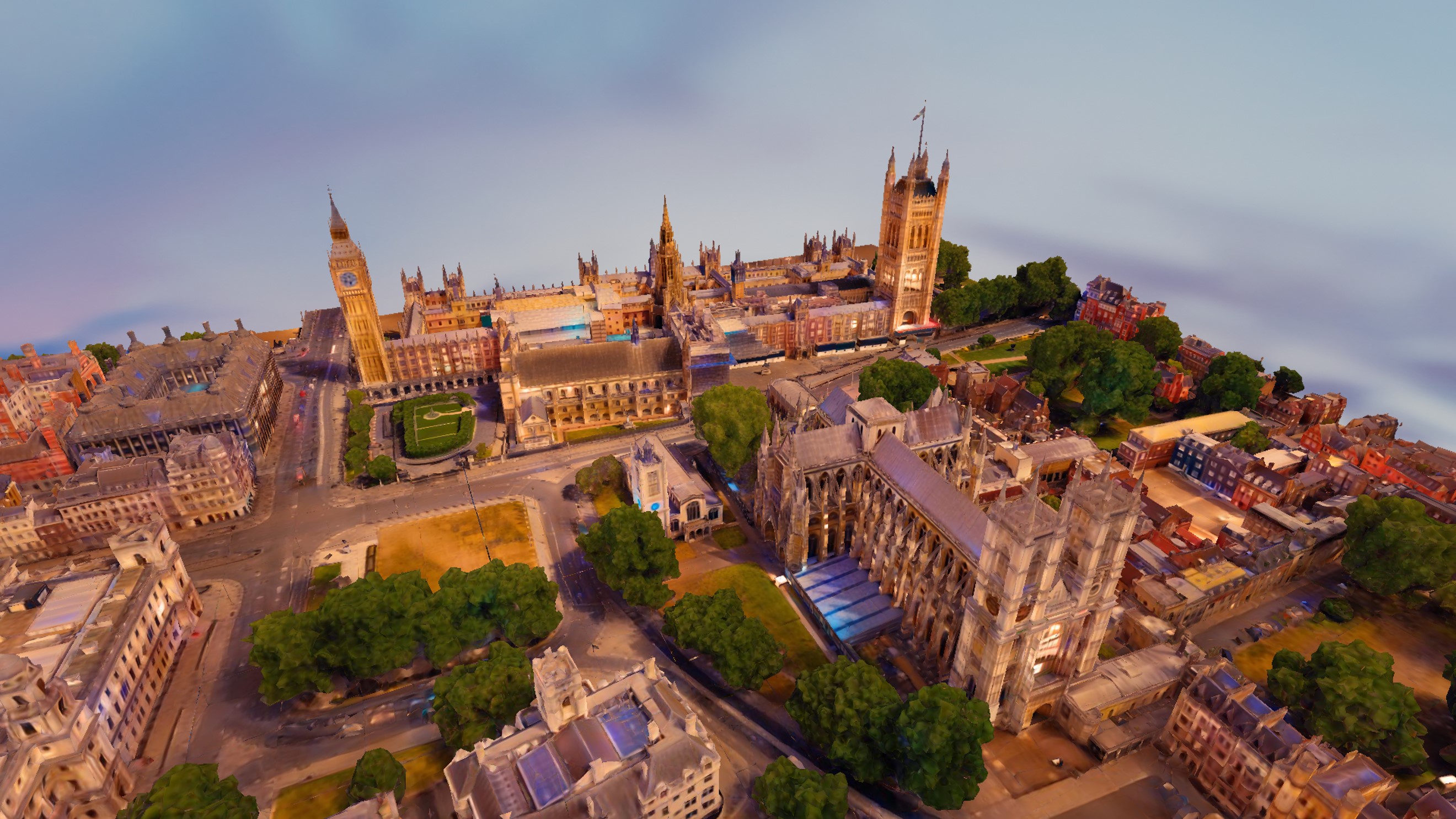}
    \includegraphics[width=\wsc\linewidth]{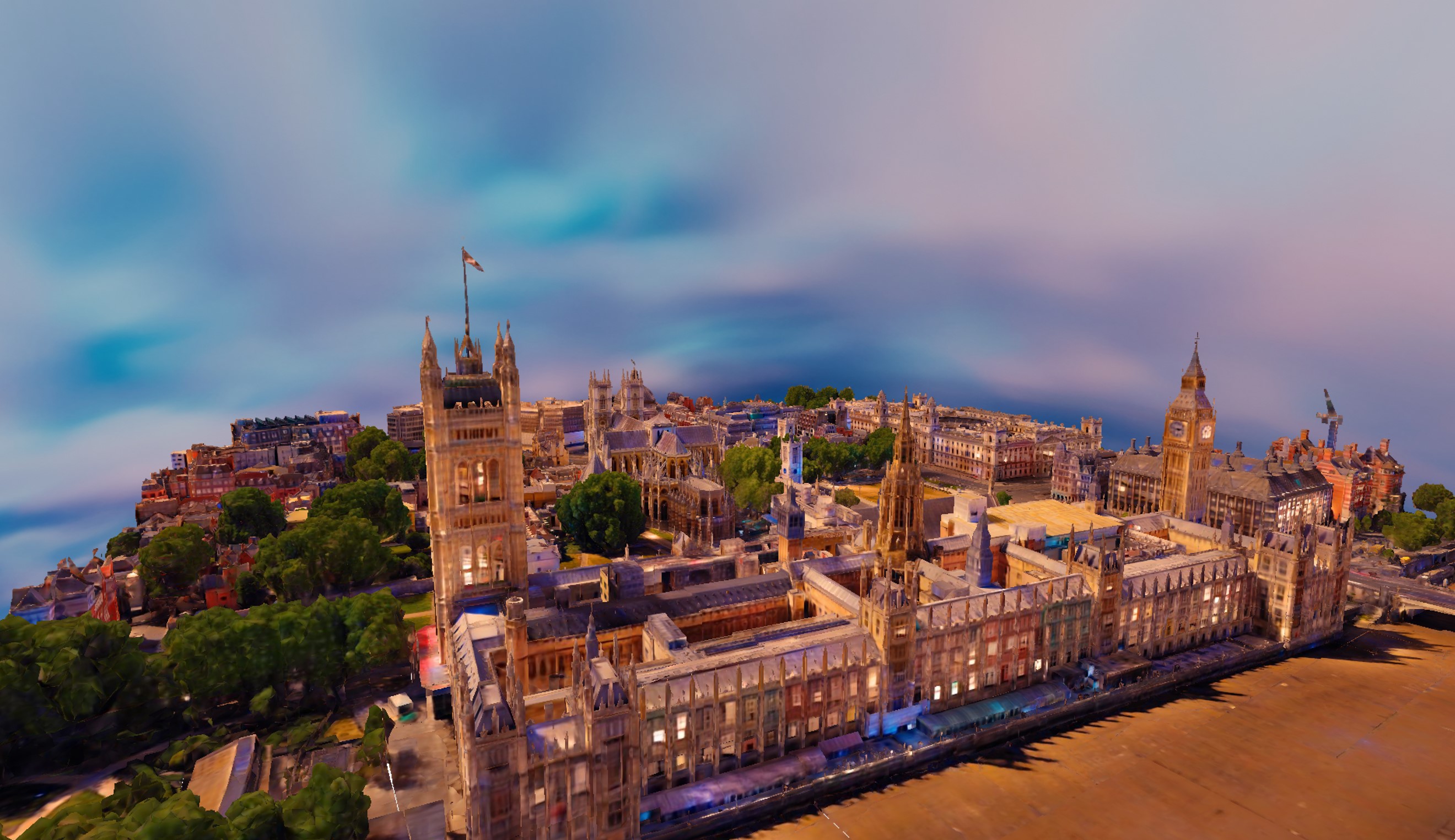}\\
    
    \includegraphics[width=\wsc\linewidth]{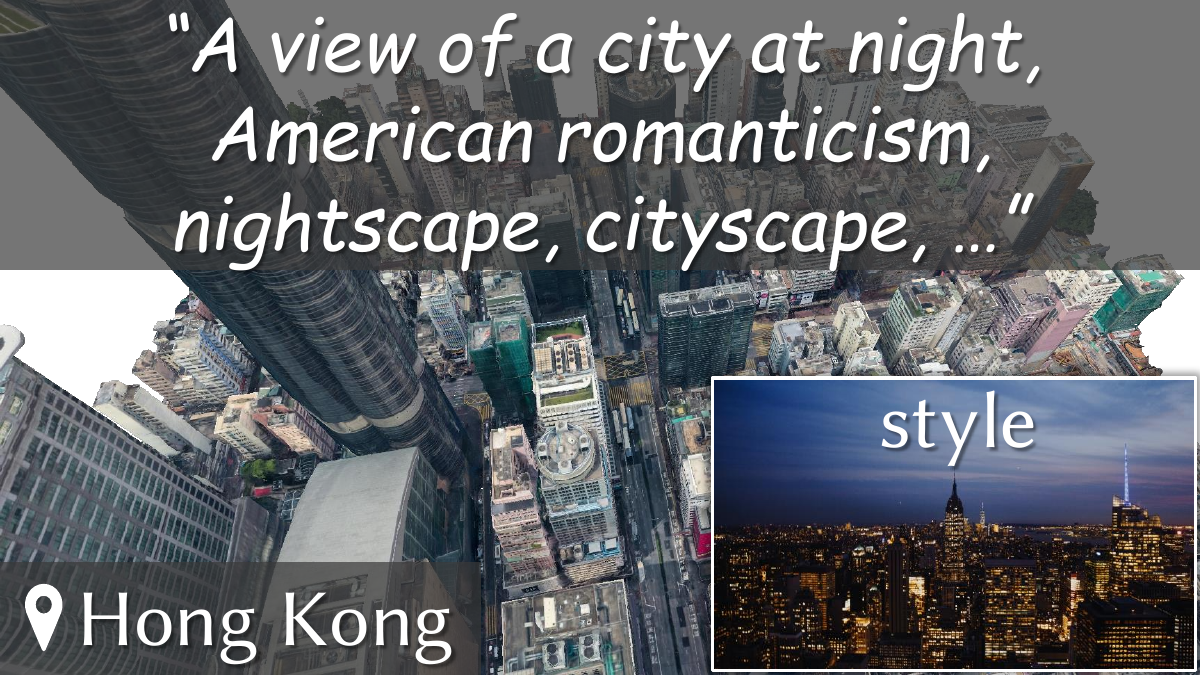}
    \includegraphics[width=\wsc\linewidth]{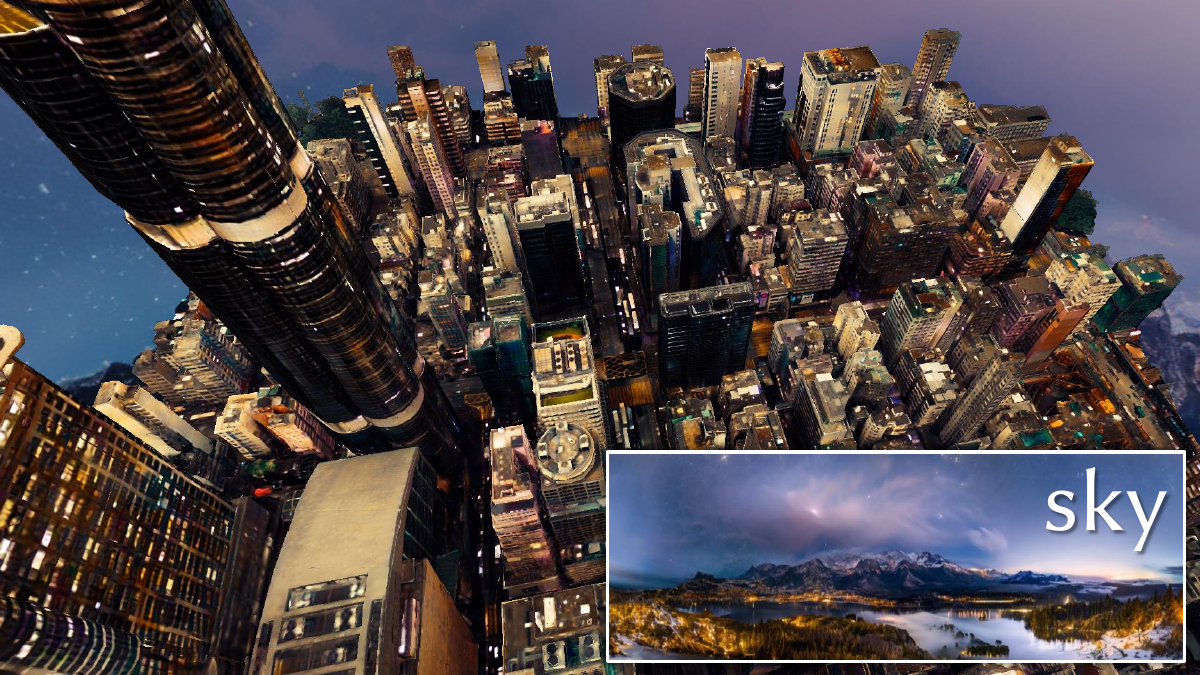}
    \includegraphics[width=\wsc\linewidth]{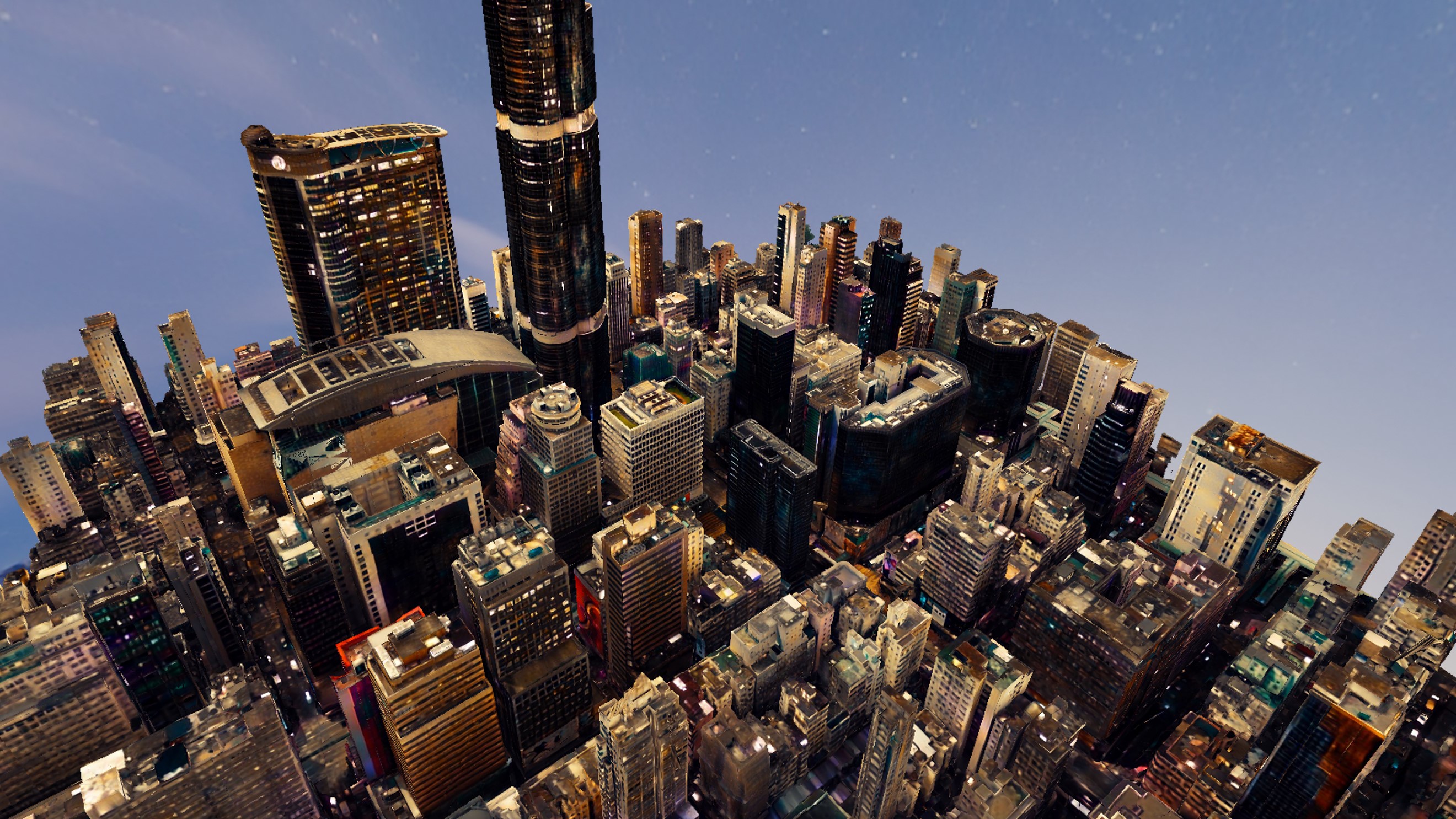}
    \includegraphics[width=\wsc\linewidth]{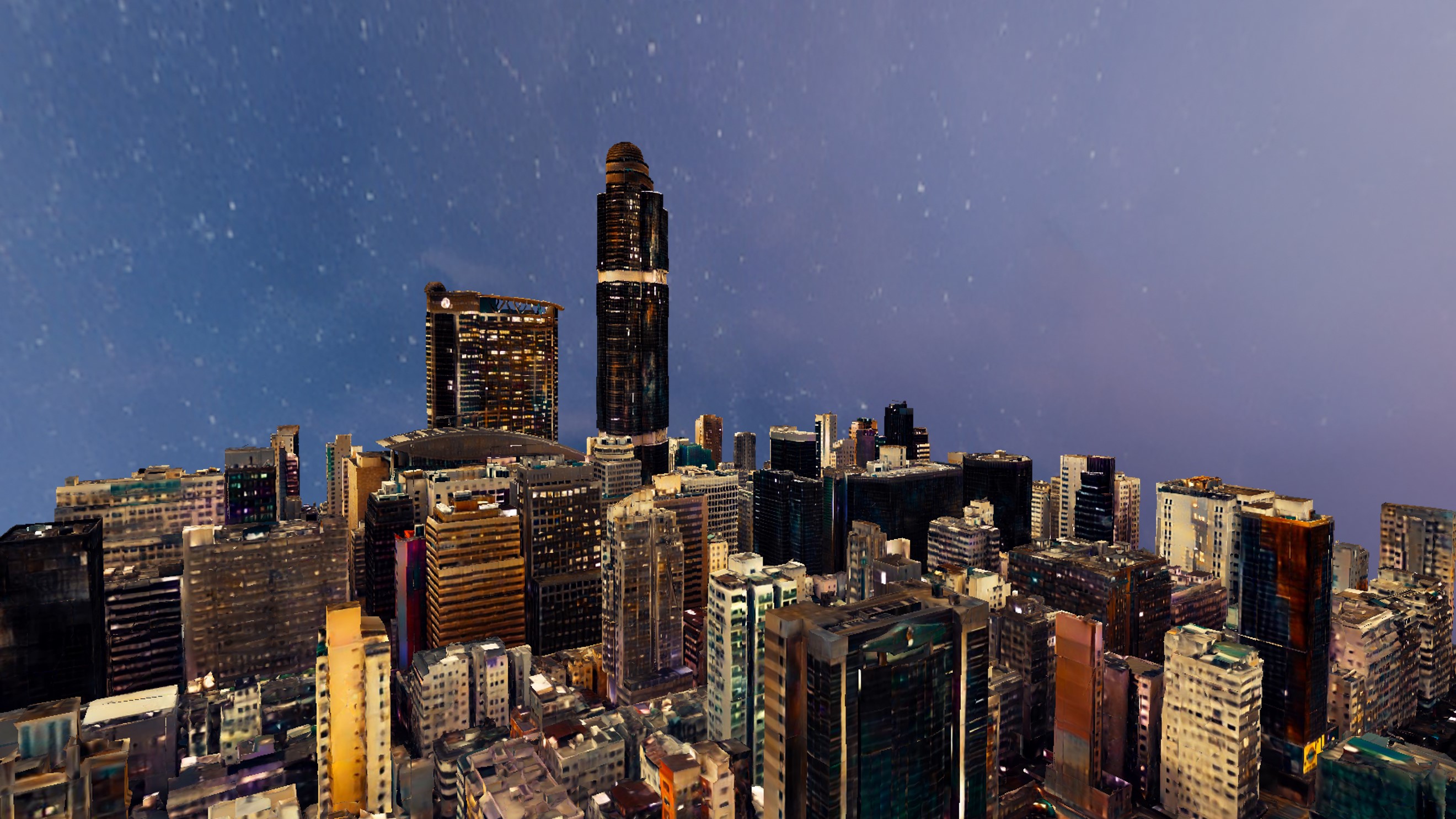} \\
    
    \includegraphics[width=\wsc\linewidth]{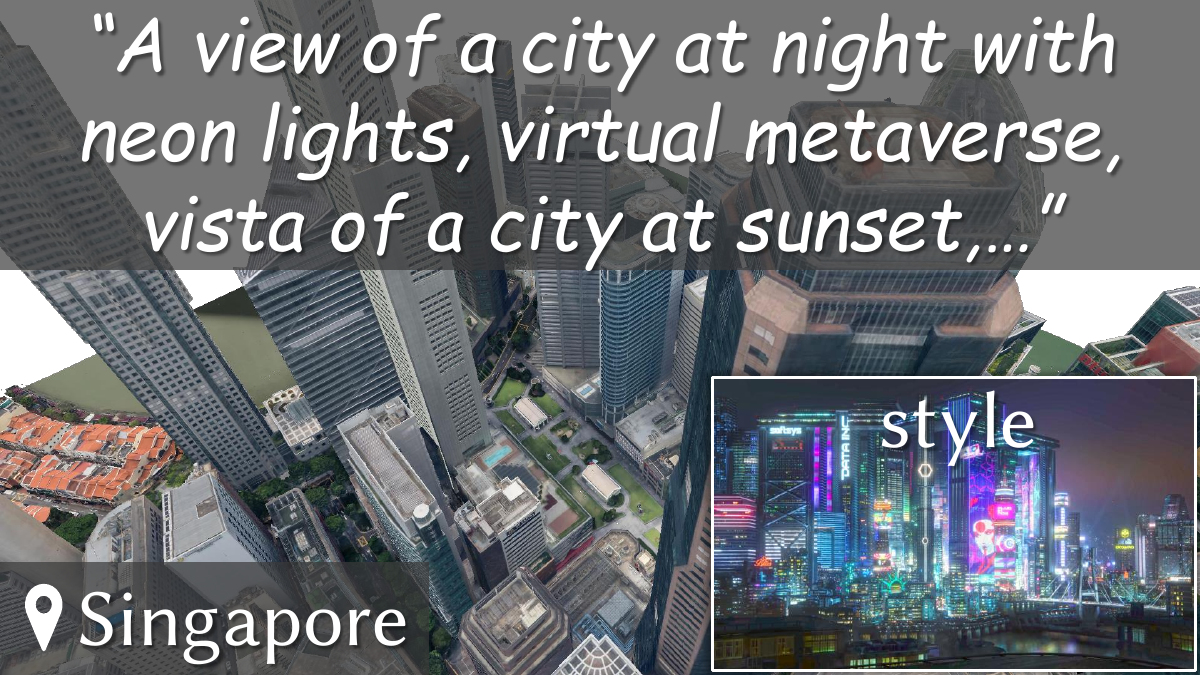}
    \includegraphics[width=\wsc\linewidth]{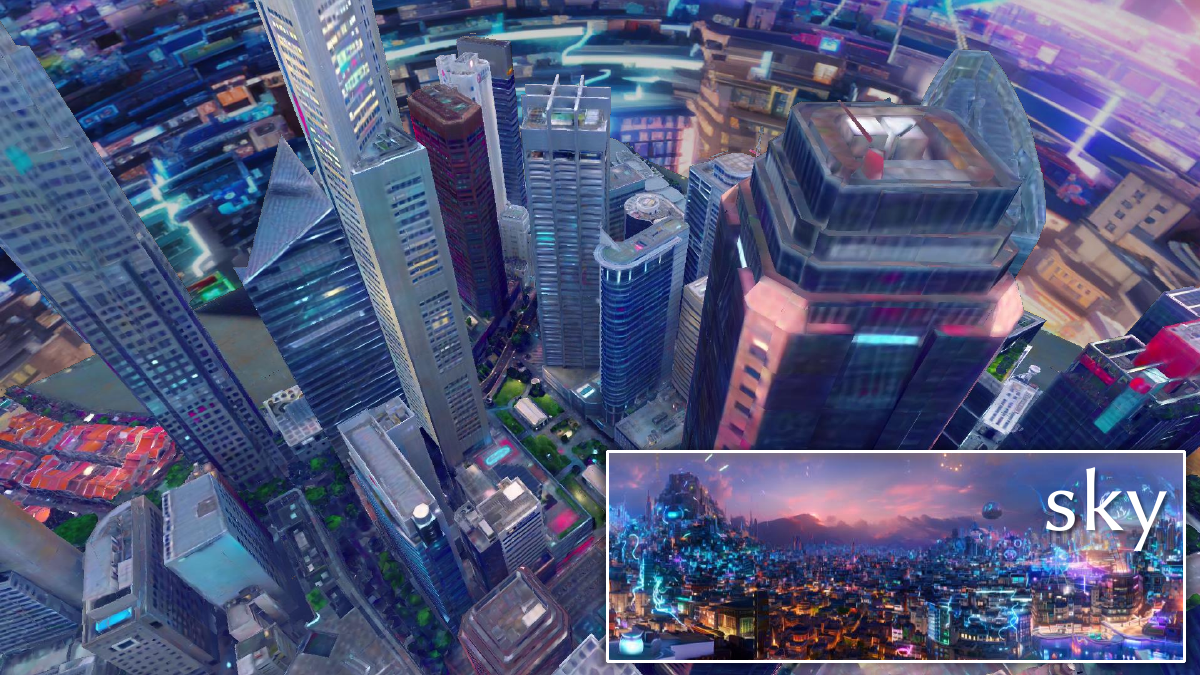}
    \includegraphics[width=\wsc\linewidth]{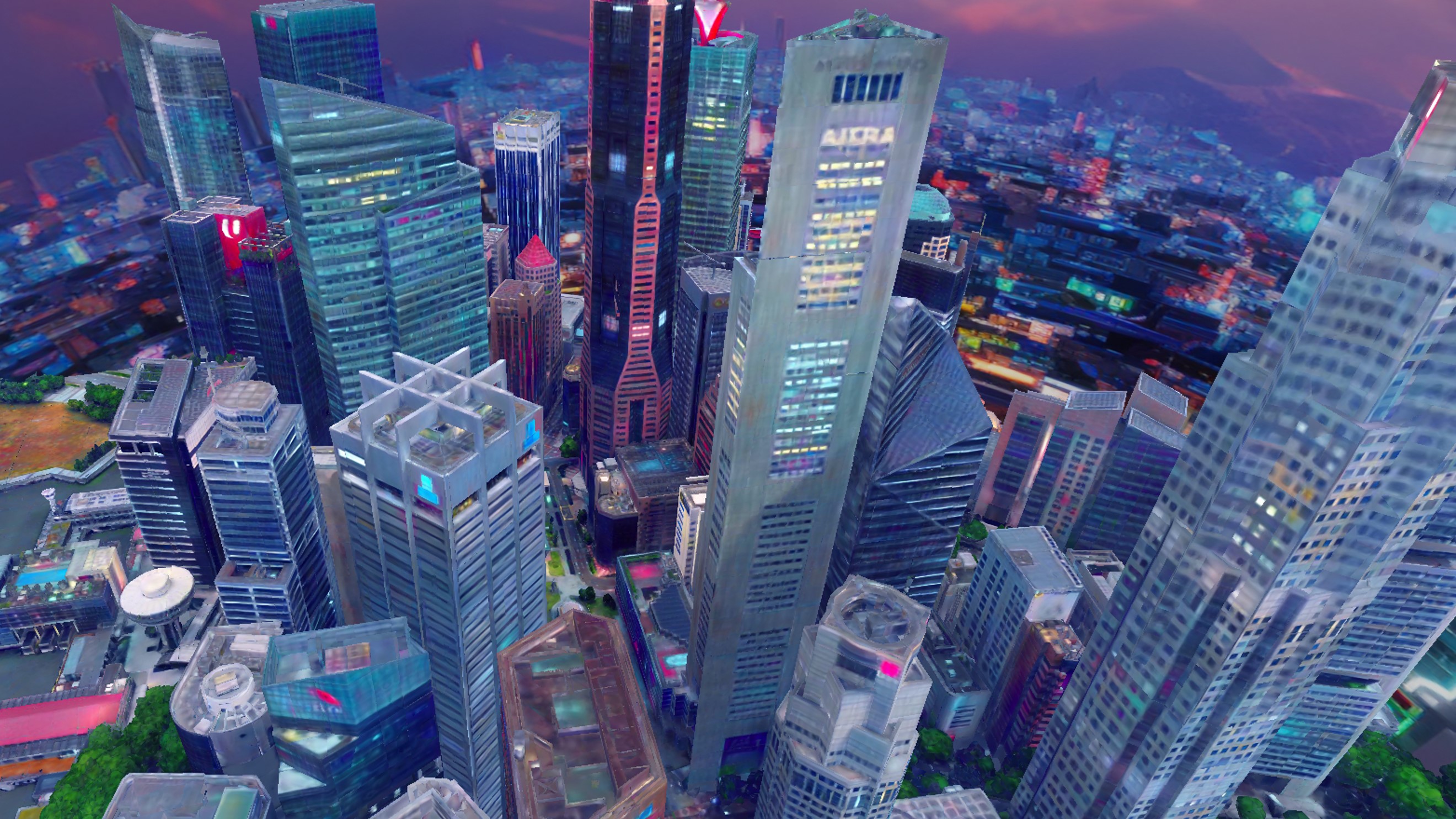}
    \includegraphics[width=\wsc\linewidth]{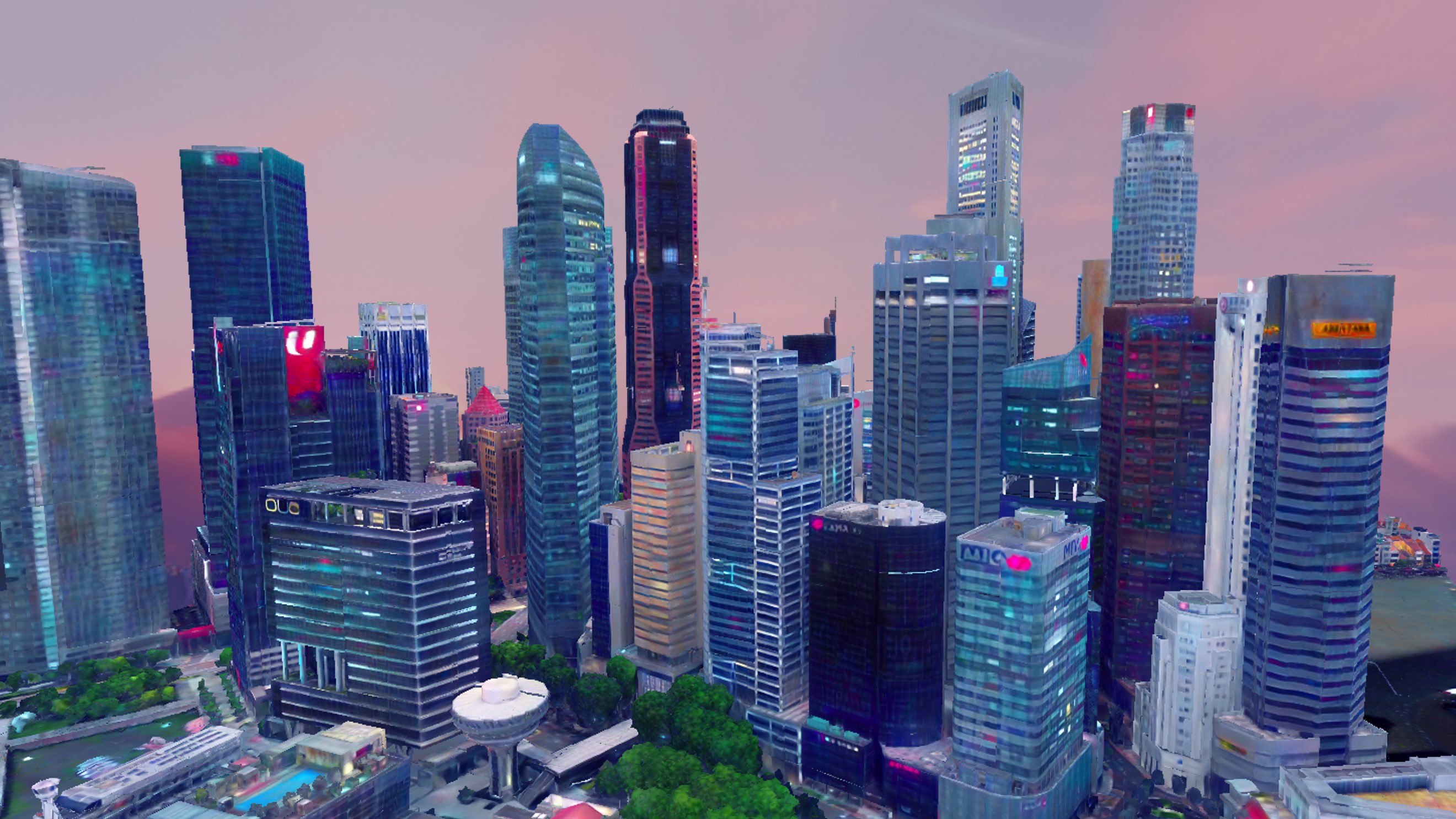} \\
    
    \includegraphics[width=\wsc\linewidth]{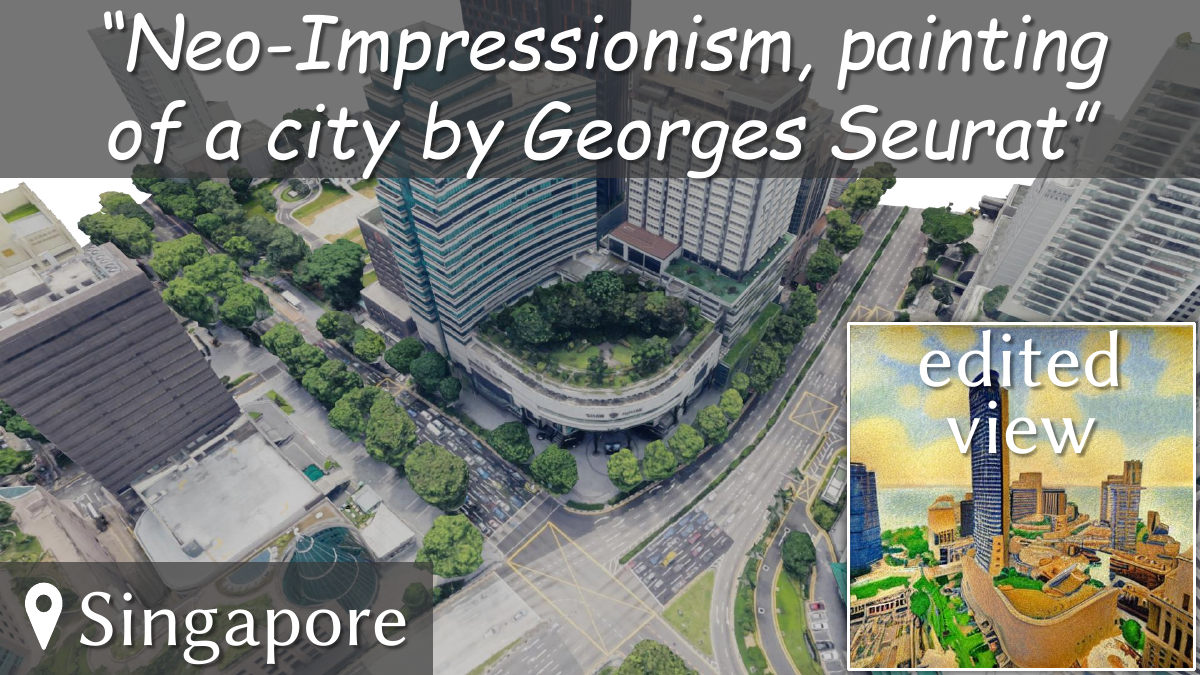}
    \includegraphics[width=\wsc\linewidth]{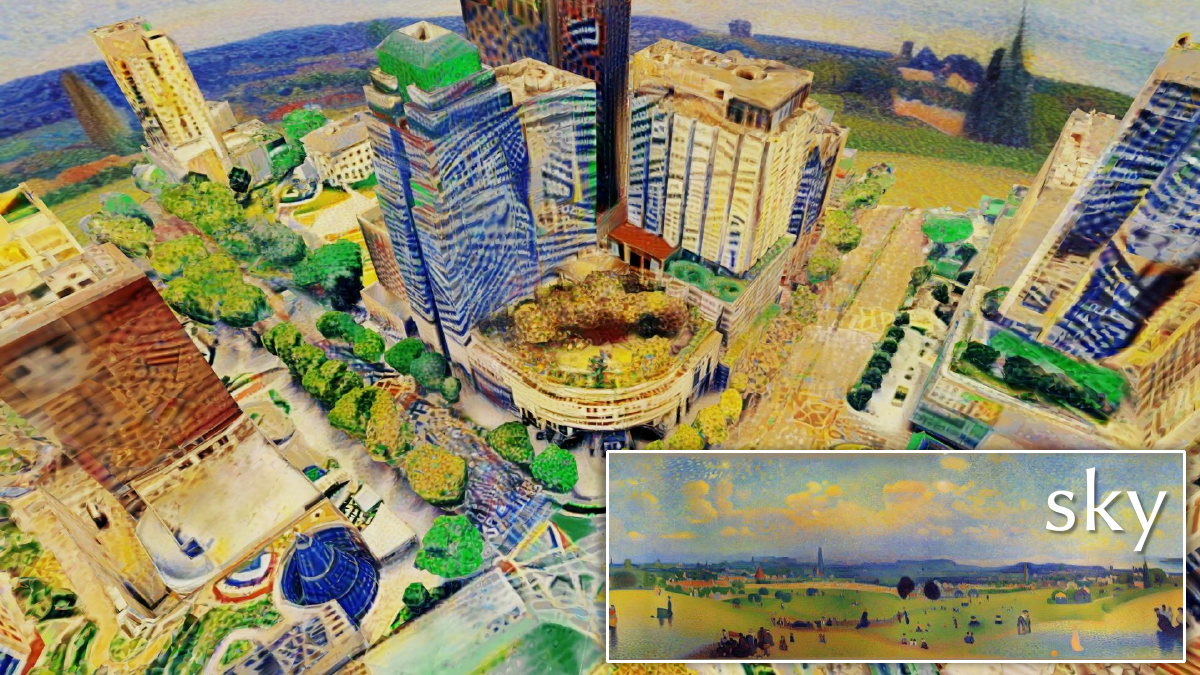}
    \includegraphics[width=\wsc\linewidth]{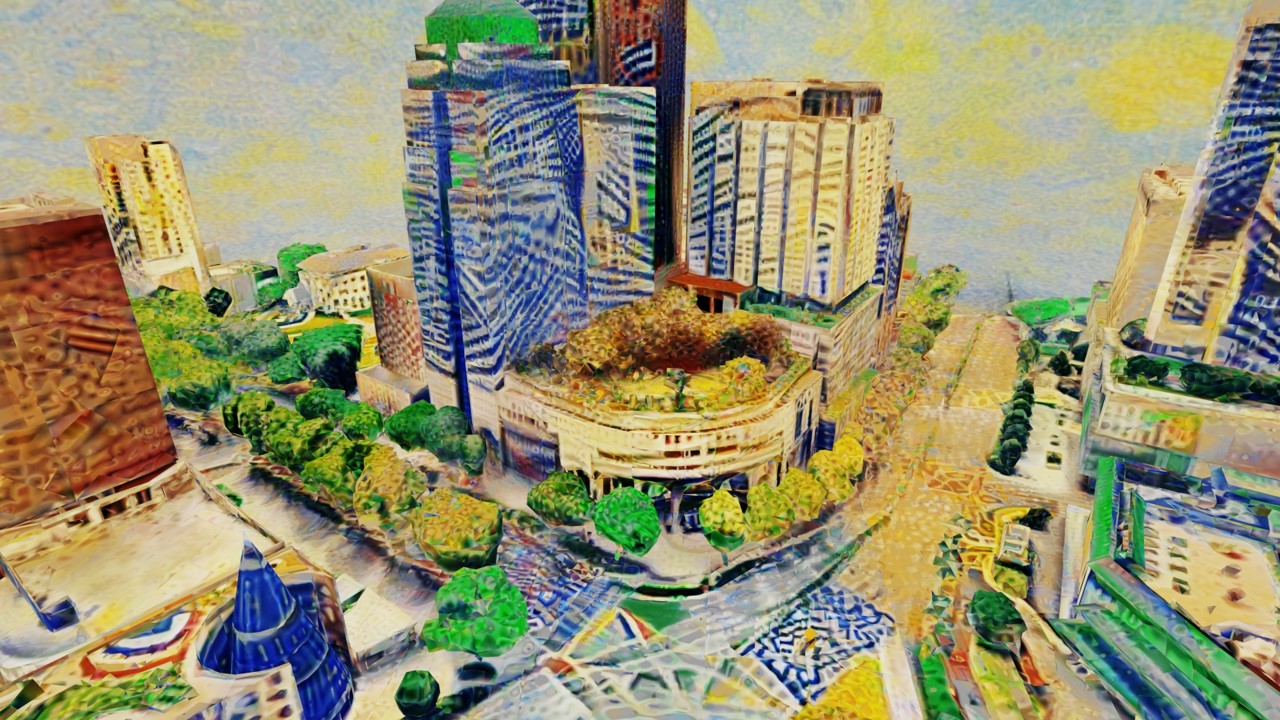}
    \includegraphics[width=\wsc\linewidth]{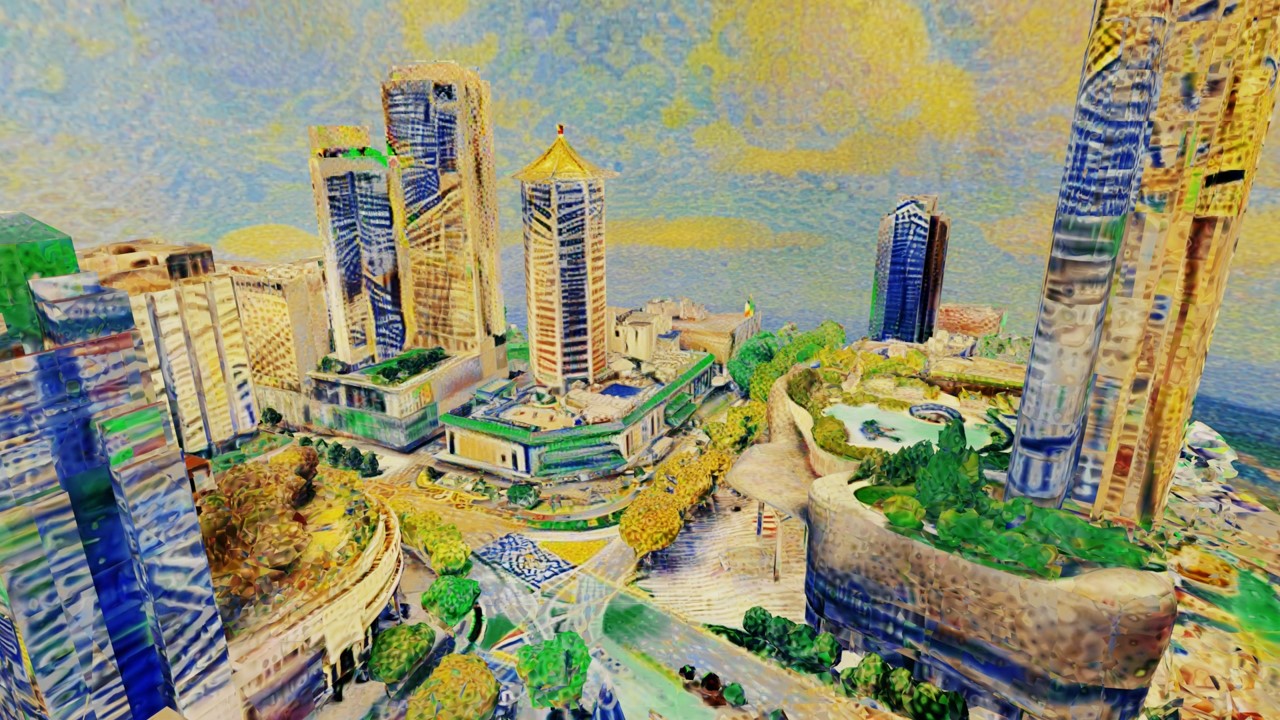}
    
    \begin{subfigure}[b] {\wsc\linewidth}
        \centering
        \caption{Input}
    \end{subfigure}
    \begin{subfigure}[b] {\wscthree\linewidth}
        \centering
        \caption{Stylized results}
    \end{subfigure} 
    \caption{More stylization results of our proposed StyleCity. In the second-to-last row, a style reference from the "Cyberpunk 2077" video game~\cite{cyberpunk_2023} demonstrates a result for futuristic style transfer. The last row displays the artistic 2D-to-3D editing result, and it tends to generate novel textual semantic patterns unseen from image reference. }
    \label{fig:exp_more_results}
\end{figure*}

\subsubsection{Applications.}
Our proposed system enables users, especially non-experts, to have the option to personalize their 3D mesh models by selecting different reference images or text prompts. 
For example, the system can be used to "imitate" \textbf{\it (1) Time-of-Day Effect} shown in Fig. \ref{fig:teaser}. We can tailor the visual appearance of the virtual environments to our preferred time, creating a more personalized and engaging experience.  
Our system also supports {\it (2) Non-photorealistic Stylization}.
For artistic reference, we optimize texture without initialization and set $\lambda_{pht}=0, \lambda_{gs}=10, \lambda_{ls}=1$. 
In addition, StyleCity supports {\it (3) 2D-to-3D Style Editing and Propagation}. This feature allows users to apply any 2D stylization method, such as image- or text-guided style editing \cite{zhang2023adding} and manual editing, to edit a 2D view $s_{edited}$ at a specific viewpoint of the scene. To ensure consistency, during optimization we additionally employ the penalty term \(\mathcal{L}_{edited}=\|s_{edited}-z_{edited}\|^2_2\) to the stylized image $z_{edited}$ at this viewpoint, and use the edited image $s_{edited}$ as the style reference. The last two rows in Fig. \ref{fig:exp_more_results} display examples of the futuristic style of "Night City" in Cyberpunk 2077~\cite{cyberpunk_2023} and 2D-to-3D editing in the artistic "Neo-Impressionism" style.

\section{Conclusion}
We present StyleCity, a vision-and-text driven stylization pipeline for large-scale city-level textured mesh stylization. 
We utilize a neural texture field to model the scene appearance and propose a novel multi-scale progressive optimization method achieving high-fidelity stylization. For harmonic stylization, we introduce scale-adaptive style optimization and new loss functions to regularize style features globally and locally.
Besides, we improve the diffusion panorama synthesis method to support style-aligned high-resolution omnidirectional sky synthesis, which serves as background for an immersive atmosphere and better semantics supervision. 
Finally, the output-baked texture supports conventional rendering in real time. 
Our system originally solves challenges in urban scene stylization in high fidelity and a controllable manner, potentially facilitating various applications including virtual production.

\bibliographystyle{splncs04}
\bibliography{bibliography}
\end{document}